\RequirePackage{fix-cm}
\documentclass[smallextended]{svjour3}       % onecolumn (second format)
\smartqed  % flush right qed marks, e.g. at end of proof
\usepackage{graphicx}

\usepackage{caption}
\usepackage{pifont}
\usepackage{times}
\usepackage{epsfig}
\usepackage{graphicx}
\usepackage{amsmath}
\usepackage{amssymb}
\usepackage{yhmath}
\usepackage{epsfig}
\usepackage{graphicx}
\usepackage{amsmath}
\usepackage{amssymb}
\usepackage{bm}
\usepackage{float}

\usepackage{bm}
\usepackage{booktabs}
\usepackage{color}
\usepackage{multirow}
\usepackage{rotfloat}
\usepackage{tabularx}
\usepackage[misc]{ifsym}

\usepackage{multirow}
\usepackage{float}

\usepackage{bbding}
\usepackage{bigstrut}

\usepackage{hyperref}

\newcommand{\red}{\textcolor{red}}

\begin{document}

\title{
Advancing Unsupervised Low-light Image Enhancement: Noise Estimation, Illumination Interpolation, and Self-Regulation
}

\author{
	Xiaofeng Liu\textsuperscript{1, \Letter}  \and
	Jiaxin Gao\textsuperscript{1}    \and
	Xin Fan\textsuperscript{1}    \and
	Risheng Liu\textsuperscript{1}         
}
% \textsuperscript{1}

\institute{ 
	\Letter \hspace{1 mm} Xiaofeng Liu \\
	\hspace*{4.0 mm}  (goodenoughliu@mail.dlut.edu.cn) \\             
%	\at 
%	\textsuperscript{1}\hspace{3mm}International School of Information Science $\&$ Engineering, Dalian University of 
%	\hspace*{4mm}Technology, Dalian, China\\ 
	\at
	\textsuperscript{1}\hspace{3mm}School of Software Technology, Dalian University of Technology, Dalian, China\\
%	\at
%	\textsuperscript{3}\hspace{3mm}Key Laboratory for Ubiquitous Network and Service Software of Liaoning Province, 
%	\hspace*{4mm}Dalian, China
%	\at
%	\textsuperscript{4}\hspace{3mm}Peng Cheng Laboratory, 
%	\hspace*{4mm} Shenzhen, China
}

%\author{Runjia Lin\textsuperscript{2}         \and
%	Jinyuan Liu\textsuperscript{1, 3}        \and
%	Risheng Liu\textsuperscript{2, 3, *}        \and
%	Xin Fan\textsuperscript{2, 3}            
%}
%% \textsuperscript{1}
%%\authorrunning{Short form of author list} % if too long for running head
%
%\institute{ 
%	\Letter \hspace{1 mm} Risheng Liu \\
%	\hspace*{4.0 mm} rsliu@dlut.edu.cn \\             
%	%             \emph{Present address:} of F. Author  %  if needed
%	\at
%	\textsuperscript{1}\hspace{3mm}School of Software Technology, Dalian University of Technology, Dalian, China\\
%	\at
%	\textsuperscript{2}\hspace{3mm}International School of Information Science $\&$ Engineering, Dalian University of 
%	\hspace*{4mm}Technology, Dalian, China\\
%	\at
%	\textsuperscript{3}\hspace{3mm}Key Laboratory for Ubiquitous Network and Service Software of Liaoning Province, 
%	\hspace*{4mm}Dalian, China

\maketitle

\begin{abstract}
Contemporary Low-Light Image Enhancement (LLIE) techniques have made notable advancements in preserving image details and enhancing contrast, achieving commendable results on specific datasets. 
Nevertheless, these approaches encounter persistent challenges in efficiently mitigating dynamic noise and accommodating diverse low-light scenarios. 
Insufficient constraints on complex pixel-wise mapping learning lead to overfitting to specific types of noise and artifacts associated with low-light conditions, reducing effectiveness in variable lighting scenarios.
To this end, we first propose a method for estimating the noise level in low light images in a quick and accurate way. This facilitates precise denoising, prevents over-smoothing, and adapts to dynamic noise patterns.
Subsequently, we devise a Learnable Illumination Interpolator (LII), which employs learnlable interpolation operations between the input and unit vector to satisfy general constraints between illumination and input. Finally, we introduce a self-regularization loss that incorporates intrinsic image properties and essential visual attributes to guide the output towards meeting human visual expectations.
Comprehensive experiments validate the competitiveness of our proposed algorithm in both qualitative and quantitative assessments. 
Notably, our noise estimation method, with linear time complexity and suitable for various denoisers, significantly improves both denoising and enhancement performance. Benefiting from this, our approach achieves a 0.675dB PSNR improvement on the LOL dataset and 0.818dB on the MIT dataset on LLIE task, even compared to supervised methods. 
The source code is available at \href{https://doi.org/10.5281/zenodo.11463142}{this DOI repository} and the specific code for noise estimation can be found at \href{https://github.com/GoogolplexGoodenough/noise_estimate}{this separate GitHub link}.

\end{abstract}

%\linenumbers

\section{Introduction}
As a significant branch of image restoration, Low-Light Image Enhancement (LLIE) aims at improving the visual quality and details of images, to gather brighter, clearer, and more detailed images captured in low-light conditions. 
By improving the quality of images captured under low-light conditions, LLIE finds wide applications in various fields and downstream machine analytics, such as autonomous driving~\cite{Driving}, dark object detection~\cite{Hu2022}\cite{Jiang2024}, and video surveillance~\cite{DBLP:conf/mlsp/PotterGLHNW20}. 
Overall, the development of low-light enhancement technology holds significant importance in improving image quality, enhancing visibility, and increasing efficiency and safety across various domains.

Unsupervised learning approaches in low-light enhancement aim to improve image quality without the need for labeled training data. These methods leverage statistical properties of low-light images or heuristic algorithms to enhance brightness, contrast, and details. 
To direct end-to-end learning for this mapping process in unsupervised way, many methods improve the brightness and contrast of low-light images by performing pixel-wise addition (e.g. adding a pixel-wise map to adjust the curve of brightness~\cite{guo2020zero}) or division(e.g. dividing learned illumination map~\cite{liu2021retinex}\cite{ma2022toward}). 
Despite efforts to uncover relationships and constraints among different quantities during the enhancement process, these methods do not explicitly impose constraints on the statistical distribution of data that aligns with human visual output. Consequently, training guidance solely from input data may result in overfitting to specific types of noise or artifacts. This overall limitation makes these methods less effective in variable and dynamic lighting conditions, where the types of degradation may vary.

\begin{figure*}
	\centering
	\includegraphics[width=0.8\linewidth]{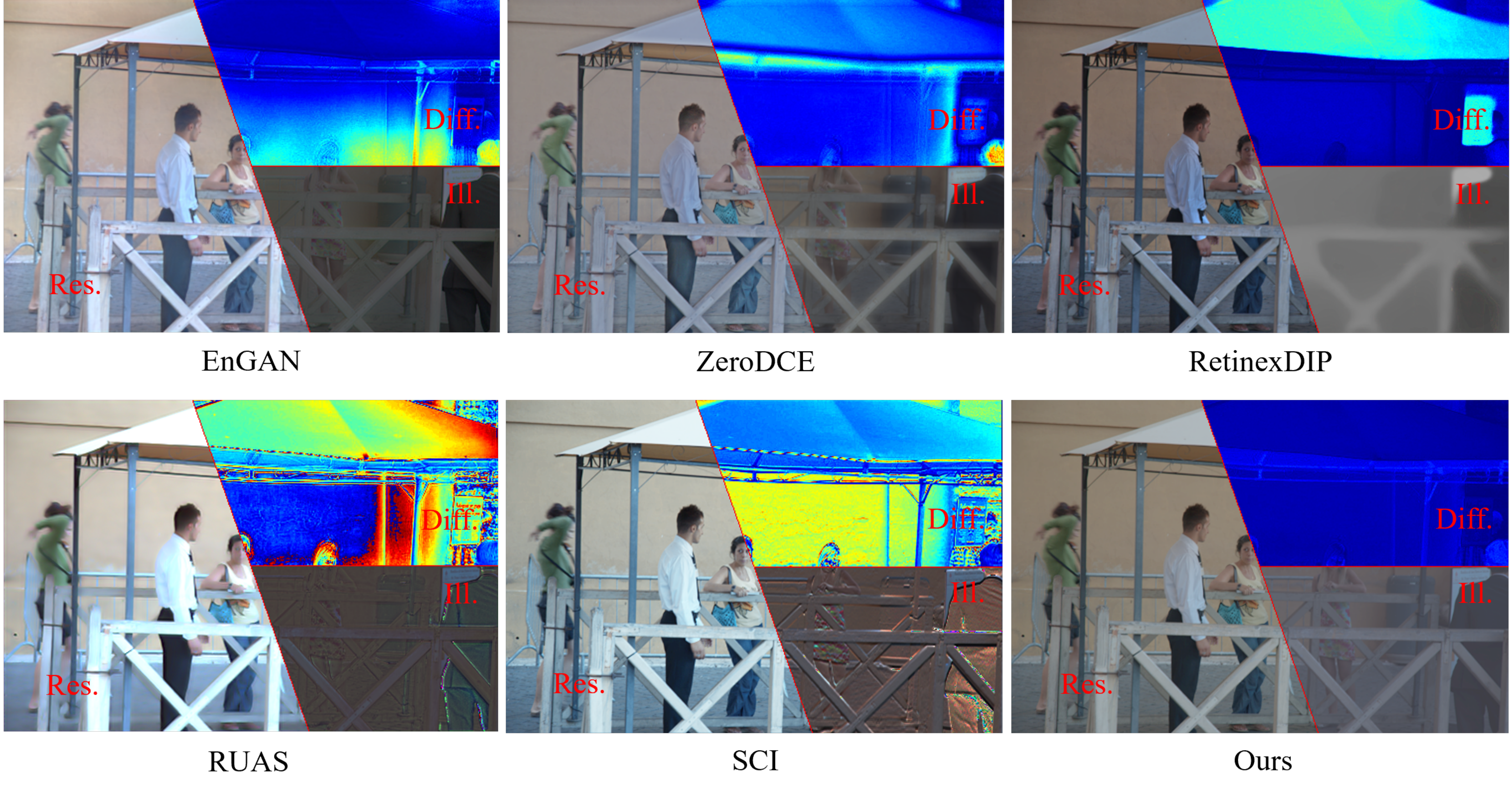}
	\caption{ 
		Comparison among recent state-of-the-art unsupervised methods, including EnGAN~\cite{jiang2021enlightengan}, ZeroDCE~\cite{guo2020zero}, RetinexDIP~\cite{zhao2021retinexdip}, RUAS~\cite{liu2021ruas}, and SCI~\cite{ma2022toward}.
		A scaled heatmap showcasing the absolute error (darker is better) and an illumination map are concatenated to the right part of the enhanced image for comprehensive comparison. Our method closely resembles the ground truth and exhibits a globally smooth yet structure-aware illumination map.
	}
	\label{fig:first}
\end{figure*}

In addition to learning the mapping from dark to bright, denoising is also  a crucial step in LLIE. Because images captured under low-light conditions often exhibit noise, which can be attributed to factors such as low signal-to-noise ratio, high ISO values and dark current noise.
While some methods~\cite{guo2020zero}\cite{ma2022toward} implicitly suppress noise by imposing smoothness constraints on the illumination map, these constraints may not effectively improve the smoothness of the enhancement results. This limitation becomes particularly apparent when dealing with high-intensity noise, making the constraints appear less effective.
On the other hand, certain techniques~\cite{liu2021retinex} explicitly address noise issues by incorporating denoising components. 
However, they do not take into account the dynamic noise distribution in degraded images, leading the model overfit to specific noise and providing limited useful visual enhancement.
To get better denoising results, previous research~\cite{zhang2018ffdnet}\cite{zhang2021plug} has demonstrated the effectiveness of integrating noise intensity as an input for enhancing non-blind denoising methods.
How to estimate the noise intensity in a fast and accurate way in low-light condition to improve the performance is what we concerned.
%\begin{figure}
%	\centering
%	\includegraphics[width=0.7\linewidth]{Fig/rat/inter_resized.pdf}
%	\caption{Compared with state-of-the-art unsupervised methods, II, the Illumination Interpolator, has ability to enhance images with different degradation factors, which simply divide the linear interpolation result between the input image and a unit vector. Details are provided in Section \ref{img_mean_mean}.}
%	\label{fig: inter_res}
%\end{figure}

To address these issues, we propose a denoising-first and enhancing-later pipeline to achieve clear visibility in low-light conditions with dynamic noise.
Firstly, we present a novel noise estimation method that relies on a simple yet efficient formula, by leveraging high-order image gradients. This method enables rapid and accurate estimation of noise levels in low-light settings with linear time complexity, which aids precise denoising, avoids over-smoothing, and enhances adaptability in noise reduction.
%thereby transforming the denoising process from a blind to a non-blind approach.
This adaptation equips our model to effectively handle a wide range of noise levels encountered in low-light conditions.
Then, we observe that even when using linear interpolation to generate an illumination map then using division to get the enhanced result, it helps maintain the brightness of the enhanced image within a stable and reasonable range.
Hence we construct a learnable illumination interpolator (LII) to generate illumination map in a linear way to handle more complex scenarios, instead of leaning complex pixel-wise mapping of illumination directly.
%Meanwhile interpolating with a factor can decrease disparities in pixel values without changing the relative relationships between pixels. These relative relationships are not influenced by the specific pixel locations. Comparatively, the interpolation outcomes between the low-light input and unit vectors exhibit a natural smoothness in terms of pixel relationships when contrasted with the original low-light input. Besides, they uphold a consistent structural integrity within the overall image composition.
Using interpolation maintains pixel value consistencies while retaining natural smoothness in pixel relationships. LII generates globally smooth yet structurally aware illumination maps, as illustrated in Figure~\ref{fig:first}, where the smoothness is  frequently employed as a regularization term to constrain illumination maps. 
%With these features of interpolation, the illumination map generated by LLI is global smooth but structure aware naturally.
Moreover, starting from the pixel intensity distribution from nature images in ImageNet, we employ a relaxed loss function designed using the mean and standard deviation of natural image color statistics. This function constrains the generated target pixel color means to be close to the distribution of colors found in natural images on the color manifold. As a result, the generated image colors appear more natural and realistic. The relaxed nature of the loss function also helps mitigate overfitting, ensuring that the results generated by this method remain relatively stable even in the presence of unknown environmental conditions.
Numerous experiments show that our proposed method consistently outperforms existing unsupervised learning methods, and in some cases, even surpasses supervised learning methods by a significant margin. It consistently ranks first in nearly all metrics, such as achieving a PSNR improvement of 0.675dB on LOL dataset and 0.818dB on MIT dataset.

Our contributions are summarized as follows:
\begin{itemize}
	%\item  Great truths are always simple. we design a simple and efficient module 
	%	\item  According to the constraint principle of illuminance and reflectance within a limited dynamic range, as a prior knowledge in the recovery process, we construct a learnable illuminance interpolator and thereby compensating for non-uniform lighting. 
	\item Leveraging the statistic features of low-light images, we firstly propose a noise intensity estimation method based on image gradients specifically designed for low-light images. 
%	This method allows us to improve the denoiser's performance with little cost. 
	The results indicate that our noise estimation method can efficiently and quickly estimate noise parameter in low-light images and improve the denoiser's performance with little cost. 
	\item  Instead of learning the complex pixel-wise mapping, we mainly learn an interpolation factor and construct a learnable illumination interpolator for generating a global smooth but structure aware illumination representation.
	%	 for LLIE problems. 
	\item  Starting from the properties of natural image manifolds, a self-regularized recovery loss is introduced as a way to encourage more natural and realistic reflectance map. 
	Our method significantly improves enhancing performance on a diverse range of datasets.
	
\end{itemize}

\section{Related Work}
\subsection{Low-light Image Enhancement}
In the realm of LLIE, researchers have delved into an assortment of network models aimed at tackling the amalgamation of degradation factors and improving representations. RetinexNet\cite{wei2018deep} introduced a semi-decoupled approach comprising a Decom-Net for decomposition and an Enhance-Net for illumination adjustment. MBLLEN\cite{lv2018mbllen} employed a deep learning method with multi-branch fusion to extract features from various levels and enhance low-light images through multiple subnets. GLADNet\cite{wang2018gladnet} dealt with LLIE via a Global illumination Aware and Detail-preserving Network. RUAS\cite{liu2021ruas} employed a collaborative reference-free learning searching strategy to discover a suitable network architecture for enhancing low-light images, obviating the need for handcrafting design.
Besides, numerous training strategies have been explored for LLIE to investigate the relationship between low-light and normal images. DRBN~\cite{yang2020fidelity} presented a semi-supervised learning approach utilizing deep recursive band networks to extract coarse-to-fine band representations and reconstitute towards high-quality images. ZeroDCE~\cite{guo2020zero} estimated pixel-wise and high-order curves for dynamic range adjustment of a given image using a lightweight deep network.  EnGAN~\cite{jiang2021enlightengan} introduced a global-local discriminator structure, self-regularized perceptual loss fusion, and attention mechanism to train the network in an adversarial way without paired images.  RetinexDIP~\cite{zhao2021retinexdip} devised a novel "generative" strategy for Retinex decomposition and a unified deep framework for low-light image enhancement, without external image data. SCI~\cite{ma2022toward} put forward a new self-calibrated illumination learning framework to reduce computational cost and accelerate the training process. 

\subsection{Image Denoising}
Previous denoising models, such as IRCNN~\cite{IRCNN} and DnCNN~\cite{DnCNN}, have traditionally relied on supervised learning methods tailored to distinct noise levels, requiring separate networks for different degradation factors.
In contrast, FFDNet~\cite{zhang2018ffdnet} leverages noise estimation as input, demonstrating the capability to effectively handle a wide range of noise levels using a singular model. Building upon this paradigm, both DRUNet~\cite{zhang2021plug} and CBDNet~\cite{CBDNet} exhibit impressive denoising proficiency by integrating noise level inputs and employing a UNet-style denoising architecture. DRUNet directly incorporates noise levels as input for to solve multiple tasks related to image restoration, whereas CBDNet estimates noise levels using convolutional networks.
The challenge arises from the dynamic nature of real-world noise, which is not solely static. This characteristic presents a significant hurdle for neural networks, hindering precise estimation through learning-based methods.
Given the dynamic attributes inherent in real-world noise, the prompt and accurate estimation of noise levels becomes crucial. This dynamic quality poses a considerable challenge for both neural networks and conventional methods in accurately estimating the intensity of dynamic noise.

\subsection{Noise Level Estimation}
Estimating noise levels from a single image is considered an inherently challenging problem. Despite efforts spanning several decades, numerous methods have emerged for this purpose, often operating under the assumption that the processed image contains sufficient flat areas. However, this assumption is not consistently applicable, particularly in scenarios involving the processing of natural images.
Pyatykh~\cite{NoisePCA2012TIP}, Liu~\cite{NoiseLevelEstimation2013TIP}, and Chen~\cite{StatisticalMethod2015TIP} have explored the utilization of Principal Component Analysis (PCA) to model observed signals and infer the noise distribution. While these methods represent advancements, they still possess high time complexity in evaluating noise distribution, rendering them unsuitable for real-time enhancement as a pre-processing progress.

%\section{Preliminaries}
%Derived from the study of the human visual system~\cite{land1977retinex}, Retinex theory serves as a color constancy model with broad applications in mitigating problems associated with uneven lighting and color bias.
%The fundamental tenet of the Retinex theory involves the partitioning of an image into its constituent reflectance and illumination components. The Retinex theory views an image as the product of its reflectance and illumination components. The reflectance component represents the color and texture information of the object's surface, while the illumination component represents the distribution of lighting. In LLIE, the relationship of low-light input $\mathbf{x}$, reflectence $\mathbf{s}$ and illumination $\mathbf{y}$ can be formulated as follows:
%\begin{equation}\label{eq:retinex}
%	\mathbf{x} = \mathbf{y} \otimes \mathbf{s}.
%\end{equation}
%
%In numerous previous studies, a common approach to more effectively segregate the reflectance and illumination elements and diminish the influence of noise and lighting fluctuations on enhancement outcomes involves the assumption of smoothness within the illumination layer.
%Meanwhile, the contours, surface characteristics, and relative placements of objects can impact the lighting distribution. This lighting distribution is, to some degree, correlated with or akin to the content found in the image, encompassing objects, textures, and geometric forms. Consequently, a structural similarity exists between the illumination layer and the low-light input.

\section{Preliminaries}
Derived from the study of the human visual system~\cite{land1977retinex}, Retinex theory serves as a color constancy model with broad applications in mitigating problems associated with uneven lighting and color bias.
The fundamental tenet of the Retinex theory involves the partitioning of an image into its constituent reflectance and illumination components. The Retinex theory views an image as the product of its reflectance and illumination components. The reflectance component represents the color and texture information of the object's surface, while the illumination component represents the distribution of lighting. In LLIE, the relationship of low-light input $\mathbf{x}$, reflectence $\mathbf{s}$ and illumination $\mathbf{y}$ can be formulated as follows:
\begin{equation}\label{eq:retinex}
	\mathbf{x} = \mathbf{y} \otimes \mathbf{s}.
\end{equation}

\section{Proposed Method}

\begin{figure*}[htbp]
	\centering 
	\begin{tabular}{c}
		\includegraphics[width=\linewidth]{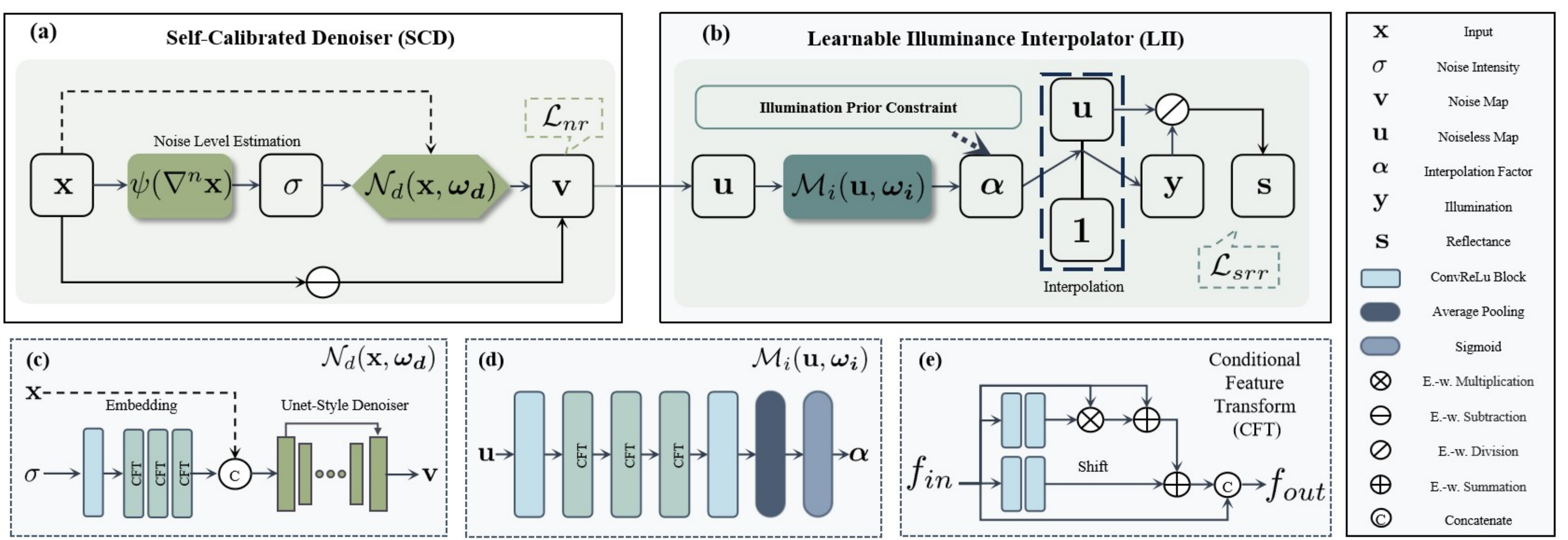}\\
	\end{tabular}%
	\caption{Illustrations of the pipeline and fundamental components. \textit{Top:} (a) Self-Calibrated Denoiser, (b)  Learnable Illumination Interpolator. \textit{Bottom:} (c) Denoise  Module $\mathcal{N}_{d}(\mathbf{x},\bm{\omega_d})$, (d) Illumination Learning Module  $\mathcal{M}_{i}(\mathbf{x},\bm{\omega_i})$, (e) Conditional Feature Modulation (CFM).
		%		 \textit{Right:}  (f) Legengds.
	} \label{fig:workflow}
	\vspace{-3mm}
\end{figure*}% 
\begin{figure}[t!]
	\begin{center}
		\begin{tabular}{c@{\extracolsep{0.35em}}c@{\extracolsep{0.35em}}c@{\extracolsep{0.35em}}c@{\extracolsep{0.35em}}} 
			
			\includegraphics[width=0.232\linewidth]{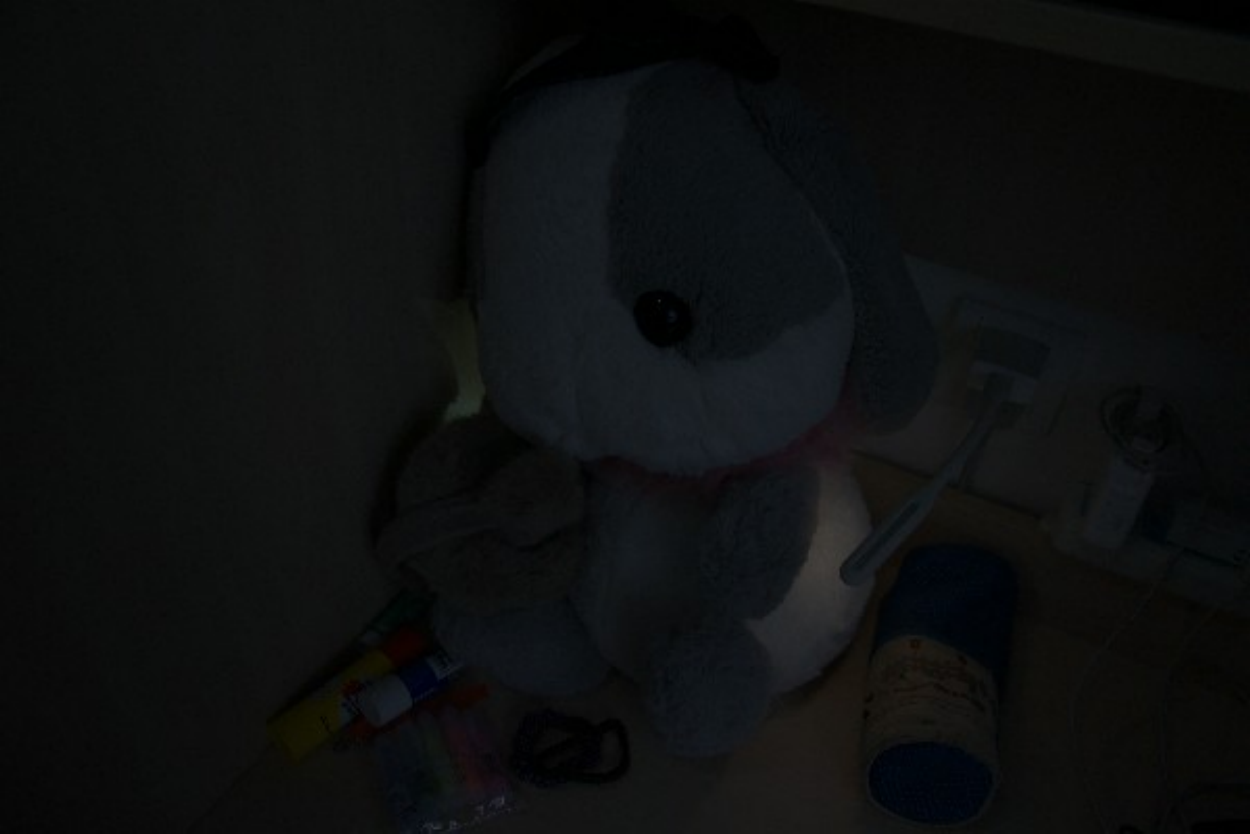} &\includegraphics[width=0.232\linewidth]{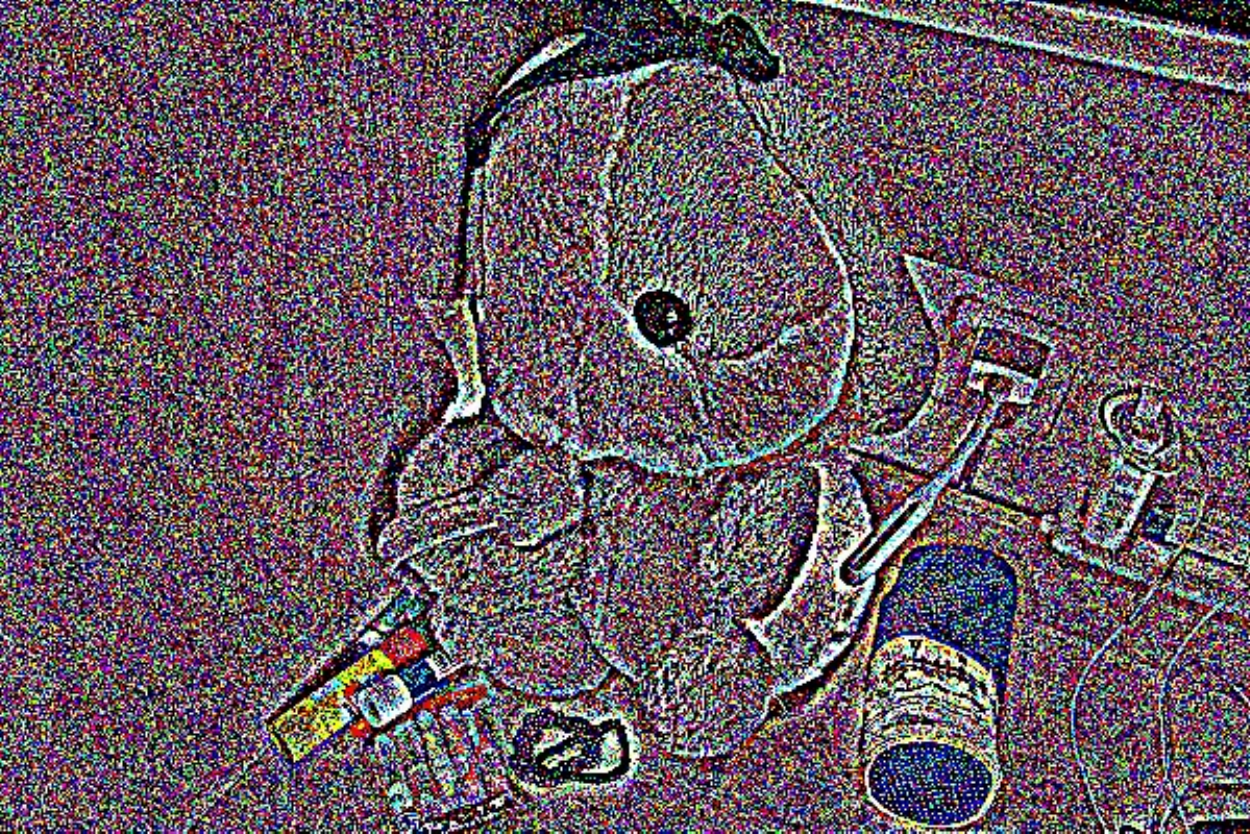} & \includegraphics[width=0.232\linewidth]{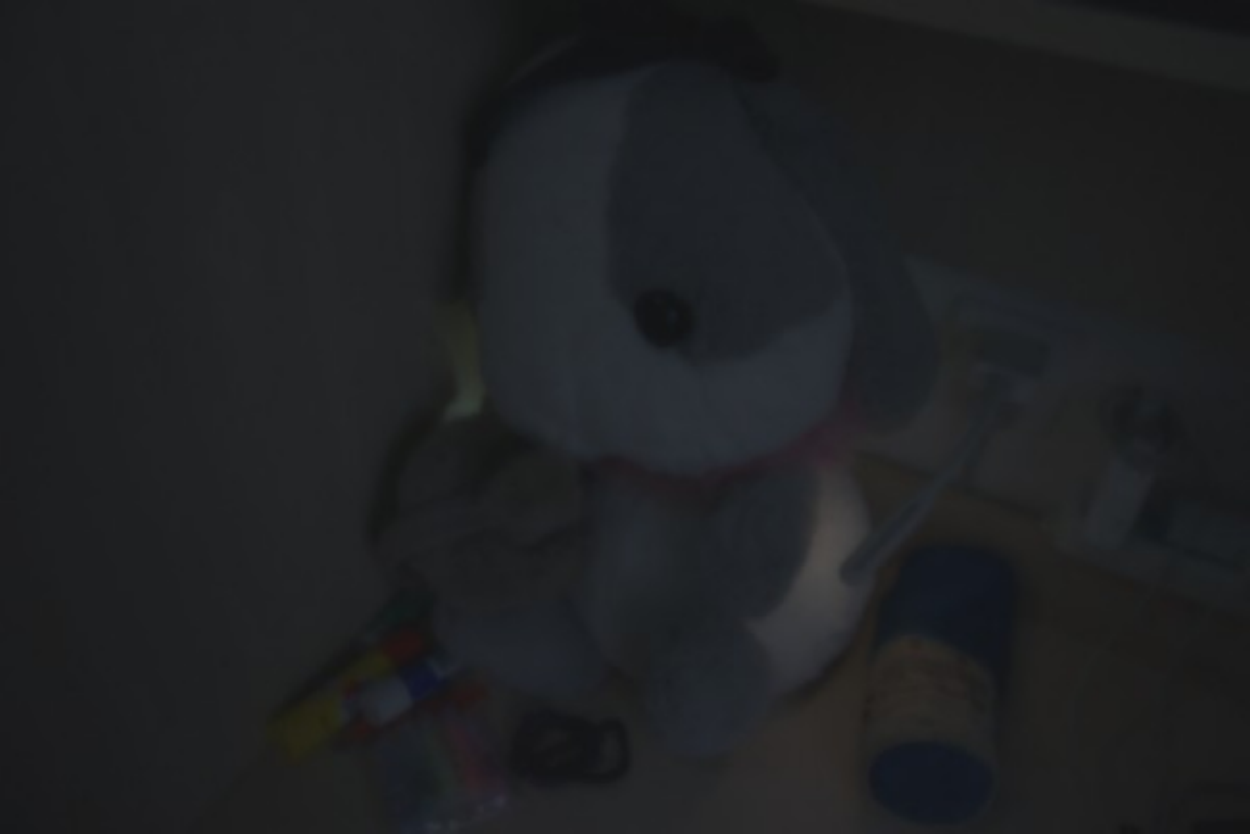} 
			& \includegraphics[width=0.232\linewidth]{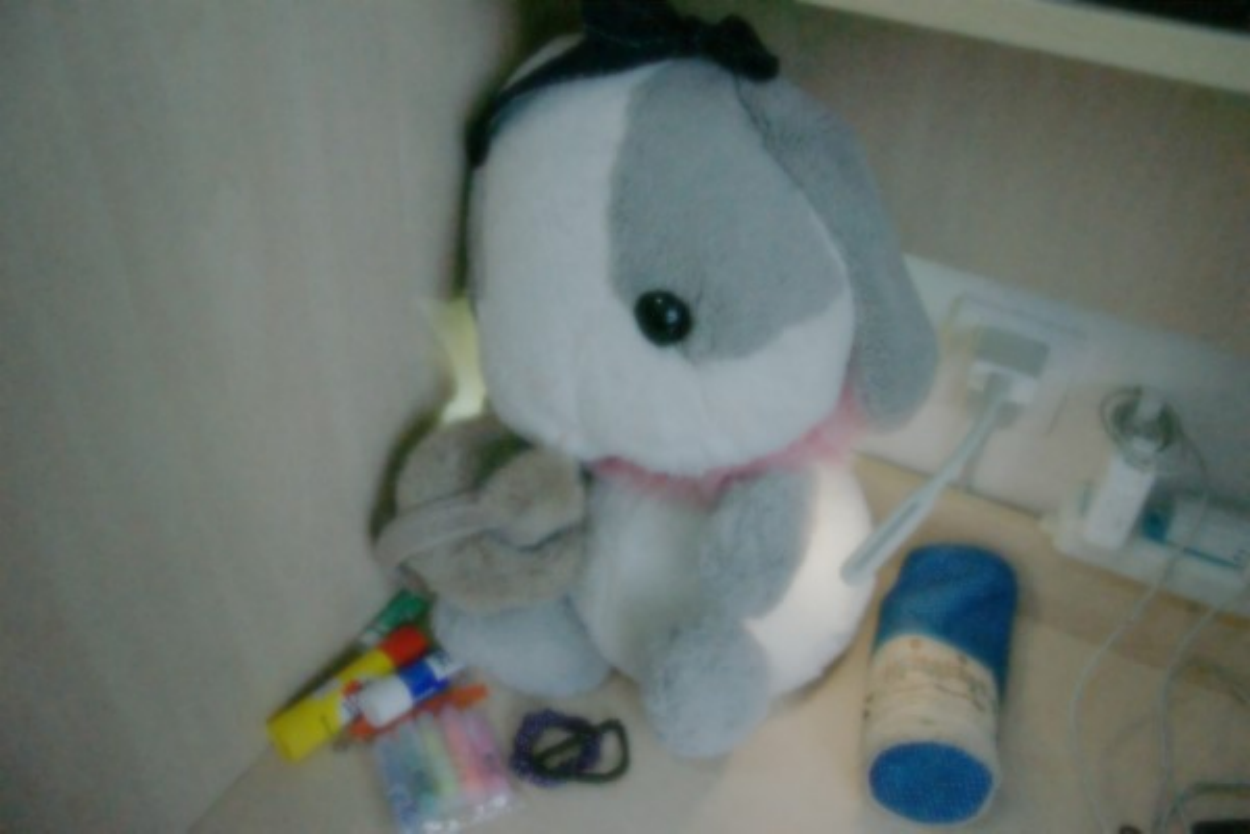}  \\  					
			\specialrule{0em}{-0.2pt}{-0pt} 	 
			Input: $\mathbf{x}$ &  Noise: $\mathbf{v}$ &  Illumination: $\mathbf{y}$ &  Reflectance: $\mathbf{s}$\\		
			\specialrule{0em}{-2pt}{-2pt} 
		\end{tabular}
	\end{center}
	\vspace{-0.2cm}
	\caption{ 
		Visualization of intermediate results (i.e., $\mathbf{x}$, $\mathbf{v}$,  $\mathbf{y}$ and  $\mathbf{s}$) regarding the workflow  based on Equation3~(\ref{eq:scd-to-lii}).
	}
	\label{fig:scd-to-lii}
\end{figure}

\subsection{First-Denoising Last-Enhancing}
Motivated by the Retinex theory~\cite{land1977retinex}, we adapt a more general framework for low-light imaging that accounts for incidental compatible noise. The framework can be formulated as $\mathbf{x}=\mathbf{y} \otimes \mathbf{s}+\mathbf{v}$, where $\mathbf{x}$ denotes the low-light input corrupted by incidental noise $\mathbf{v}$, $\mathbf{s}$ represents the reflectance, $\mathbf{y}$ represents the illumination, and $\otimes$ denotes the element-wise product operation.

%Prior work in simultaneous noise and illumination removal can be broadly classified into two categories: one involves denoising in a separate reflectance branch, while the other performs denoising directly after obtaining the reflectance result. The former involves the aggregation of the reflection and illumination branches, while the latter yields a sharp output. In contrast, our approach offers a new perspective by first removing noise from the low-light image itself, followed by illumination learning. This method adheres to the physical principles of low-light imaging and decouples the tasks of noise removal and illumination learning, distinguishing it from most previous work which considers denoising in the reflectance.

Specifically, we construct an unsupervised learning framework with noise removal followed by Illumination learning, including a precursor Self-Calibrated Denoiser (\textbf{SCD}) and a subsequent Learnable Illumination Interpolator (\textbf{LII}). The learning paradigm can be formulated as follows: 
\begin{equation}\label{eq:scd-to-lii}
	\mathtt{SCD-to-LII}:
	\left\{
	\begin{aligned}
		\mathbf{u} &= \mathbf{x} - \mathbf{v}, \\
		%		~\mathtt{where}~\mathbf{v} = \mathcal{N}_{d}(\mathbf{x};\bm{\omega_d}),\\ 
		\mathbf{s} &= \mathbf{u} \oslash \mathbf{y},
		%		~\mathtt{where}~\mathbf{y} = \mathcal{M}_{i}(\mathbf{u};\bm{\omega_i}), 
		%		\mathbf{s} &= \mathbf{x} \oslash \mathbf{y},~\mathtt{where}~\mathbf{y} = \mathcal{M}_{i}(\mathbf{u};\bm{\omega_i}), 
	\end{aligned}\right. 
\end{equation}
where $\mathbf{v}$ and $\mathbf{u}$ denote the noise map and noiseless low-light map, respectively. For further understanding, a visualization of the intermediate results regarding the workflow is illustrated in Figure~\ref{fig:scd-to-lii}, where $\mathcal{N}_{d}$ and $\mathcal{M}_{i}$ denote the noise removal module and illumination
learning module, parameterized by $\bm{\omega_d}$ and $\bm{\omega_i}$, respectively.

We prioritize denoising primarily due to the following reasons. Firstly, under low-light conditions, it is easier to accurately estimate noise characteristics. On the other hand, brightening the image first can amplify noise intensity, and our noise estimation accuracy diminishes under high-brightness conditions, making it challenging for denoisers. This dual effect makes noise removal more difficult.
In comparison to color restoration, existing methods tend to exhibit relatively pale noise handling capabilities. Therefore, our priority is to focus on image denoising, followed by brightening, to achieve superior visual results.

\subsection{Self-Calibrated Denoiser}
Prior research has demonstrated that estimating noise intensity is beneficial for denoising. In the following, we attempt to explore the noise intensity distribution of low-light input from its intrinsic properties (e.g., gradient information), and develop a self-calibrated denoiser to bootstrap the image to remove the intrinsic noise in an unsupervised manner. 

\subsubsection{Noise Estimation}
The gradient characteristics of image can reflect the distribution of potential noise. Thus, we use the higher order gradient of the image to estimate the noise intensity. After a rigorous theoretical derivation the following equation is given
% \footnote{Further details and comprehensive proof are provided in the \ref{app: proof}.} 
:
\begin{equation}\label{eq:scd1} 
	\sigma \approx \psi(\nabla^{n}\mathbf{x}):= \frac{\sqrt{\pi}E|\nabla^{n}\mathbf{x}|}{\sqrt{2\Sigma_{k=0}^{n} [C_{n}^{k}]^2}}.
\end{equation}	
%\noindent Here, $\nabla^{n} \mathbf{x}$ represents the $n$-order image gradient. Given an image with pixel coordinate position $p$, the gradient can be computed as $\nabla\mathbf{x}(p)= \mathbf{x}(p)-\mathbf{x}(p+1)$, and for $n>1$, it can be calculated recursively using $\nabla^{n} \mathbf{x}(p)=\nabla^{n-1} \mathbf{x}(p)-\nabla^{n-1} \mathbf{x}(p+1)$.
%And $C$ represents the mathematical notation for combinations, i.e. $C_{n}^{k} = \frac{n!}{k!(n - k)!}$. The notation $E$ represents the expectation.
In this equation, $\nabla^{n} \mathbf{x}$ signifies the $n$-order image gradient. For a given image with pixel coordinate position $p$, the gradient can be calculated as $\nabla\mathbf{x}(p)= \mathbf{x}(p)-\mathbf{x}(p+1)$. When $n$ is greater than 1, the computation can be recursively performed using $\nabla^{n} \mathbf{x}(p)=\nabla^{n-1} \mathbf{x}(p)-\nabla^{n-1} \mathbf{x}(p+1)$.
The symbol $C$ represents the mathematical notation for combinations, defined as $C_{n}^{k} = \frac{n!}{k!(n - k)!}$. Here, $E$ signifies the expectation.

\textbf{Proof:}
Here we initially consider the Gaussian noise distribution
%\footnote{The wider range of realistic complex noise will be deemed for future work.} 
$\mathbf{v} \sim \mathcal{N}(0, \sigma^2)$ with a standard deviation of $\sigma$, and define the following relationship: $\mathbf{x}=\mathbf{u}+\mathbf{v}$ between the noisy input $\mathbf{x}$ and noise-free image $\mathbf{u}$. 	 
First, we define the first-order gradient and $n$-order gradient ($n>1$) of the image at pixel coordinate position $p$, i.e., $\nabla \mathbf{x}(p) = \mathbf{x}(p) -  \mathbf{x}(p + 1)$ and $\nabla^{n}\mathbf{x}(p) = \nabla^{n-1}\mathbf{x}(p) - \nabla^{n-1}\mathbf{x}(p + 1)$. 
Decoupling the noise map $\mathbf{v}$   and clear images $\mathbf{u}$, we obtain $\nabla^{n}\mathbf{x}(p) = \nabla^{n}\mathbf{u}(p) + \nabla^{n-1}\mathbf{v}(p) - \nabla^{n-1}\mathbf{v}(p + 1)$.  Recursively, we further obtain 	  
%	 \begin{equation*} \small
	%	 	\begin{aligned}
		%	 		\nabla^{n}\mathbf{x}(p) & = \nabla^{n}\mathbf{u}(p) + C_{2}^{0} \nabla^{n-2}\mathbf{v}(p) - C_{2}^{1} \nabla^{n-2}\mathbf{v}(p + 1) + C_{2}^{2} \nabla^{n-2}\mathbf{v}(p + 2), \\
		%	 		%& = ... \\
		%	 		& = ... = \nabla^{n} \mathbf{u}(p) +  \Sigma_{k=0}^{n-1}(-1)^{k} C_{n}^{k}\nabla \mathbf{v}(p+k), \\
		%	 		& = \nabla^{n} \mathbf{u}(p) +  \Sigma_{k=0}^{n}(-1)^{k} C_{n}^{k} \mathbf{v}(p+k).
		%	 	\end{aligned}
	%	 \end{equation*}
$\nabla^{n}\mathbf{x}(p)=\nabla^{n} \mathbf{u}(p) +  T$, where $T=\Sigma_{k=0}^{n}(-1)^{k} C_{n}^{k} \mathbf{v}(p+k).$  Since each noisy pixel is independently and identically distributed,  i.e., $\mathbf{v}(p) \sim \mathcal{N}(0, \sigma^2)$, we have $T \sim \mathcal{N}(0,\Sigma_{k=0}^{n}[C_{n}^{k}]^2\sigma^2)$ and $E|T| = \sigma\sqrt{\frac{2}{\pi}\Sigma_{k=0}^{n}[C_{n}^{k}]^2}$, where $E$ denotes the expectation sign.  

Further, according to $
E|\nabla^{n}\mathbf{x}| = E|\nabla^{n}\mathbf{u} + T|$, we can obtain the following inequality: $
E|T| - E|\nabla^{n}\mathbf{u}| \leq  E|\nabla^{n}\mathbf{x}| \leq E|T| + E|\nabla^{n}\mathbf{u}|.$ 
Since the natural noise-free image satisfies the smoothness property, the following rules are satisfied $\nabla \mathbf{u}(p_s)\rightarrow 0 $ and $	\nabla \mathbf{u}(p_e)\rightarrow \mathbf{u}(p_e)$ for the smooth region pixel point $p_s$ and the edge pixel point $p_e$. Further we obtain $E|\nabla^n \mathbf{u}| \approx E|\nabla \mathbf{u}|$.  Substituting into the inequality, we have $E|T| - E|\nabla \mathbf{u}| \leq  E|\nabla^{n}\mathbf{x}| \leq E|T| + E|\nabla \mathbf{u}|$. When $n$ is large enough, $E|\nabla^n \mathbf{u}| \approx E|\nabla \mathbf{u}| = const \ll \sqrt{\Sigma_{k=0}^{n}[C_{n}^{k}]^2}$, $E|\nabla^n \mathbf{u}| \ll E|T|$,  it follows from the pinch-force theorem that $
E|\nabla^{n}\mathbf{x}| \approx E|T| = \sigma\sqrt{\frac{2}{\pi}\Sigma_{k=0}^{n}[C_{n}^{k}]^2}.$ Therefore, the Gaussian noise intensity can be estimated by statistical analysis of $|\nabla^{n}\mathbf{x}|$, which yields $\sigma \approx \psi(\nabla^{n}\mathbf{x}):= \frac{\sqrt{\pi}E|\nabla^{n}\mathbf{x}|}{\sqrt{2\Sigma_{k=0}^{n} [C_{n}^{k}]^2}}$.

\textbf{Time Complexity.}
%At the heart of our method lies the accurate derivation of noise standard deviation through the utilization of 1st or higher-order gradients extracted from images. 
Due to the simple but efficient representation, the noise estimation approach can reach linear time complexity.
By capitalizing on statistical information encompassing the entire image, our approach eliminates the necessity of patch selection based on additional principles. This fundamental characteristic allows our method to operate with a time complexity of $\mathbf{O}(r)$ for $r$-dimensional flattened image input, affirming its linear computational demand. This inherent linearity empowers our method with exceptional suitability for real-time computing applications.
Notably, our approach simplifies the noise calculation process by relying on Equation (\ref{eq:scd1}), foregoing the need for additional complex algorithms. This simplicity not only ensures accuracy but also makes our method highly compatible with GPU implementations. Consequently, our method enables accelerated inference without compromising precision or computational efficiency.

\subsubsection{Noise Remover}
Based on the rapid noise estimation as an additional input, we've developed a Self-Calibrated Denoiser (SCD), represented in (c) of Figure~\ref{fig:workflow}, formulated as:
\begin{equation}\label{eq:scd2}
	\mathcal{N}_{d}(\mathbf{x};\bm{\omega_d}): 
	\left\{
	\begin{aligned}
		%f_{\bm{\sigma}}^{latent} 
		\bm{f_{\sigma}} &= {\mathcal{F}}_{CFT}\left({\mathcal{F}}_{conv}(\sigma)\right),\\
		\mathbf{v} &= \mathcal{G}(\sigma, \sigma_{g}) \otimes {\mathcal{F}}_{Unet}\left({\mathcal{F}}_{C}\big(\bm{f_{\sigma}}, \mathbf{x} \big)\right), 
	\end{aligned}\right.
\end{equation}
Here, the noise level $\bm{\sigma}$ is estimated using Equation~(\ref{eq:scd1}). This estimated noise level is embedded within the conditional feature transformation layer ${\mathcal{F}}_{CFT}$, yielding the latent feature map $\bm{f{\sigma}}$ detailed in (e) of Figure~\ref{fig:workflow}. Subsequently, the cascaded raw input $\mathbf{x}$ is directed into the Unet-style based denoiser ${\mathcal{F}}_{Unet}$ alongside a noise gate function $\mathcal{G}$ to generate the noise map $\mathbf{v}$. The function $\mathcal{G}(\sigma, \sigma{g}) = \text{sign}(\text{ReLU}(\sigma - \sigma_{g}))$ regulates the process, preserving crucial details and preventing excessive smoothing, particularly in noiseless images.
%We begin the pretraining of the SCD module by employing techniques aligned with the DDPM approach. This involves the introduction of Gaussian noise with varying intensities into clear low-light images, followed by attempts in pre-training to remove this additional noise.

\subsection{Learnable Illumination Interpolator}
To learn better illumination map,  we generalize the following uniform prior knowledge. It reflects the mathematical rules about the original input, illumination and reflection that need to be satisfied in a limited dynamic range.

Here, we aim to examine the numerical relationships among each component to unveil their intrinsic connections. By imposing normalization operation restrictions on the raw input, we ensure that the reflectance and illumination components, confined within a limited dynamic range during recovery, adhere to the following prior constraints: $0\leq \mathbf{x} \leq \mathbf{s} \leq\mathbf{1}$, and $0\leq \mathbf{x} \leq \mathbf{y} \leq\mathbf{1}$, where $\mathbf{1}$ represents a unit vector.
Considering the potential structural consistency among $\mathbf{x}$, $\mathbf{s}$, and $\mathbf{y}$, the challenge of separating illumination from reflectance within an image, inspired by the principles of retinex theory, can be addressed by formulating an explicit interpolation operator or function. This construction facilitates the effective separation of these components based on their intrinsic relationships.

\begin{figure}
	\centering
	\includegraphics[width=0.8\linewidth]{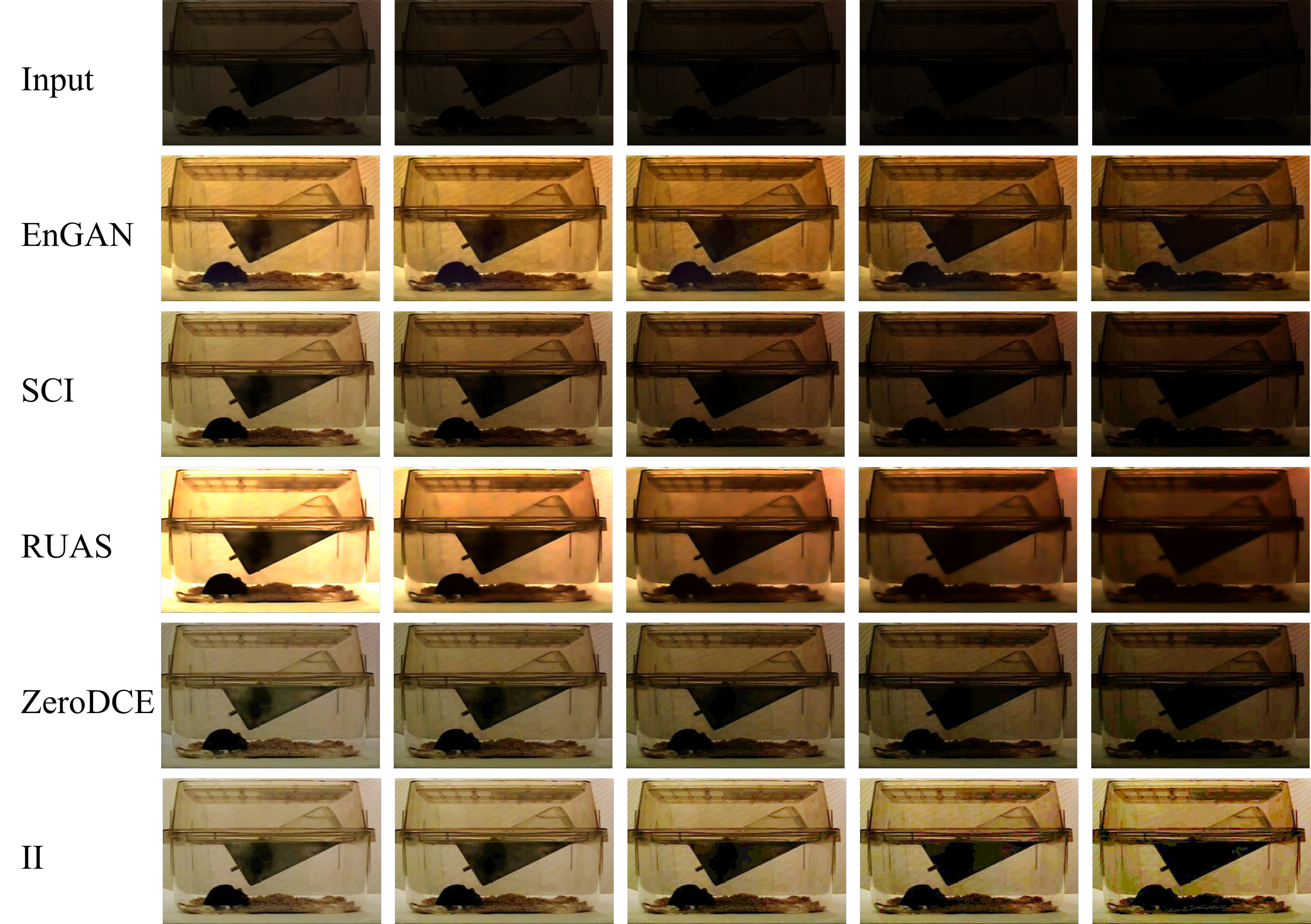}
	\caption{Compared with state-of-the-art unsupervised methods, II, the Illumination Interpolator, has ability to enhance images with different degradation factors, which simply divide the linear interpolation result between the input image and a unit vector.}
	\label{fig: inter_res}
\end{figure}

Building upon the established relationship, we propose a straightforward method to estimate illumination $\mathbf{y}$ by employing a linear interpolation function, denoted as $\mathbf{y}=g(\mathbf{x})\otimes \mathbf{x} +(\mathbf{1}-g(\mathbf{x}))$. Notably, our experimental observations highlight a notable enhancement effect achievable by formulating the interpolation function $g(\mathbf{x})$ in relation to the image mean $\mathtt{Mean}$, specifically as $g(\mathbf{x})=\mathbf{1}-\mathtt{Mean}(\mathbf{x})$. This observation is evidenced by the results obtained from the Illumination Interpolator(II) in Figure \ref{fig: inter_res}. \label{img_mean_mean}
Intuitively, brighter areas tend to have an illumination value closer to 1, while darker areas retain illumination values closer to their original intensities. Consequently, upon division, as described in Equation (\ref{eq:scd-to-lii}), the resultant image is brighter than the original. The image mean $\mathtt{Mean}$ encapsulates statistical information from the images and can be adapted to accommodate various images under different lighting conditions.
Although interpolation exhibits potential for enhancement, it remains fixed and may not consistently align with our desired outcomes.

To further improve the representational capability and the nonlinear nature of the function of $g$, 
we construct learnable network module $\mathcal{M}_{i}(\mathbf{u},\bm{\omega_{i}})$ (parameterized by $\bm{\omega_{i}}$) to simulate the process of illumination generation in the form of interpolation. Specifically, the noise-free low-light  image $\mathbf{u}$ is transmitted to the network to generate illumination factors and then illumination is calculated in a weighted form.  The above illumination learning process is formulated as
%\begin{equation}\label{eq:LII3} 
%	\mathbf{y}=\mathbf{u} \otimes \underbrace{\mathcal{M}_{i}(\mathbf{u},\bm{\omega_{i}})}_{\bm{\alpha}} ~+~ \mathbf{e} \otimes \underbrace{\mathcal{M}_{i}(\mathbf{u},\bm{\omega_{i}}}_{\bm{\beta}}),
%\end{equation}
\begin{equation}\label{eq:LII3} 
	\mathbf{y}=\mathbf{u} \otimes \bm{\alpha}~+~ \mathbf{1} - \bm{\alpha},~\mathtt{where}~\bm{\alpha} = \mathcal{M}_{i}(\mathbf{u},\bm{\omega_{i}}),
\end{equation}
where $\mathbf{1}$ denotes the unit vector and $\bm{\alpha}$ is the learned illumination interpolation factor. As shown in (d) of Figure~\ref{fig:workflow}, $\mathcal{M}_{i}(\mathbf{u},\bm{\omega_{i}})$ is constructed as a network module.
%The interpolation is a natural structure constraint on fidelity and smooth properties of illumination.
The interpolation is a natural structure constraint on fidelity and smooth properties of illumination and we will delve into the compelling advantages of LII with such structure constraint in Section Discussion~\ref{CharmOfLLI}.

\subsection{Reference-Free Loss Function}  
The total loss function is expressed as 
\begin{equation}\label{eq:loss}
	\mathcal{L}_{total}={\lambda}_{srr}*\mathcal{L}_{srr}+{\lambda}_{nr}*\mathcal{L}_{nr},
\end{equation}
where $\mathcal{L}_{srr}$ and $\mathcal{L}_{nr}$ are self-regularized recovery loss, and noise removal loss respectively. ${\lambda}_{srr}$ and ${\lambda}_{nr}$ are the corresponding weights. 
%		In our training, we empirically set %${\lambda}_{srr}$=1 and ${\lambda}_{nr}$=0.1.
%		${\lambda}_{srr}={\lambda}_{nr}$=1.

%	\noindent
\textbf{Self-Regularized Recovery Loss.} Inspired by the gray world assumption and following normal image's pixel  intensity distribution, we propose the self-regularized recovery loss $\mathcal{L}_{srr}$ to encourage color naturalness of the result and it can be formulated as
\begin{equation}\label{eq:loss1}
	\mathcal{L}_{srr}(\mathbf{s})= e^{relu(|\bar{\mathbf{s}} - \eta_t| - \sigma_t)}, 
\end{equation}  
where $\bar{\mathbf{s}}$ is the mean value of image $\mathbf{s}$ in different channels, $\eta_t$ and $\sigma_t$ are constants, and presents the mean and standard deviation value of the desired enhanced image. $\eta_t$ is set to $[0.485, 0.456, 0.406]$ and $\sigma_t$ is set to $[0.229, 0.224, 0.225]$ according to statistical properties of \textit{Imagenet} \cite{deng2009imagenet}, revealing a statistic value of image intensity. These values represent common attributes of typical images and hold practical value, being widely applied in preprocessing processes like detection and segmentation.
Further discussion is provided in Section Discussion~\ref{CharmOfSrr}.
%			when images normalized into range $[0, 1]$. 
%The color loss ensures that the brightness of each channel is constrained within a small margin, similar to normal images. If the brightness is at a normal level, the loss will be zero, but if it is over-exposed or under-exposed, the loss will increase rapidly. This approach allows us to effectively enhance low light images without over-exposure or under-exposure.
%The color loss allows us to optimize parameters effectively without the need for complex, prior-based domain knowledge. Unlike other methods that require adjusting hyper parameters between different loss terms, our novel illumination architecture enables us to use only the color loss, simplifying the training process.

%	\noindent
\textbf{Self-Adaptable Noise Removal Loss.} To preserve edge details during the denoising process,  we introduce an adaptive denoising loss function that dynamically balances the weights of fidelity term and regularization term~\cite{rudin1992nonlinear}, thereby ensuring optimal performance across varying noise intensities encountered in unsupervised denoising. It can be formulate as
\begin{equation}\label{eq:loss2}
	\mathcal{L}_{nr}(\mathbf{y})= ||\mathbf{u}-\mathbf{x}||^2 + \lambda_{TV}  * \sigma(\mathbf{x}) * \mathtt{TV}(\mathbf{u}), 
\end{equation}  
in which the  regularized term $\mathtt{TV}$ is the standard total variation with a hyper- parameter $\lambda_{TV}>0$, and the noise coefficient $\sigma(\mathbf{x})$ is the noise aware weight for balancing the smoothness and details of the input $\mathbf{x}$, estimated by the Equation (\ref{eq:scd1}).

\subsection{Discussion}  

%		\noindent

\begin{figure}[H]
	\begin{center}
		\includegraphics
		[width=0.96\linewidth]
		{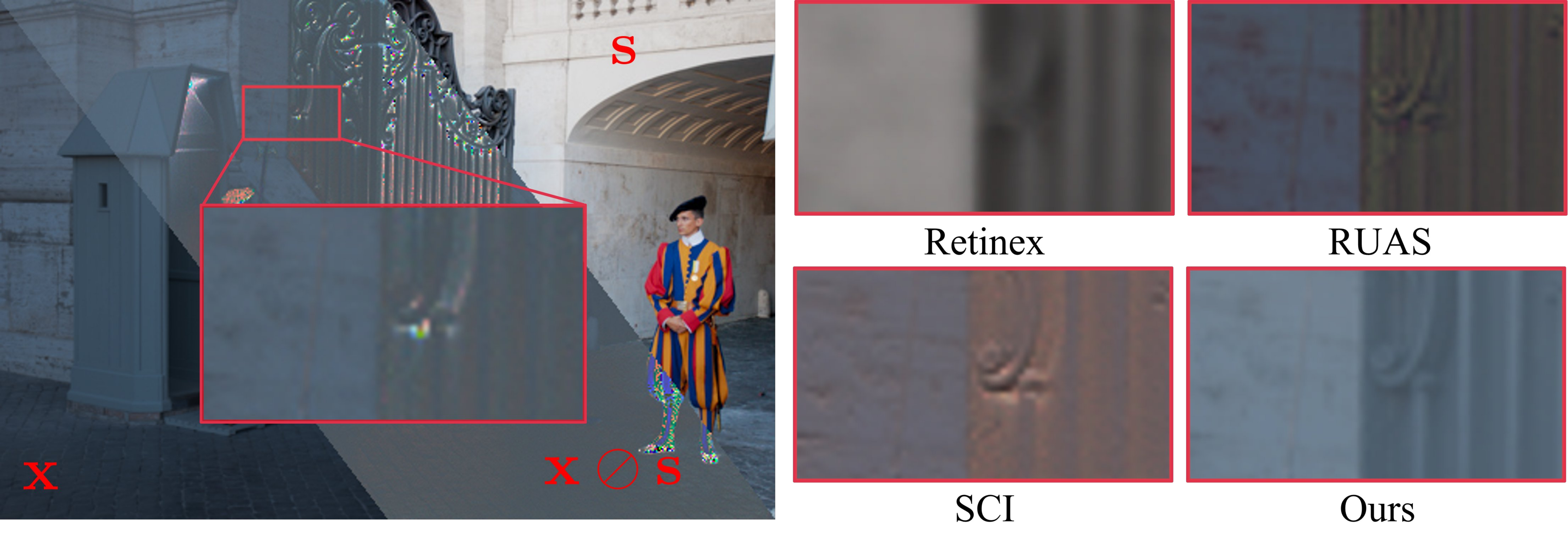}
	\end{center}
	\vspace{-0.2cm}
	\caption{ 
		Visual comparison of estimated illumination map.	\textit{Left:} Combination reference produced by the ground truth, \textit{Right:} Illumination estimated via reverse procedure (i.e., $\mathbf{x} \oslash \mathbf{s}$).
		%Pixel value distribution of randomly sampled column (yellow box). \textit{Middle:}, 
	}
	\label{fig:lii}\vspace{-0.2cm}
\end{figure}

\textbf{Charm of Illumination Interpolator.} ~\label{CharmOfLLI} Compared to Retinex-induced LLIE methods, the proposed illumination interpolator learns a difference function from the original image to the illumination map. 
%With its advanced interpolation structure, the LII naturally satisfies the assumptions for illumination smoothing and structural fidelity.  
The interpolation between input and unit vector effectively performs edge-preserving smoothing on the input image. 
On one hand, the differences between pixels decrease after interpolation, resulting in an overall smoother appearance. 
On the other hand, the interpolation process does not disturb the relative spatial relationships between pixels, thus preserving edges and ensuring the overall structure of the image remains stable.
As a result, the output maintains consistency in relative brightness and structural fidelity with the input. 
That's to say, the output of LII naturally satisfies the general assumptions for illumination smoothing and structural fidelity, which are the main
Note that comparison results of the Retinex-induced LLIE methods is illustrated in Figure~\ref{fig:lii}. Considering the Retinex-induced model, we calculate the illumination map $\mathbf{y}$ of various methods by using $\mathbf{x} \oslash \mathbf{s}$. Obviously, the illumination map estimated by ours is smoother and retains the edge detail information of the original image.  

%			LII provides a natural way to ensure the fidelity and smooth assumption of illumination and our loss draws a desired blueprint of the final result's properties. Although without constrained directly on illumination map as other retinex-based methods do, our method outperforms at least as well as other methods. To further demonstrate this feature, considering the most common loss function $L = \lambda_1 \Vert \mathbf{x} - \mathbf{y} \Vert^2 + \lambda_2 \Vert \nabla \mathbf{y} \Vert$ for illumination maps,  we use the common constraint term $\Vert \nabla \mathbf{y} \Vert$ and the translation fidelity term $\Vert \hat{\mathbf{x}} -\hat{\mathbf{y}} \Vert^2$ for low-light images as measurements to quantitatively analyze  illumination maps, where
%			$\hat{\mathbf{x}} = \mathbf{x} - \mathtt{Mean}(\mathbf{x})$ and $\hat{\mathbf{y}} = \mathbf{y} - \mathtt{Mean}(\mathbf{y})$.
%			To avoid the interference of noise, we conduct quantitative evaluation experiment on MIT dataset.
%			As shown in Figure \ref{fig:charm}, illumination maps estimated by our method are smoother and more similar to the dark input in biased structure.

\begin{figure}[H]
	\vspace{-0.2cm}
	\begin{center}
		\begin{tabular}{c@{\extracolsep{0.35em}}c@{\extracolsep{0.35em}}} 			
			
			\includegraphics[width=0.46\linewidth, height=0.345\linewidth]{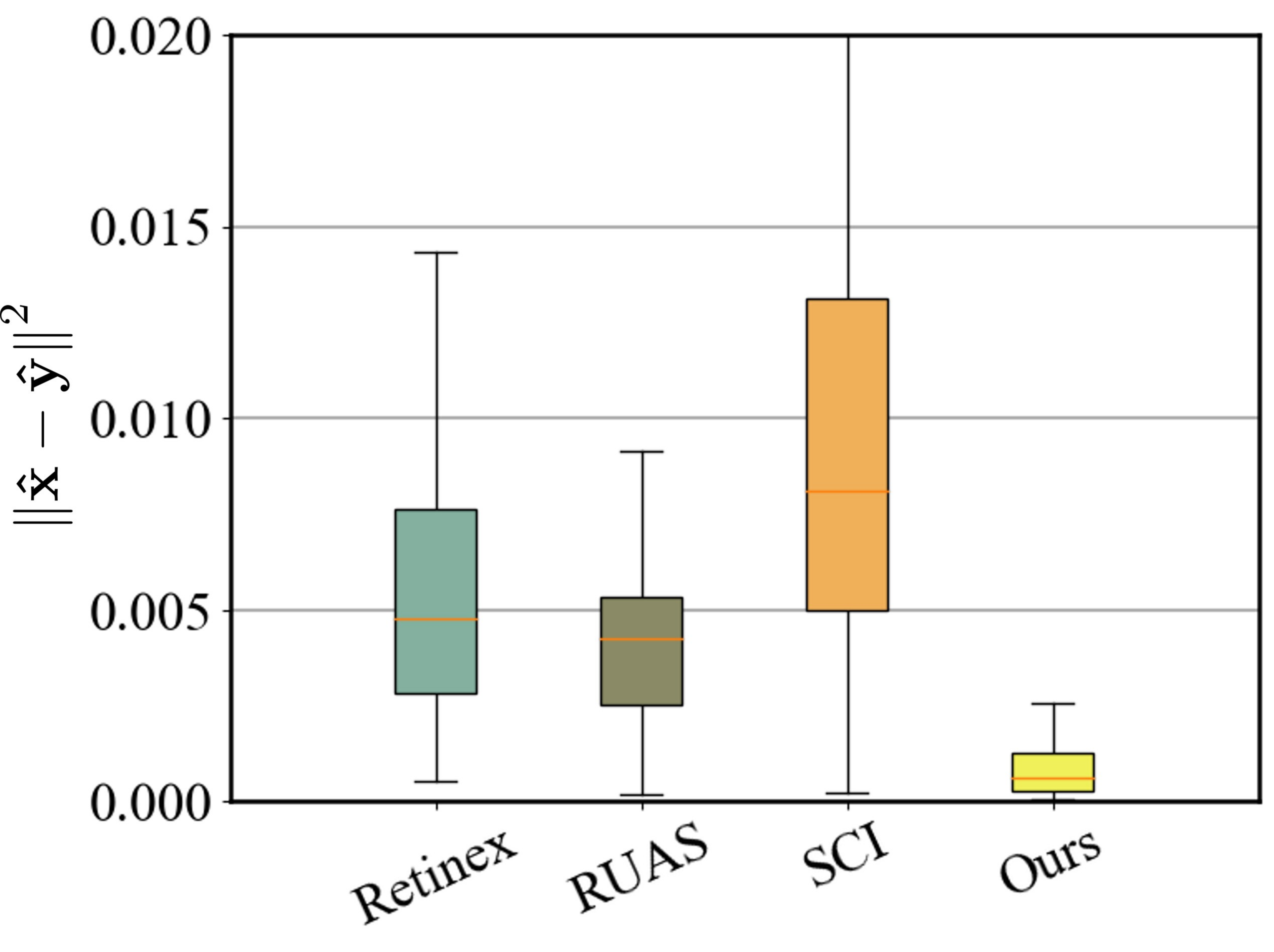} & \includegraphics[width=0.46\linewidth, height=0.345\linewidth]{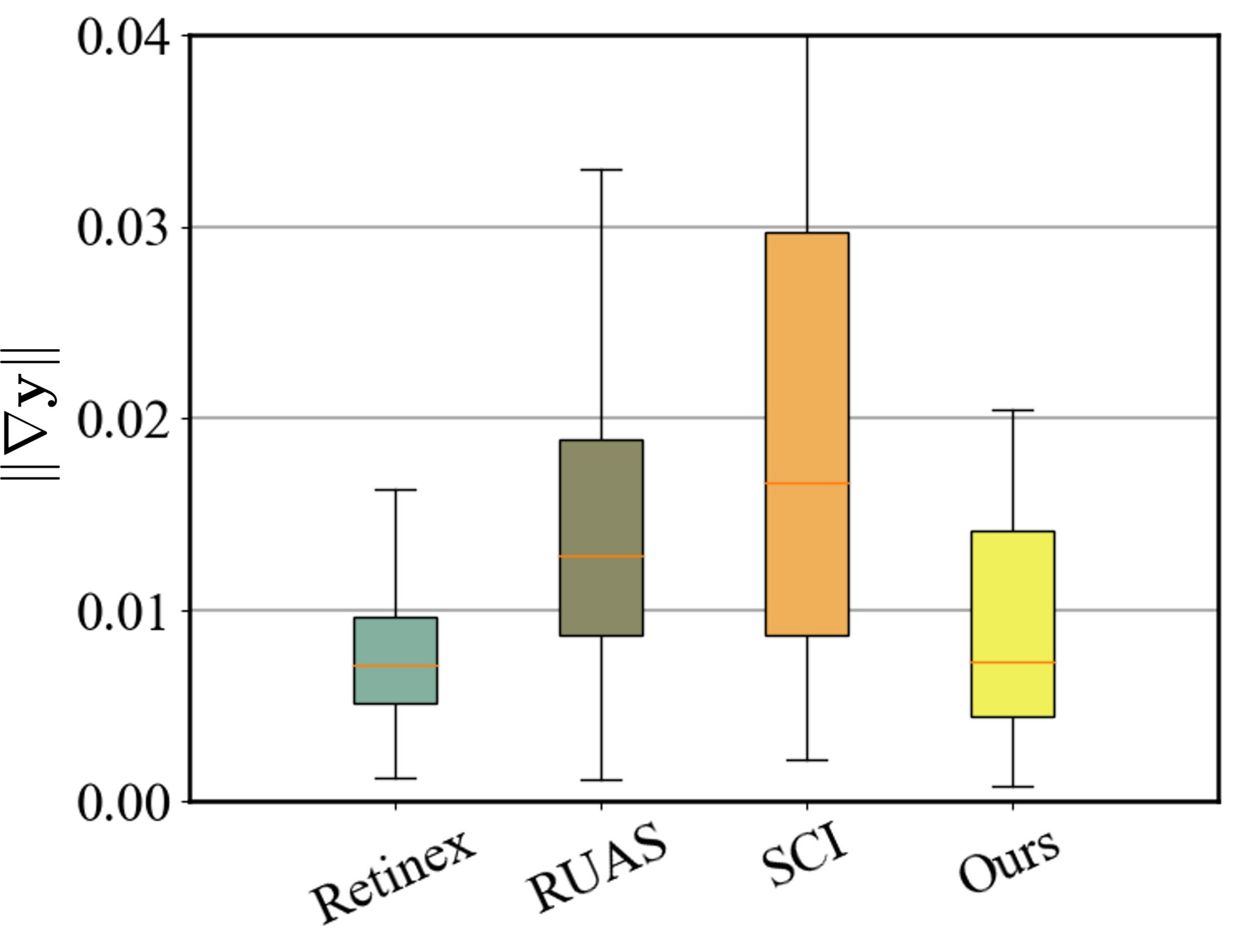}  \\
		\end{tabular}
	\end{center}
	\vspace{-0.05cm}
	\caption{ 
		%			Charm of combination training. 
		Statistic results of estimated illumination map.
		Our illumination maps' distribution outperforms at least as well as other retinex-based methods, despite without using corresponding loss explicitly.
	}
	\label{fig:charm_fid}
	\vspace{-0.15cm}
\end{figure}

LII provides a natural way to ensure the fidelity and smooth assumption of illumination and our loss draws a desired blueprint of the final result's properties. 
For a certain factor $\bm{\alpha}$, the gradient of illumination map can be calculated by $\nabla \bm{y} = \bm{\alpha} \nabla\bm{x}$, according to Equation~(\ref{eq:LII3}).
Although without constrained directly on illumination map as other retinex-based methods do, our method outperforms at least as well as other retinex-based approaches. We quantitatively analyze illumination maps by measuring the common constraint term $\Vert \nabla \mathbf{y} \Vert$ and the translated fidelity term $\Vert \hat{\mathbf{x}} -\hat{\mathbf{y}} \Vert^2$, where $\hat{\mathbf{x}} = \mathbf{x} - \mathtt{Mean}(\mathbf{x})$ and $\hat{\mathbf{y}} = \mathbf{y} - \mathtt{Mean}(\mathbf{y})$. We use the MIT dataset for our experiment to avoid interference from noise. Results in Figure \ref{fig:charm_fid} show that our method produces smoother and more similar illumination maps to the low-light input.

\begin{figure}[htb]
	\vspace{-0.2cm}
	\begin{center}		
		\includegraphics[width=0.96\linewidth]{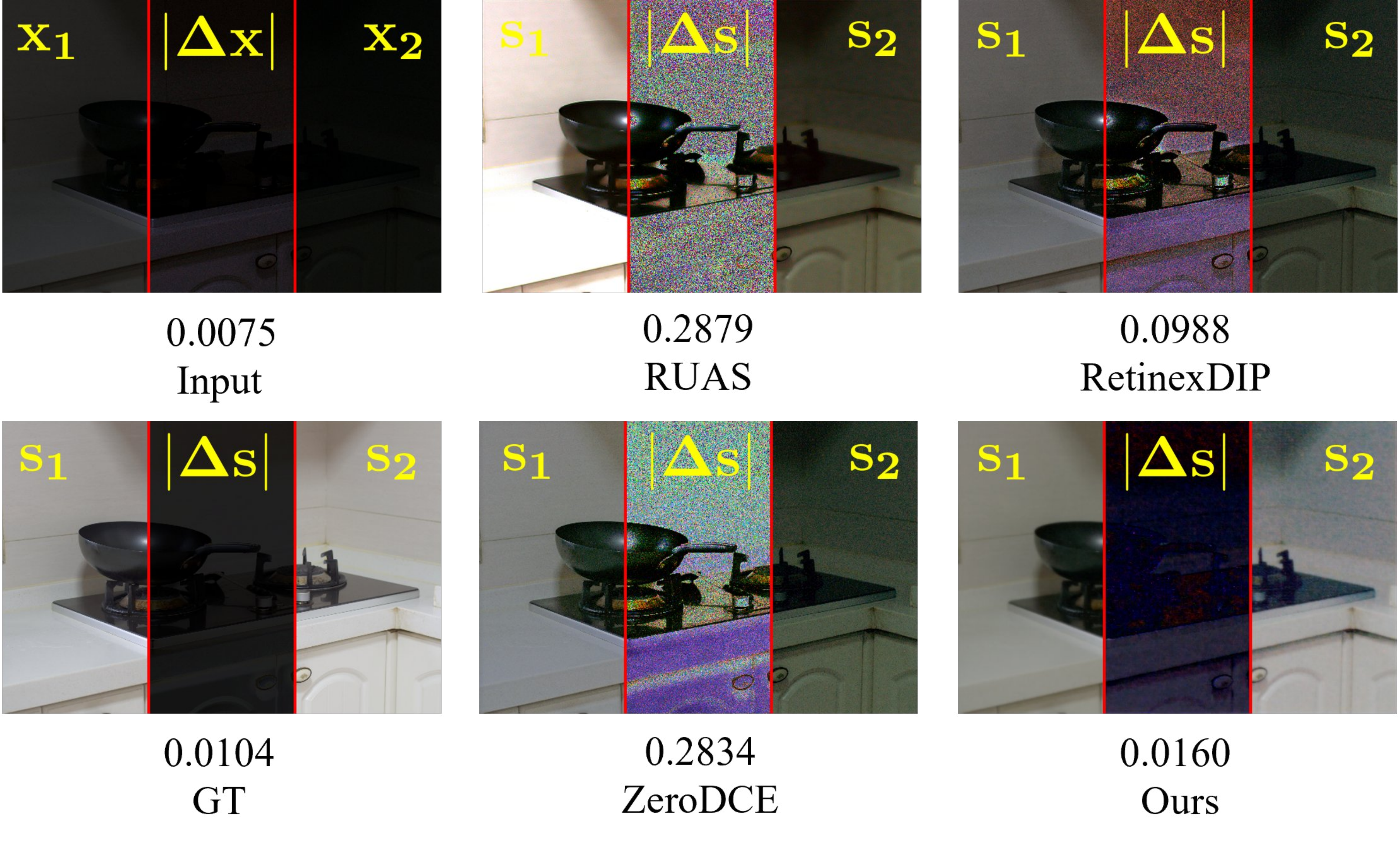} 
	\end{center}
	%	\vspace{-0.3cm}
	\caption{ 
		Charm of self-regularized recovery loss for robust enhancement. 
		$\mathbf{s_i}$ presents the enhanced result of $\mathbf{x_i}$ and $\mathbf{\Delta s} = 10(\mathbf{s_{1} - s_{2}})^2$ means scaled square difference between results for better comparison (darker is better). Mean square error are listed at the bottom to quantify the difference of similar inputs/results (smaller is better). 
	}
	\label{fig:charm_of_color_loss}
	%	\vspace{-0.1cm}
\end{figure}

%		\noindent
\textbf{Charm of Self-Regularized Recovery Loss.} \label{CharmOfSrr}
The loss ensures that the brightness of each channel is constrained within a small margin, similar to normal images. If the brightness is at a normal level, the loss will be zero, but if it is over-exposed or under-exposed, the loss will increase rapidly. 
From the perspective of image manifold, this approach considers mapping the high-dimensional information of the image into a low-dimensional manifold region that reflects the color distribution of the image, allowing us to effectively enhance low-light images without over-exposure or under-exposure.
Additionally, the color loss function permits the efficient optimization of parameters without the need for complex, prior-based domain knowledge, with LLI's linear convex interpolation structure. Compared to other methods that require adjusting hyper parameters between distinct loss terms, our novel illumination architecture utilizes only the color loss function, thus simplifying the training process.
Moreover, when handling low-light images with subtle differences, our proposed method generates results that exhibit smaller differences compared to other approaches. This outcome further validates the robustness, as shown in Figure~\ref{fig:charm_of_color_loss}.

\section{Experiment}

In this section, we initially showcase the effectiveness and efficiency of our noise estimation method within low-light conditions and showcase its positive impact on denoising in such scenarios with dynamic noise.
Then, we conducted comprehensive evaluations encompassing both qualitative and quantitative analyses using well-established low-light datasets. 
%Additionally, to substantiate the reliability and applicability of our method in practical settings, we performed a series of experiments in real-world scenarios. These experiments were meticulously designed to validate the resilience and adaptability of our approach across various real-world conditions.
Moreover, to provide deeper insights into our method's components, we executed an extensive set of ablation experiments. These experiments aimed to elucidate the functionalities of each module within our approach and to compare the effectiveness of our modules against specific components extracted from state-of-the-art methods.

\subsection{Noise Estimation Evaluation}

\begin{figure}[htbp]
	\centering
	\begin{center}
		\begin{tabular*}{\linewidth}{c@{\extracolsep{0.5em}}c@{\extracolsep{0.35em}}} 			
			
			\includegraphics[width=0.48\linewidth, height=0.39\linewidth]{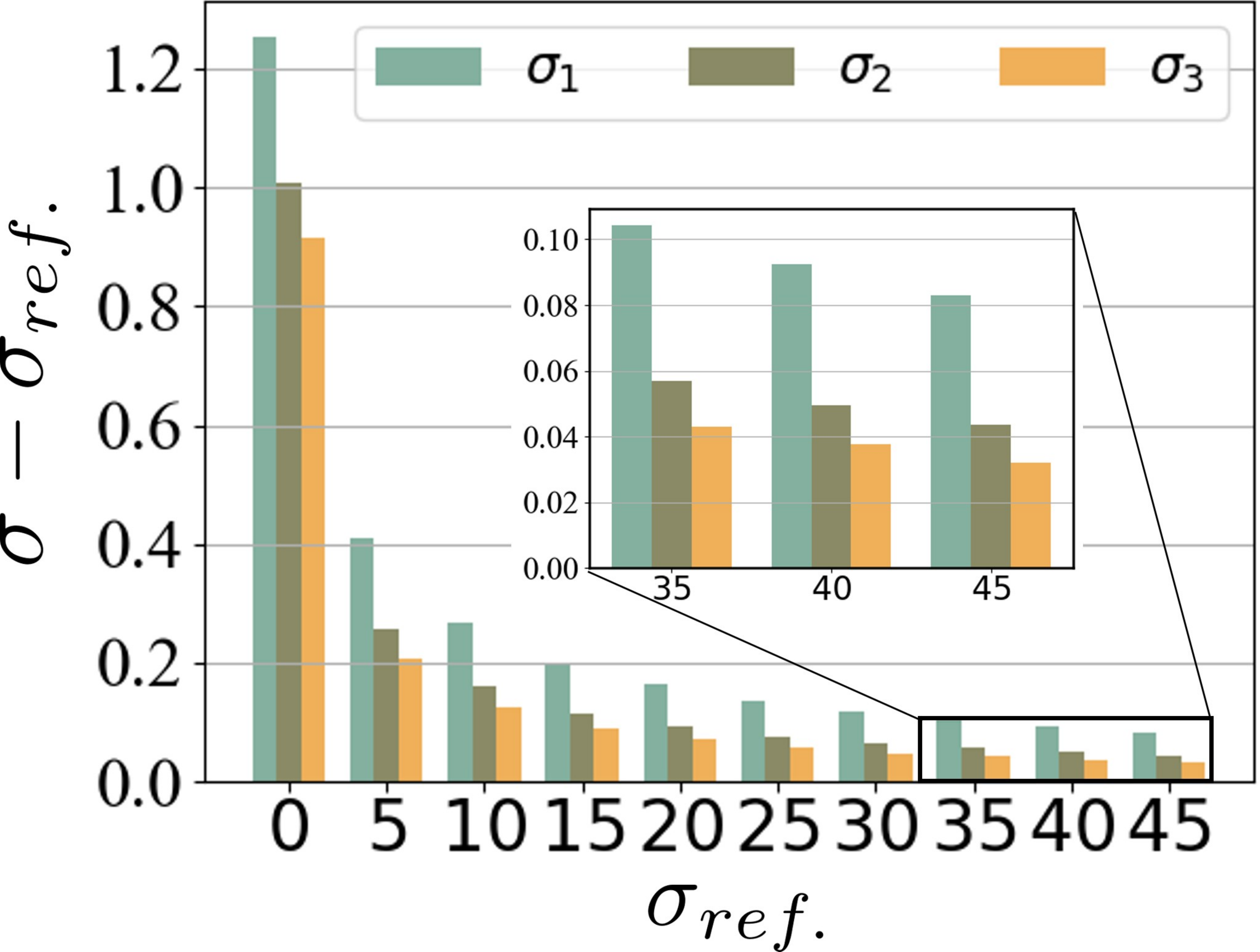}   
			& \includegraphics[width=0.48\linewidth, height=0.39\linewidth]{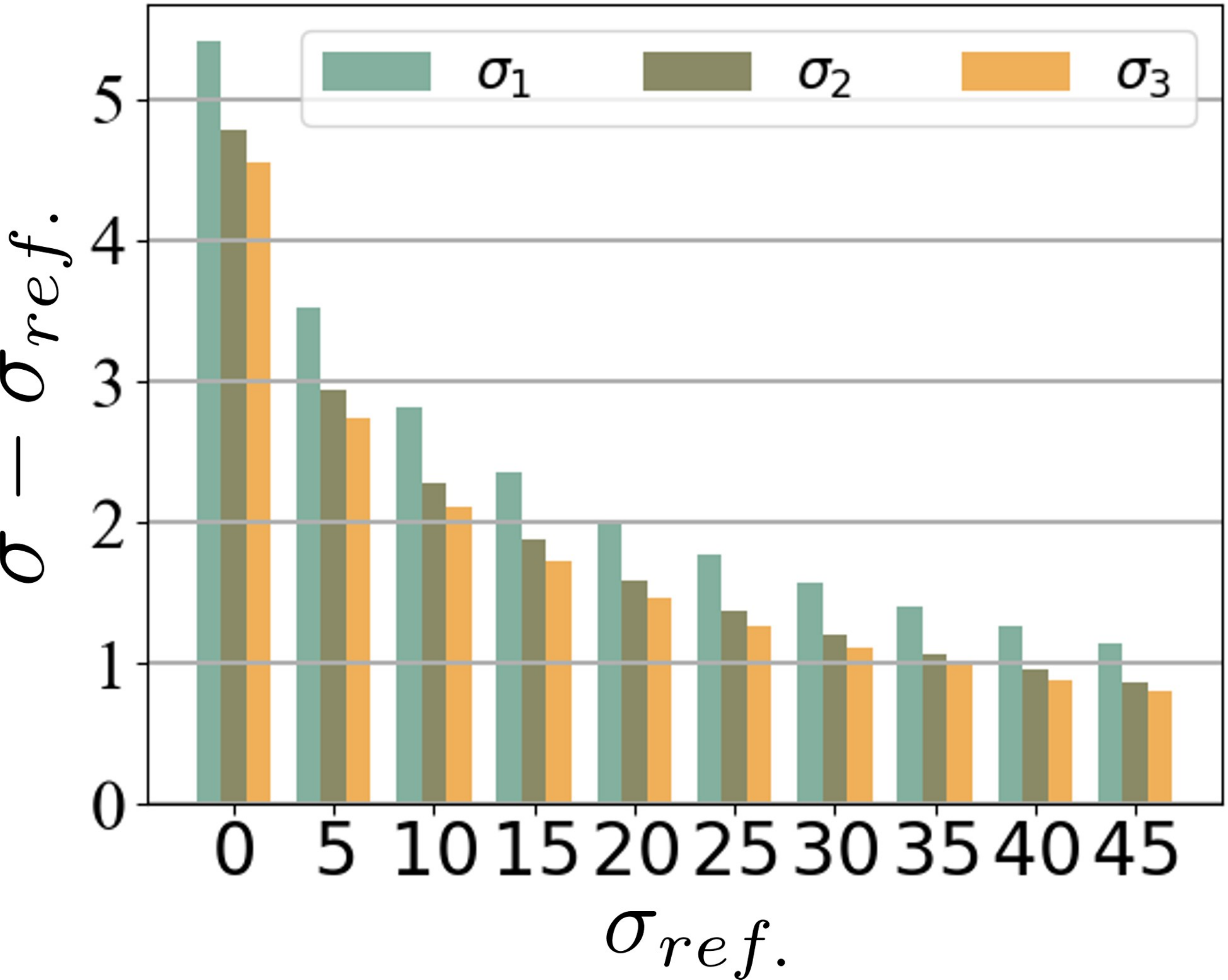}  \\ 			
			\specialrule{0em}{-0.5pt}{-0pt} 
			MIT dark images & LOL normal images \\ \specialrule{0em}{-0.5pt}{-0pt}  
		\end{tabular*}
	\end{center}
	\vspace{-0.2cm}
	\caption{
		Comparison of noise level estimation error in different datasets. $\sigma$ represents the estimated noise level, while $\sigma_{ref.}$ denotes the ground truth of the added noise. 
	}
	\label{fig:high-order}\vspace{-0.2cm}
\end{figure}

	In this section, we mainly focus on the effectiveness of our proposed noise estimation method for low light conditions.
	
	To verify the effectiveness of using the higher-order gradient of images to estimate the image noise intensity ${\sigma}$, we randomly selected 10 noiseless low-light images on the MIT dataset and 10 noiseless normal images on the LOL dataset, respectively. On this basis, we then added Gaussian noise with different noise intensity ${\sigma}_{ref.}$ and estimated the noise intensity ${\sigma}_n$ using a noise estimator of order $n$ (i.e., $n=1,2,3$). The process was repeated 100 times for each image. As shown in Figure~\ref{fig:high-order}, %with increasing noise intensity, 
	the noise estimation accuracy increases with increasing noise intensity and also with increasing order of the image gradient used. The experiments prove that the 1-st order noise estimator remains high accuracy for high intensity noise($1\%$ error when sigma is greater than 20 in low-light images), and therefore the experiment defaults to the 1st order noise estimator.
	It can be observed that our method is better suited for estimating noise intensity under low-light conditions. Due to the generally lower pixel intensity under low-light conditions, the impact of image texture details on the noise estimation is relatively minor, compared to the estimated noise under normal lighting conditions.

	\begin{table}[htbp]
		\centering
		%	\begin{threeparttable}
			\renewcommand{\arraystretch}{1.2}
%			\smaller
			\caption{The average absolute error(ABE) and CPU running time of different noise estimation methods on random selected low-light images from MIT dataset. The best result is in red whereas the second best one is in blue.}
			\label{tab:noise_level_estimation}
			\vspace{10pt}
			% Table generated by Excel2LaTeX from sheet 'noise'
			\begin{tabular}{|c|cc|cccc|c|}
				\hline
				\multicolumn{1}{|c|}{\multirow{2}[1]{*}{Metric}} &
				\multicolumn{2}{c|}{\multirow{2}[1]{*}{Methods}} & \multicolumn{4}{c|}{$\sigma_{ref.}$} & \multicolumn{1}{c|}{\multirow{2}[1]{*}{Time(s)}} \bigstrut[t]\\
				& \multicolumn{2}{c|}{} & 5 & 15 & 25 & 35 &  \\
				\hline
				\hline
				\multirow{4}[0]{*}{ABE$\downarrow$}
				& \multicolumn{2}{c|}{Pyatykh.~\cite{NoisePCA2012TIP}} & 0.073 & 0.452 & 0.922 & 1.338 & 1.847 \\
				&\multicolumn{2}{c|}{Liu.~\cite{NoiseLevelEstimation2013TIP}} & \textcolor{blue}{\textbf{0.036}} & 0.197 & 0.373 & 0.506 & 7.916 \\
				&\multicolumn{2}{c|}{Chen.~\cite{StatisticalMethod2015TIP}} & \red{\textbf{0.030}} & \red{\textbf{0.067}} & 0.119 & 0.181 & 0.038 \\
				&\multicolumn{2}{c|}{IVHC~\cite{IVHC2016TIP}} & 0.275 & 0.335 & 0.550 & 0.734 & 0.012 \\
				\hline
				\hline
				\multirow{3}[0]{*}{ {ABE$\downarrow$}} &
				\multirow{3}[2]{*}{Ours} & $\sigma_1$ & 0.409 & 0.201 & 0.137 & 0.104 & \red{\textbf{0.003}} \\
				& & $\sigma_2$ & 0.255 & 0.116 & \textcolor{blue}{\textbf{0.077}} & \textcolor{blue}{\textbf{0.057}} & \textcolor{blue}{\textbf{0.003}} \\
				& & $\sigma_3$ & 0.206 & \textcolor{blue}{\textbf{0.089}} & \red{\textbf{0.059}} & \red{\textbf{0.043}} & 0.006 \\
				\hline
			\end{tabular}%
		\end{table}

	For a more comprehensive quantitative assessment, we conducted a comparative analysis of our methods against various noise estimation techniques, emphasizing the assessment of absolute estimated errors and CPU running times. In this comparative study, we utilized identical images selected from the MIT dataset and subsequently introduced Gaussian noise with varying intensities. The outcomes are presented in Table~\ref{tab:noise_level_estimation}. 
	As the additional noise intensity increases, the average absolute error demonstrates an upward trend for the compared methods, whereas our method exhibits a decline in error rate.
	Evidently, our proposed method demonstrates superior accuracy across most noise levels while demanding significantly less computational time. Notably, even without GPU acceleration, ours running time is approximately fourfold shorter compared to the second-ranked method.

	To comprehensively evaluate the advantages of integrating estimated noise levels, we executed a series of experiments. In instances where denoising models like IRCNN~\cite{IRCNN} and DnCNN~\cite{DnCNN} traditionally do not consider noise levels as additional inputs, we adopted a methodology involving the embedding of our estimated noise into a vector. This embedded vector was seamlessly integrated into the shadow layer of these widely-used denoising models through the incorporation of multiple linear layers.
	Regarding CBDNet~\cite{CBDNet}, a model typically reliant on neural network-based noise estimation, we replaced its original estimation module with our proposed noise estimation method.
	All models underwent training using the common Mean Squared Error (MSE) loss function for image restoration. Training procedures utilized a dataset comprising 596 clear low-light images sourced from the MIT dataset. Subsequently, testing involved 99 additional images, each augmented with Gaussian noise characterized by dynamically varying intensities.
	Importantly, during training, the noise intensity ranged randomly from 2 to 50, while during testing, it varied from 5 to 35. This random variation ensured that the denoising models remained oblivious to specific noise levels, maintaining a blind status throughout the entire process.
	Each individual model underwent training for a total of 4000 iterations.
	This comprehensive experimental setup facilitated a thorough evaluation of how our estimated noise impacted the denoising models' performance.

	\begin{table}[htbp]
		\centering
		%	\begin{threeparttable}
			\renewcommand{\arraystretch}{1.2}
%			\smaller
			\caption{Denoising performance of applying noise estimation method on different models.}
			\label{tab:noise_input_influence}
			\vspace{10pt}
			% Table generated by Excel2LaTeX from sheet 'noise'
			\begin{tabular}{|c|cc|cc|}
				\hline
				\multirow{2}[2]{*}{Methods} & \multicolumn{2}{c|}{Original} & \multicolumn{2}{c|}{w/ Noise Input} \bigstrut[t]\\
				& PSNR & \#Params(M) & PSNR & \#Params(M) \bigstrut[b]\\
				\hline
				\hline
				IRCNN & 34.6645 & 0.19  & $35.2346_{0.57\uparrow}$ & $0.29_{0.10\uparrow}$ \bigstrut[t]\\
				DnCNN & 28.6796 & 0.56  & $35.6878_{7.01\uparrow}$ & $0.66_{0.10\uparrow}$ \\
				CBDNet & 36.6844 & 4.36  & $37.0639_{0.38\uparrow}$ & $4.34_{0.02\downarrow}$ \bigstrut[b]\\
				\hline
			\end{tabular}%
		\end{table}

		From the findings presented in Table~\ref{tab:noise_input_influence}, it is evident that methods incorporating our estimated noise level input significantly outperform the original results. Notably, our rapid statistical noise estimation method showcases superior efficiency compared to the noise estimation network in CBDNet, for it achieves higher performance even with fewer parameters without using network to estimate the noise intensity.

		\subsection{Low-light Image Enhancement}
		We have conducted experiments to demonstrate the accuracy and efficiency of our noise estimation method, which has showcased promising potential for further low-light image denoising. Building upon this progress, we are investigating the advantages of employing our proposed noise estimation method within a denoising-first and enhancing-later framework for low-light image enhancement.
		The datasets and training configurations are enumerated below:
		
		%		\noindent
		\textbf{Benchmark Datasets and Evaluation Metrics.} 
		To rigorously evaluate the effectiveness of our proposed method in low-light image enhancement, we conducted assessments across multiple datasets. The evaluation encompassed well-established datasets such as MIT~\cite{bychkovsky2011learning} and LOL~\cite{wei2018deep}, renowned for their diverse light condition images. 
		%Additionally, we included two recently introduced datasets: Darkface~\cite{poor_visibility_benchmark} tailored for face detection and Zurich~\cite{sakaridis2019guided} designed specifically for nighttime semantic segmentation tasks.
		The MIT dataset contains noiseless low-light images, serving as a benchmark for comparison, while the LOL dataset includes images captured under various light conditions, often exhibiting significant noise.
		Additionally, we included several unpaired datasets including DICM~\cite{DICM}, LIME~\cite{LIME}, MEF~\cite{MEF}, NPE~\cite{NPE} and VV~\cite{VV} for testing directly.
		In evaluating the efficacy of our method, we employed a comprehensive set of evaluation metrics. This includes two full-reference metrics, PSNR and SSIM~\cite{wang2004image}, known for their precision in assessing image quality. Furthermore, we incorporated no-reference metrics, DE~\cite{DE}, EME~\cite{EME}, LOE~\cite{LOE} and NIQE~\cite{NIQE}, essential for evaluating images in the absence of reference images.
		%For the semantic segmentation task, we employed three additional full-reference metrics: aAcc, mIoU, and mAcc, providing a holistic assessment of segmentation performance under low-light conditions.
		
		\textbf{Training Configuration.}
		During the training phase, we adopted an unsupervised learning approach tailored to each dataset. Initially, the images were segmented into patches, each sized at $128\times128$, and a batch size of $16$ was utilized. Our training process employed the Adam optimizer with an initial learning rate set at $1 \times 10^{-4}$ and betas of $0.9$ and $0.999$, a configuration that proved effective in our experiments.
		We executed alternating optimization cycles between the Self-Calibrated Denoiser  (SCD) and the Learnable Illumination Interpolator(LII). Each optimization loop involved 50 successive updates to the LII module, followed by 50 updates to the SCD module. This alternating optimization strategy continued for 4000 iterations to effectively train the model.

		%\subsection{Benchmark Evaluation}
		\textbf{Benchmark Evaluation.}
		We performed quantitative and qualitative evaluations of the proposed method against several state-of-the-art LLIE methods, including traditional methods (i.e. LIME~\cite{LIME} and Structure-Revealing Retinex Model~\cite{SRRM}
%		, Brain-like Retinex~\cite{Brain_CAI2023109195}
		), supervised learning methods (i.e., 
		%	 Retinex~\cite{wei2018deep}, MBLLEN~\cite{lv2018mbllen}, KinD~\cite{zhang2019kindling} and DRBN~\cite{yang2020fidelity}), 
		Retinex~\cite{wei2018deep}, GLADNet~\cite{wang2018gladnet}, DRBN~\cite{yang2020fidelity} and MBLLEN~\cite{lv2018mbllen}),
		and unsupervised learning methods (i.e, EnGAN~\cite{jiang2021enlightengan}, ZeroDCE~\cite{guo2020zero}, RetinexDIP~\cite{zhao2021retinexdip}, RUAS~\cite{liu2021retinex}, and SCI
		~\cite{ma2022toward}
		). 
		Presented below are the experimental findings following the outlined experimental conditions and settings.

\begin{figure*}[htbp]
			\begin{center}
				\begin{tabular}{c@{\extracolsep{0.35em}}c@{\extracolsep{0.35em}}c@{\extracolsep{0.35em}}c@{\extracolsep{0.35em}}c@{\extracolsep{0.35em}}c@{\extracolsep{0.35em}}}
					\includegraphics[width=0.158\linewidth]{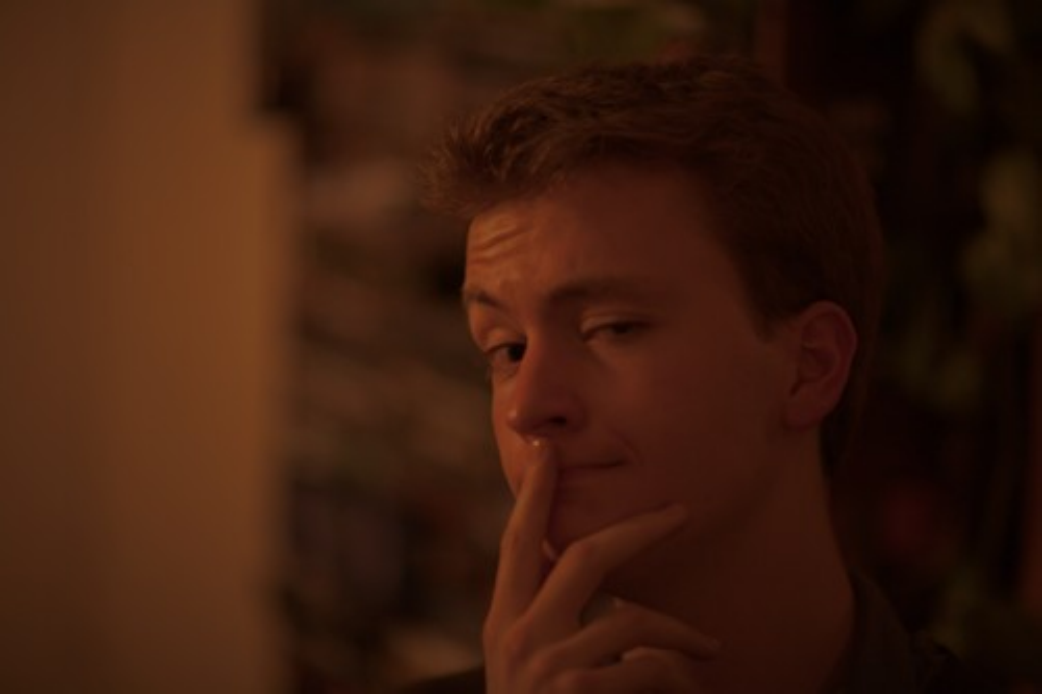}&\includegraphics[width=0.158\linewidth]{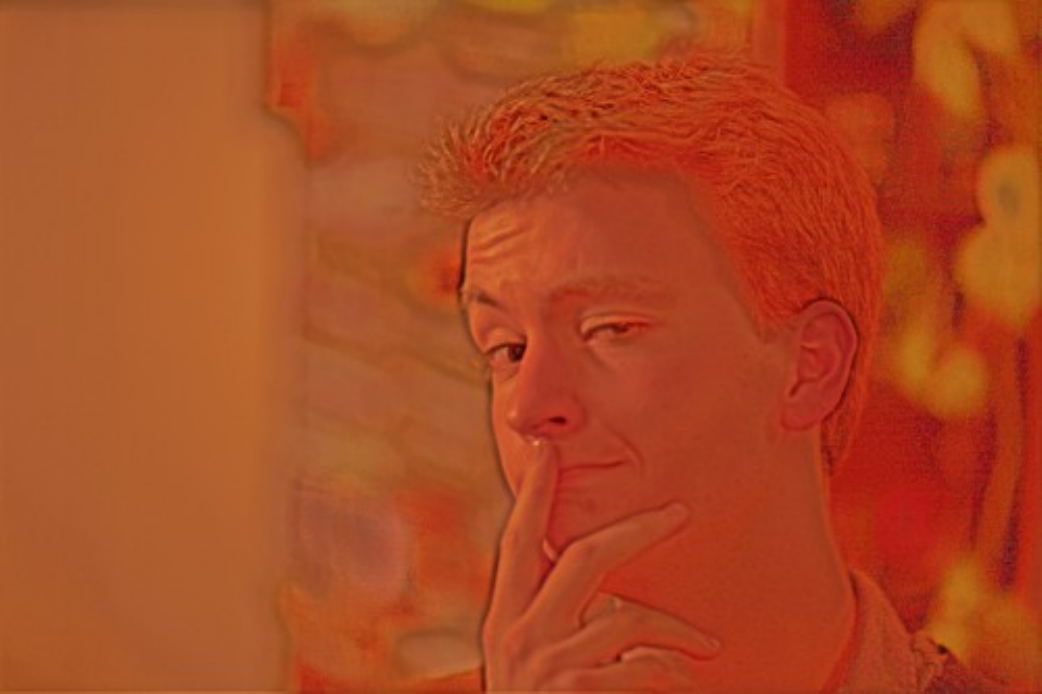}&\includegraphics[width=0.158\linewidth]{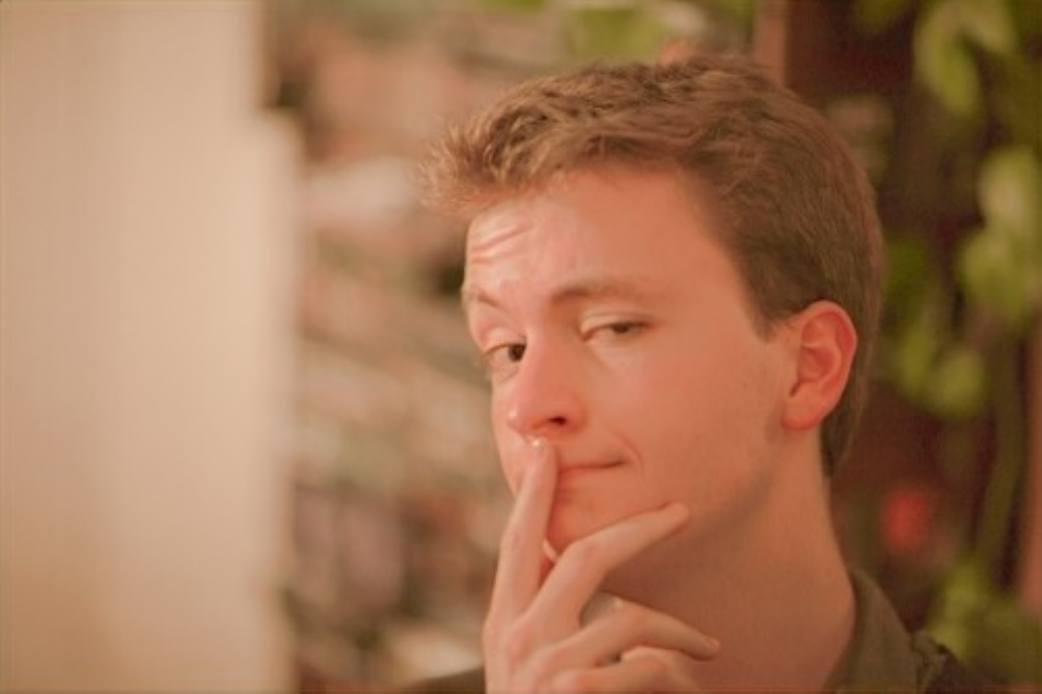}&\includegraphics[width=0.158\linewidth]{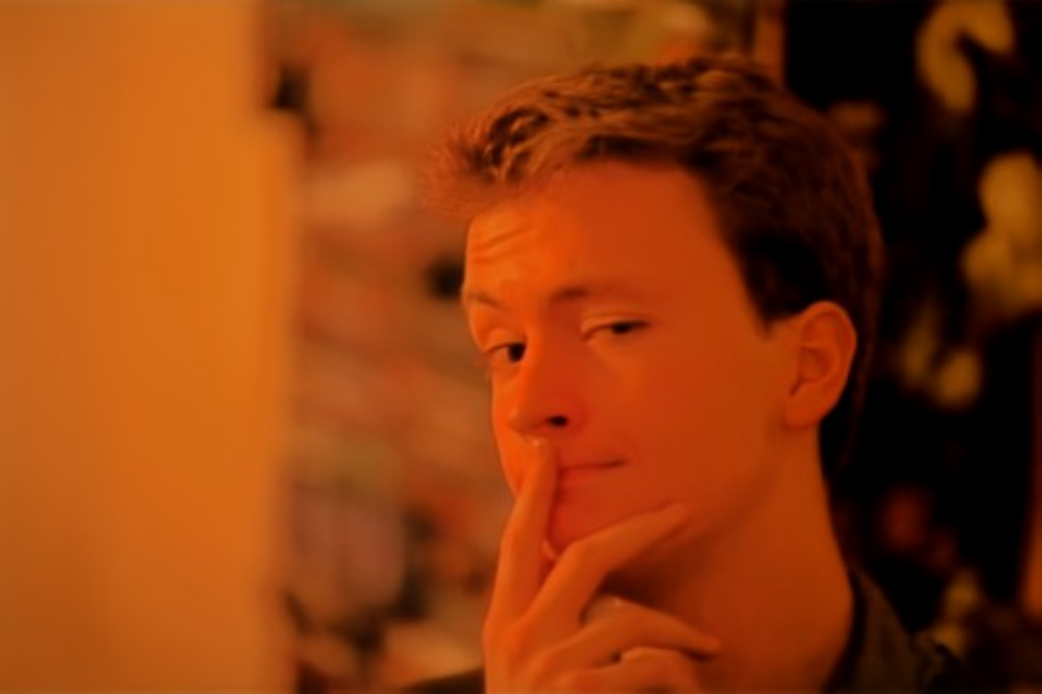} &\includegraphics[width=0.158\linewidth]{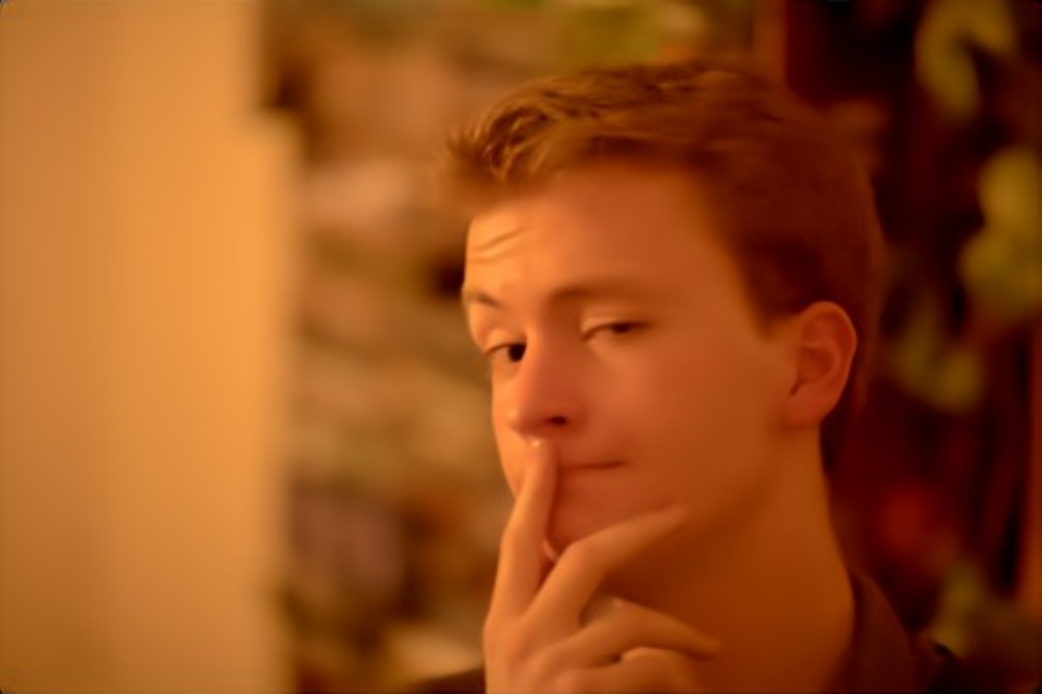}&\includegraphics[width=0.155\linewidth]{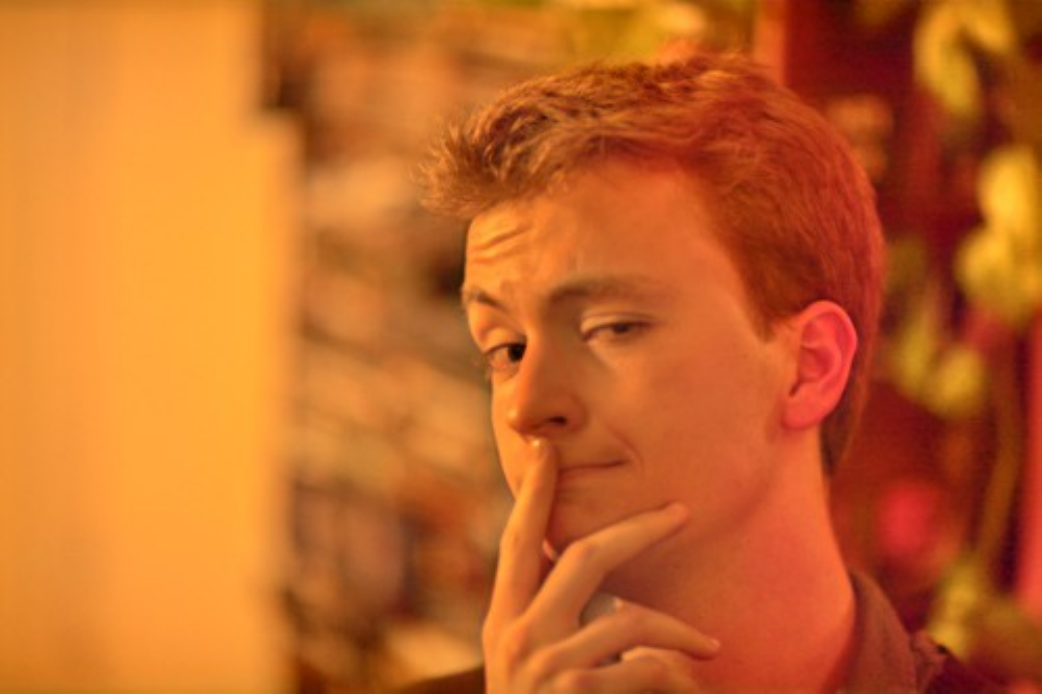} \\ \specialrule{0em}{-0.5pt}{-1pt} 
					
					\includegraphics[width=0.158\linewidth]{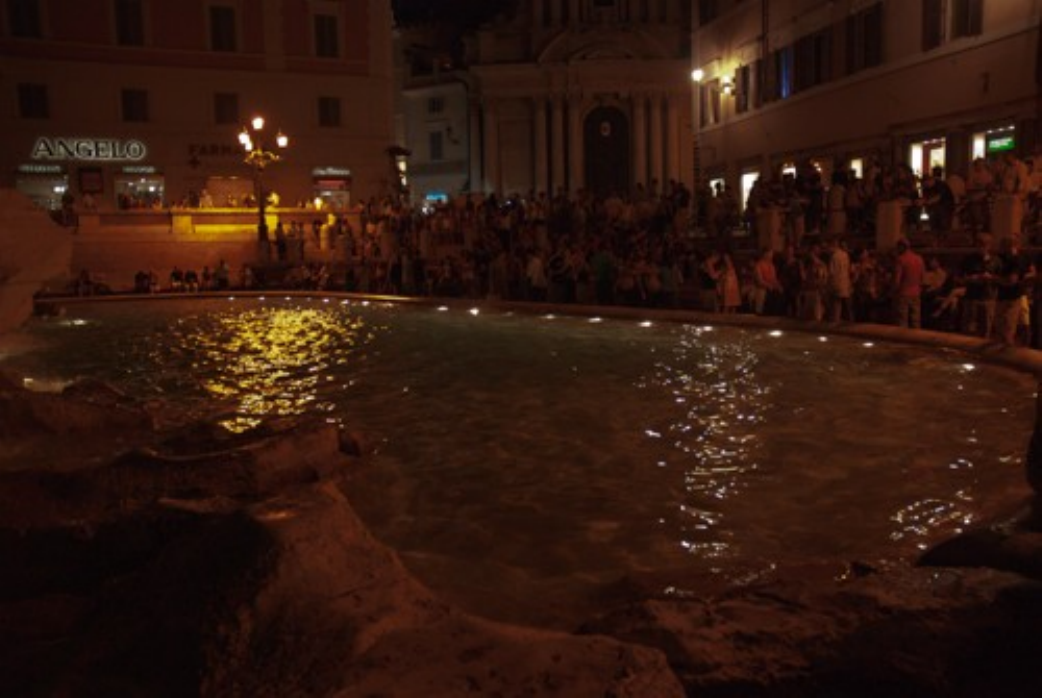}&\includegraphics[width=0.158\linewidth]{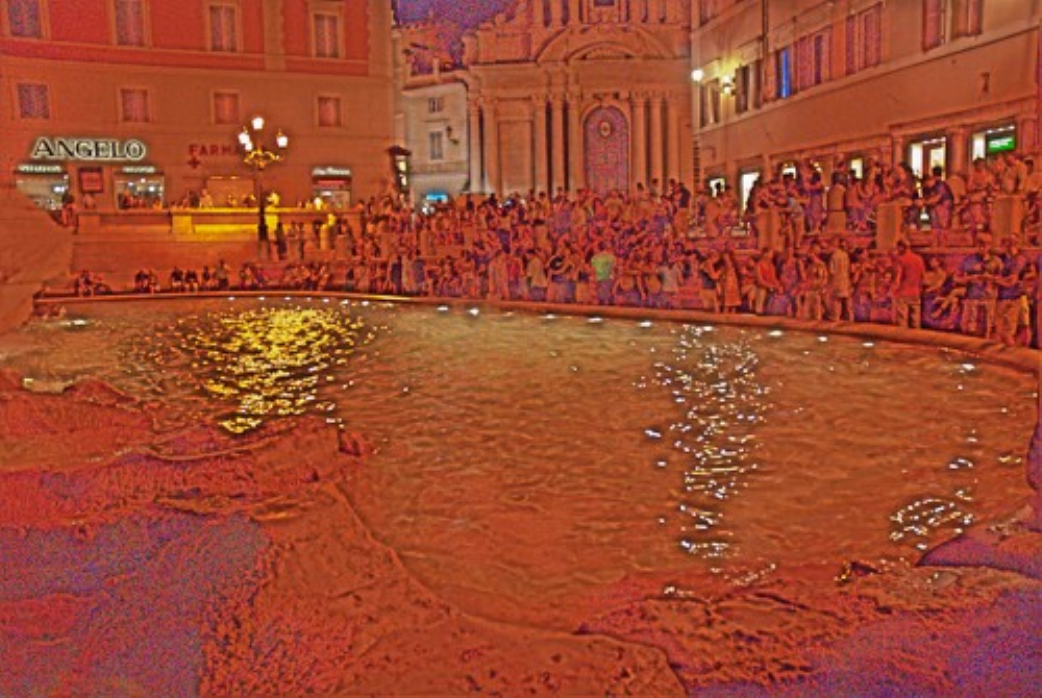}&\includegraphics[width=0.158\linewidth]{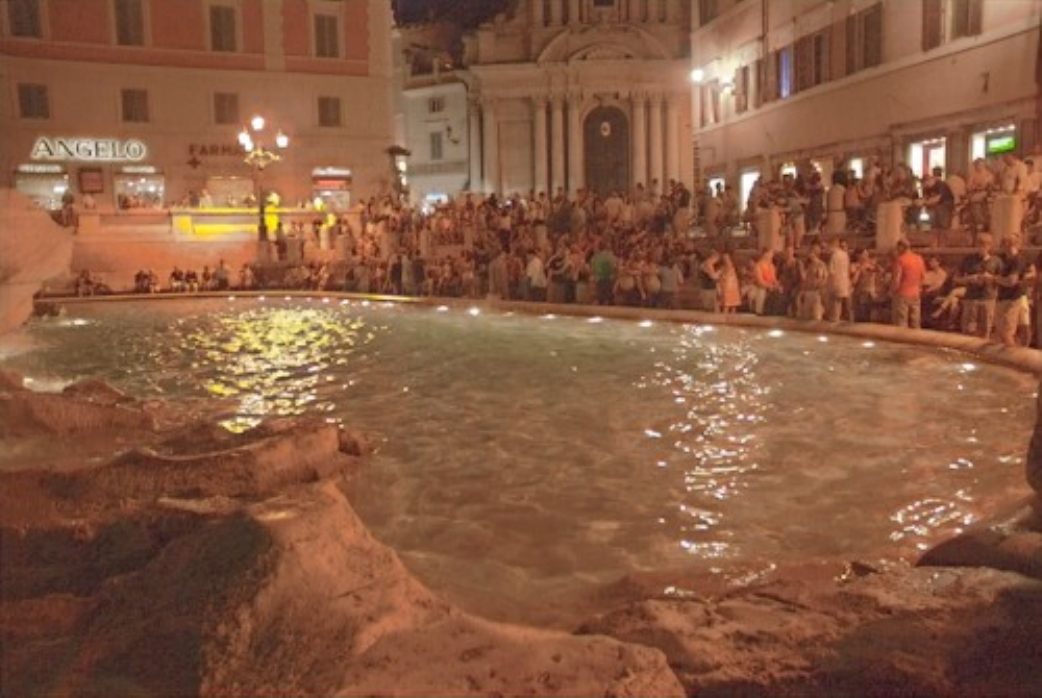}&\includegraphics[width=0.158\linewidth]{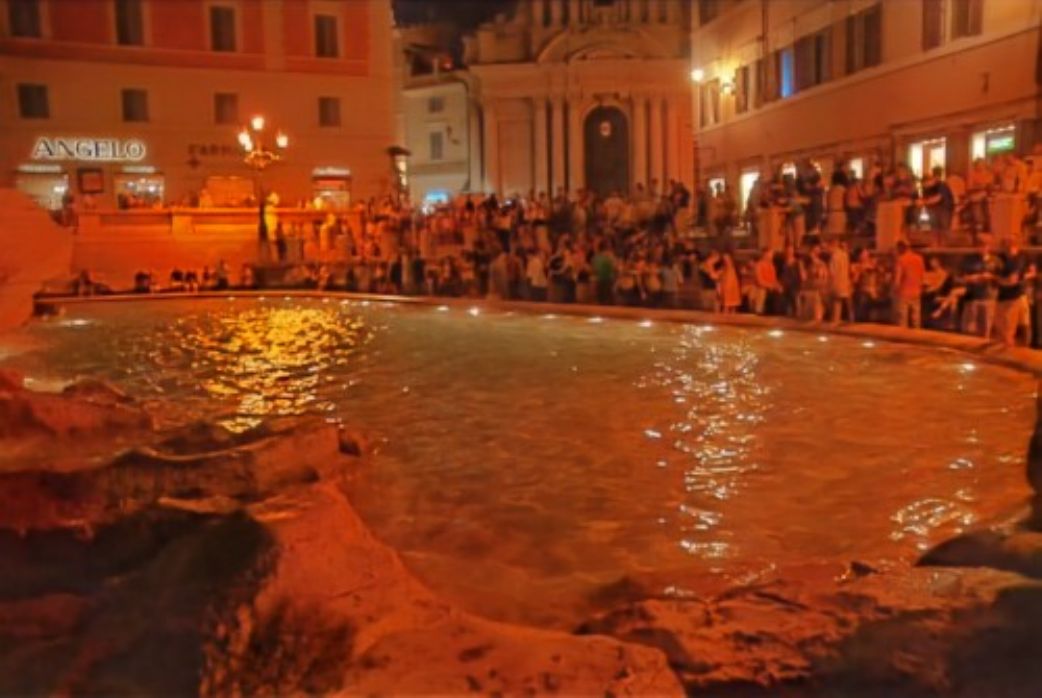} &\includegraphics[width=0.158\linewidth]{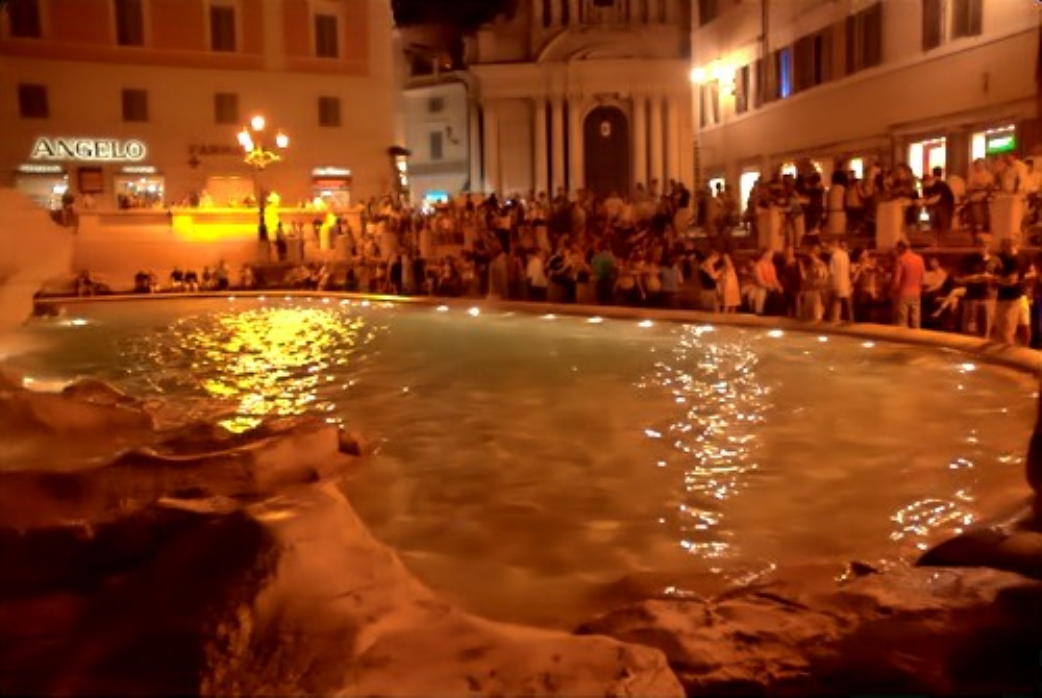}&		\includegraphics[width=0.158\linewidth]{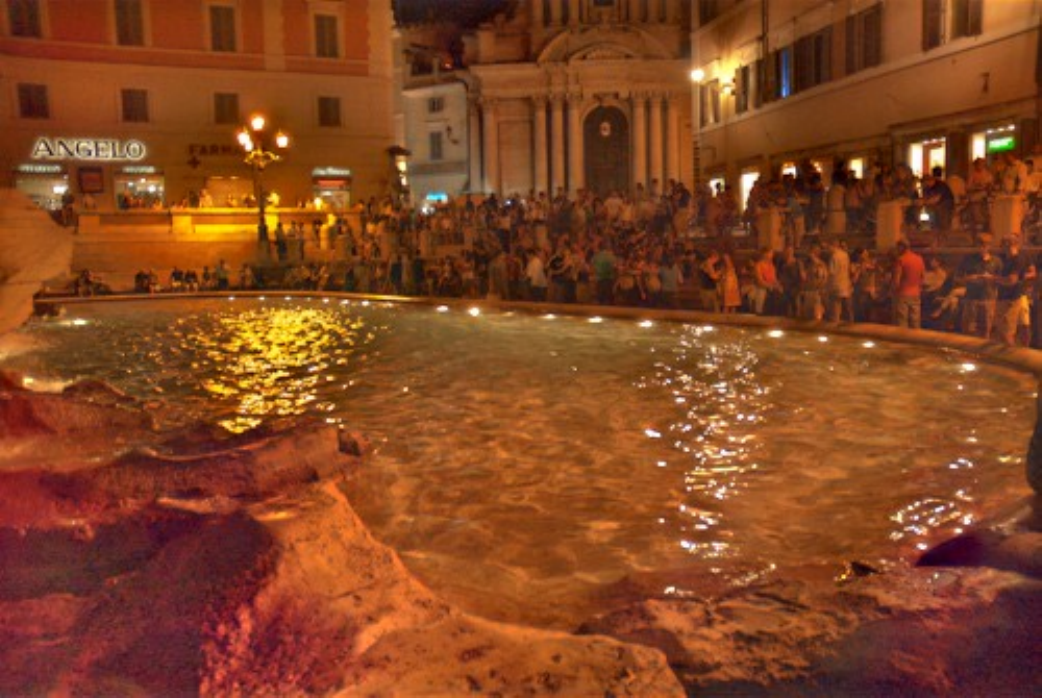}\\ \specialrule{0em}{-0.5pt}{-1pt} 		

					Input & Retinex & GLADNet & DRBN &MBLLEN & EnGAN \\  			
					\includegraphics[width=0.158\linewidth]{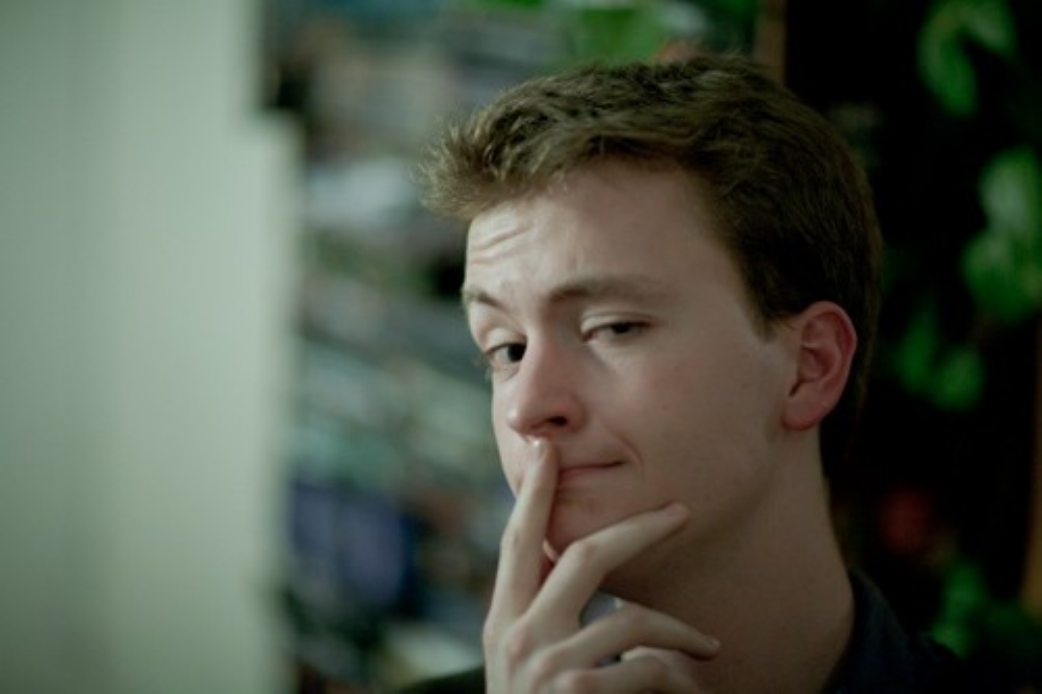}  & \includegraphics[width=0.158\linewidth]{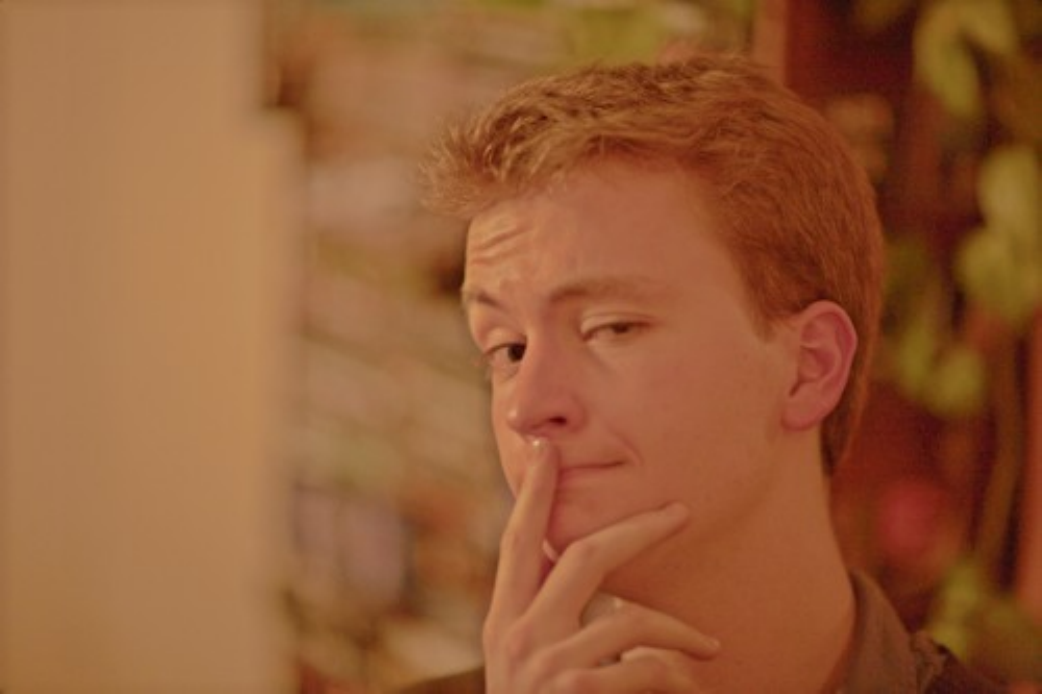}&\includegraphics[width=0.158\linewidth]{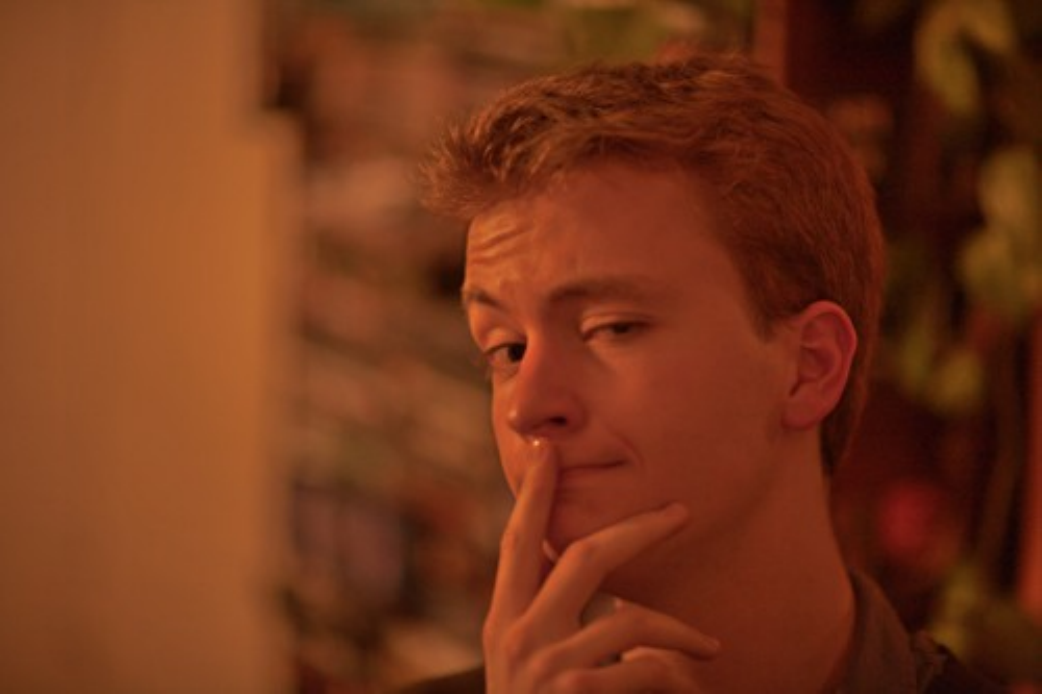}&\includegraphics[width=0.158\linewidth]{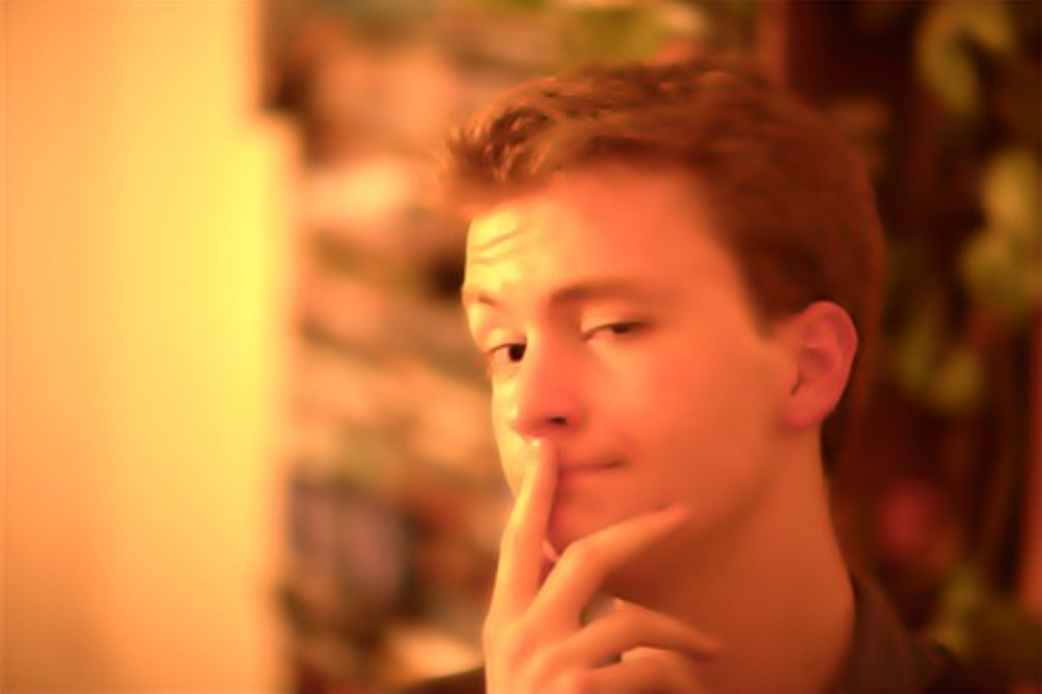}&\includegraphics[width=0.158\linewidth]{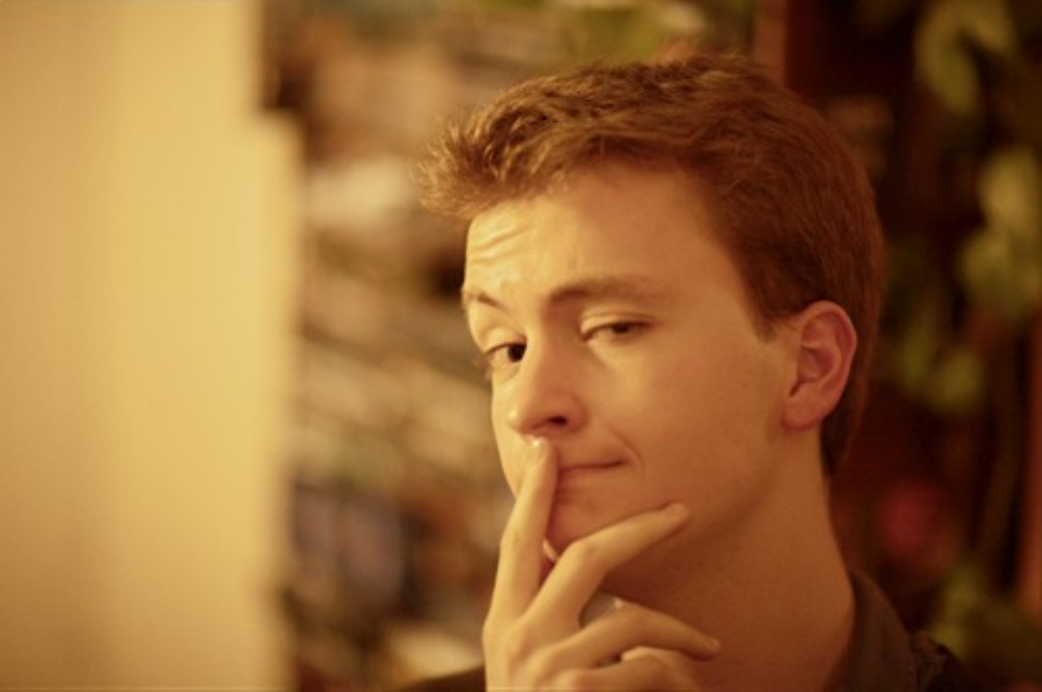}&\includegraphics[width=0.158\linewidth]{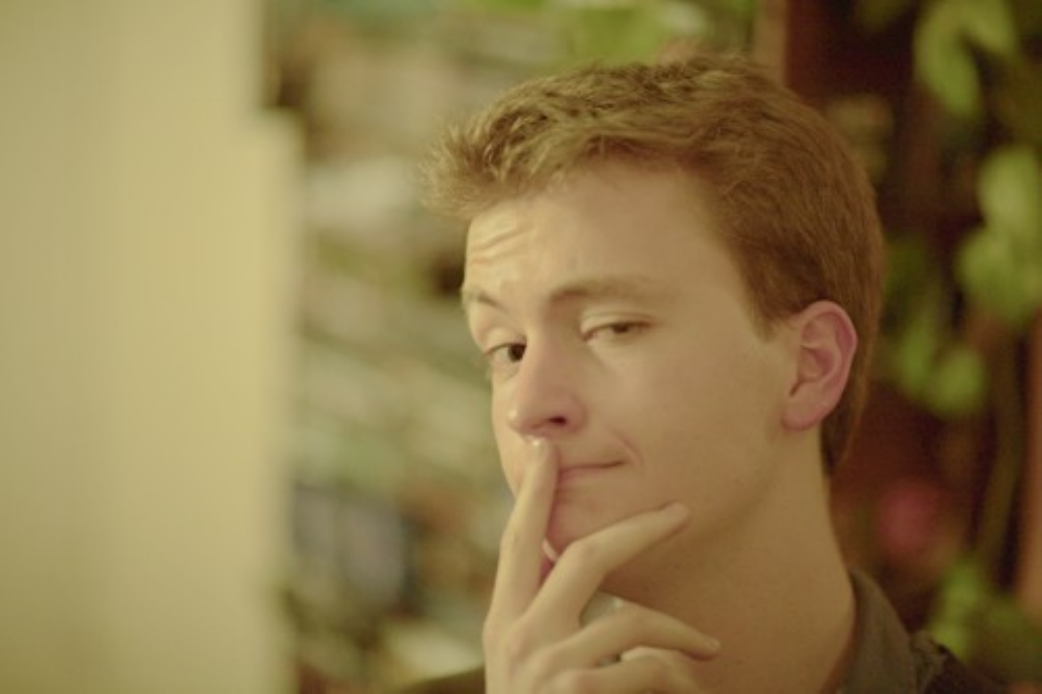} \\ 			
					\specialrule{0em}{-0.5pt}{-1pt} 
					
					\includegraphics[width=0.158\linewidth]{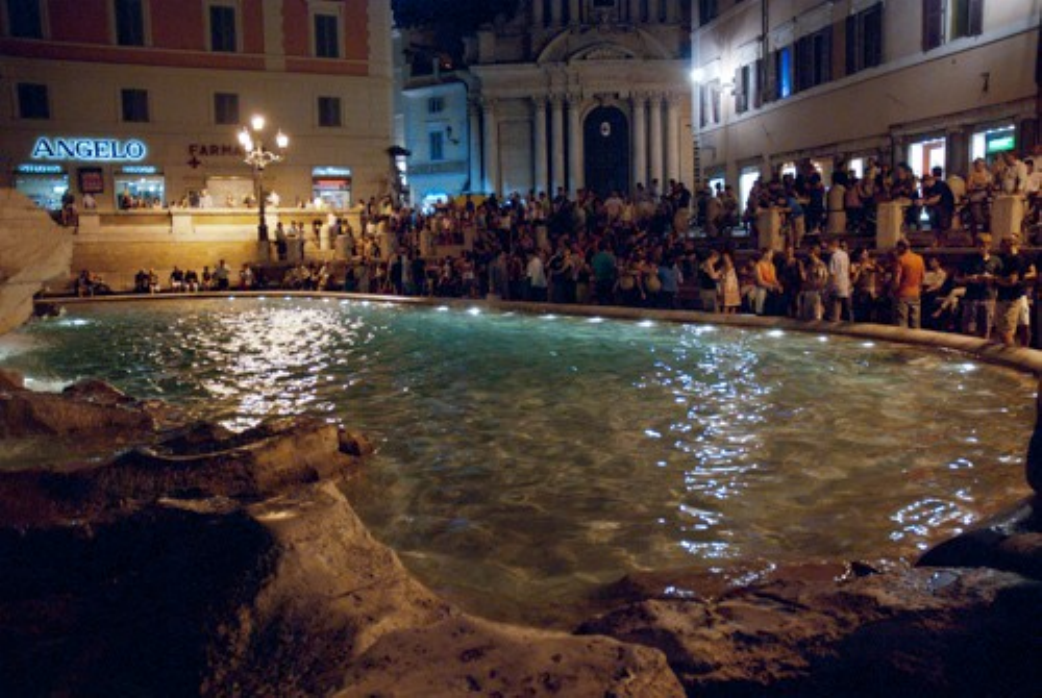}  & \includegraphics[width=0.158\linewidth]{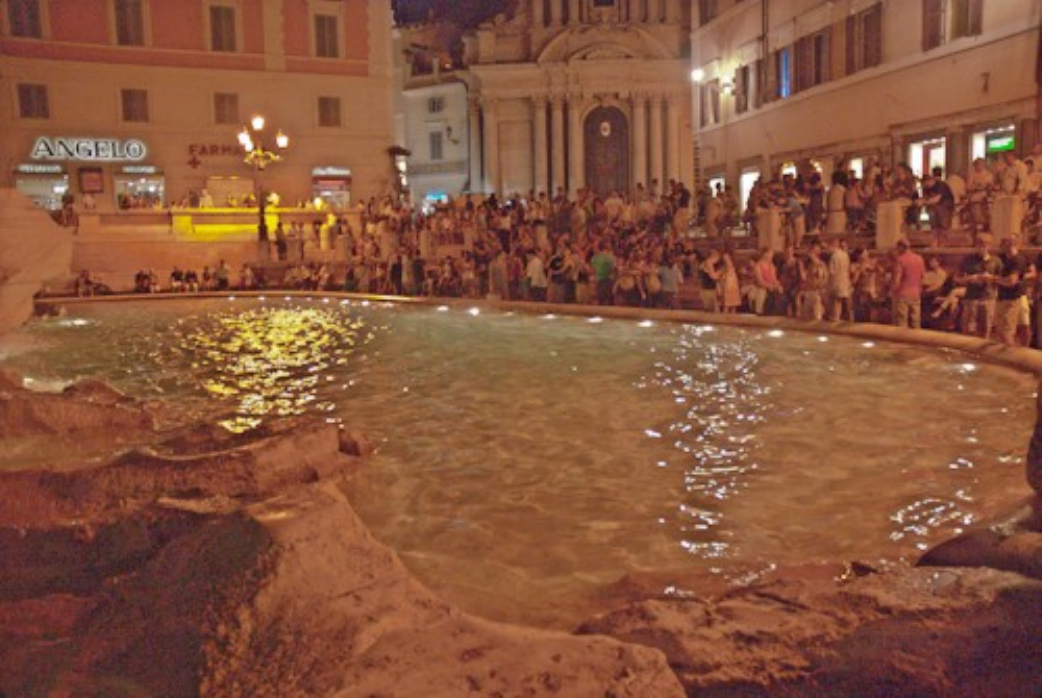}&\includegraphics[width=0.158\linewidth]{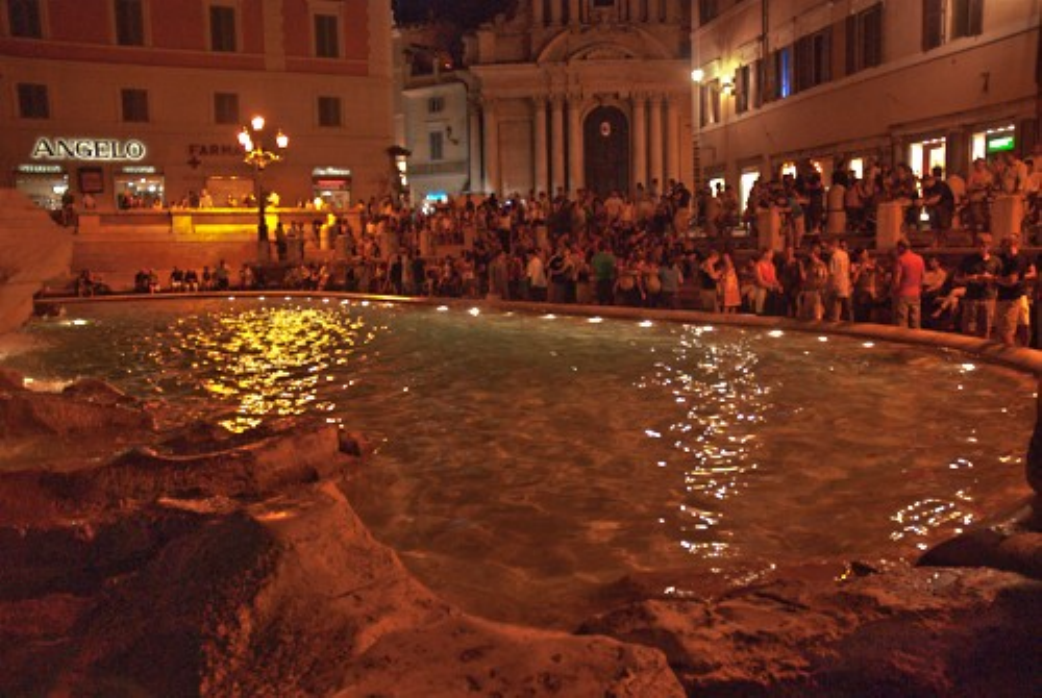}&\includegraphics[width=0.158\linewidth]{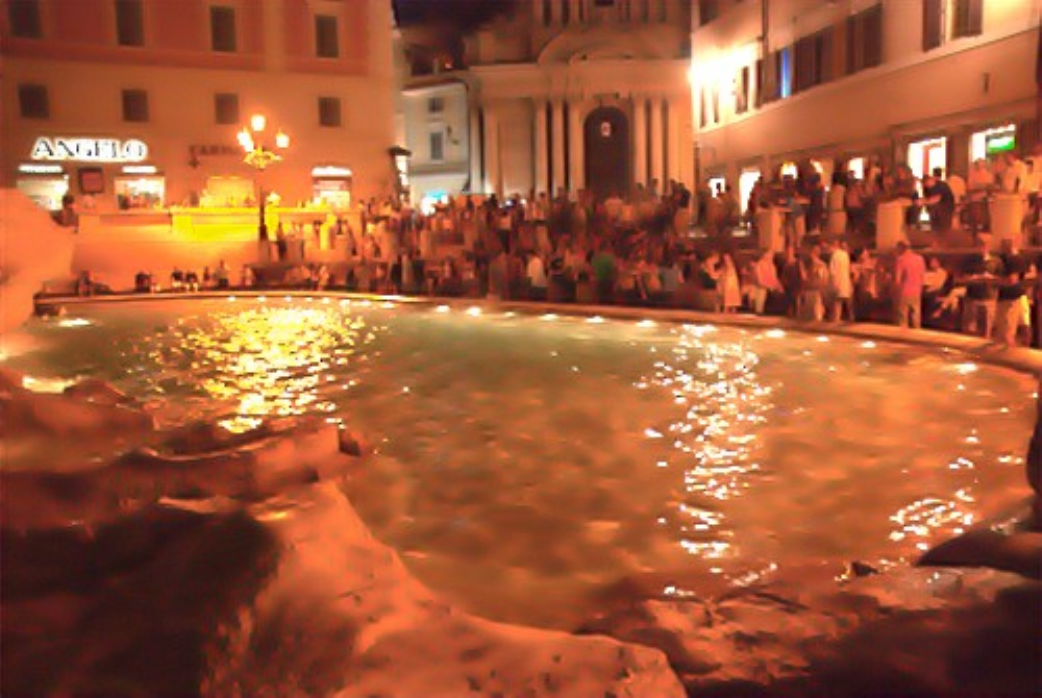}&\includegraphics[width=0.158\linewidth]{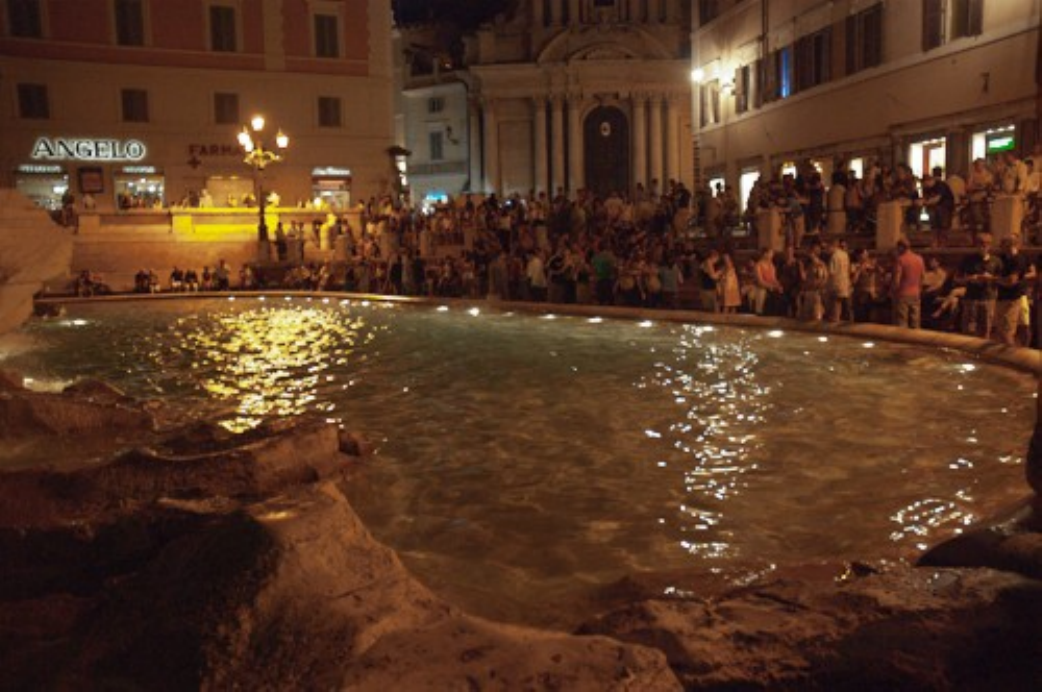}&\includegraphics[width=0.158\linewidth]{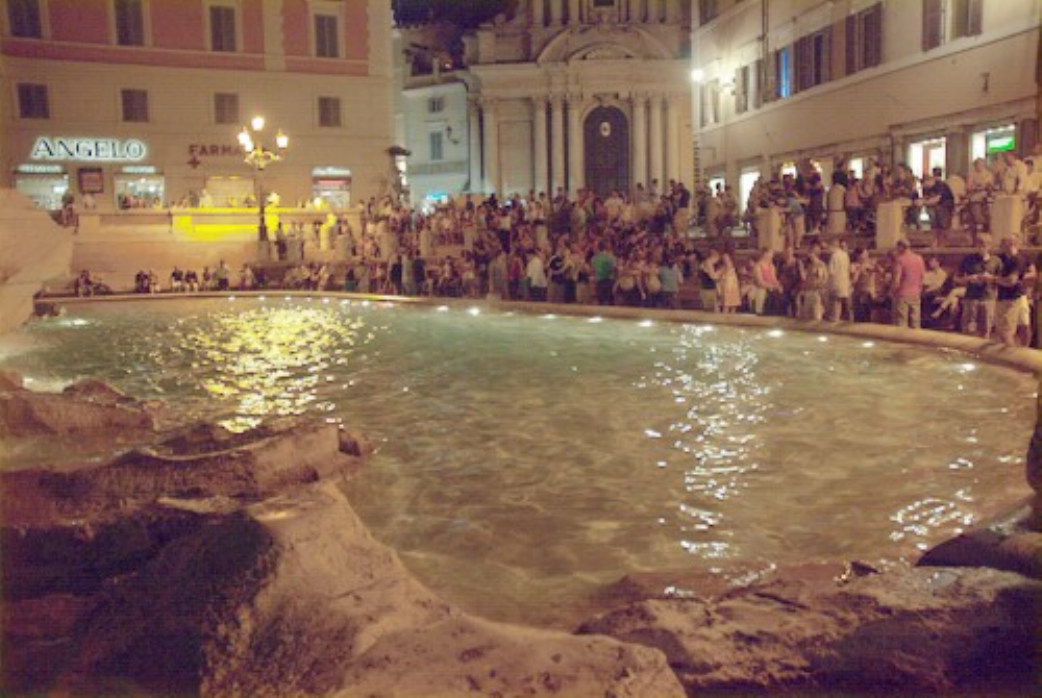} \\ 			
					\specialrule{0em}{-0.5pt}{-1pt} 
					Reference & ZeroDCE & RetinexDIP & RUAS & SCI & Ours \\		
					\specialrule{0em}{-2pt}{-2pt}  
				\end{tabular}
			\end{center}
			\vspace{-0.2cm}
			\caption{ 
				Visual results of state-of-the-art LLIE methods  and ours on the MIT dataset. \textit{Zoom in for best view.}	
			}
			\label{fig:MIT2}\vspace{-0.2cm}
		\end{figure*}

		\textbf{Qualitative evaluation.} 
		%	Qualitative results are displayed in Figures~\ref{fig:MIT2}-\ref{fig:Darkface2}. 
		Qualitative results are displayed in Figure~\ref{fig:MIT2}. 
		As illustrated, previous efforts did not achieve the desired enhancement results, resulting in inconspicuous details, unnatural colors and overexposure or underexposure. 
		In contrast, proposed method achieves the best visual quality with vivid colors and outstanding texture details.
		%Note that the relevant experiments on Darkface in Figure~\ref{fig:Darkface2} are based only on the constructed learnable illumination interpolator module, demonstrating the superiority of the proposed LII even in the absence of SNR.
		%In Figure \ref{fig:dark-zurich}, we observe that existing methods, such as RUAS, ZeroDCE, and SCI, produce over-exposed results without external hyperparameters finetuning or early stop strategies. 
%		Although RetinexDIP exhibits better visual performance, its iterative optimization process is too time-consuming to achieve real-time enhancement speeds.
		In contrast, our approach is both fast and flexible, and delivers highly effective results. We achieve real-time enhancement without the need for external hyperparameters finetuning or early stop strategies, and demonstrate superior performance compared to existing methods.
		%		Due to space limitations, more visual comparisons are available in the supplemental materials.
		
\begin{figure*}[htbp]
			\begin{center}
				\includegraphics[width=0.96\linewidth]{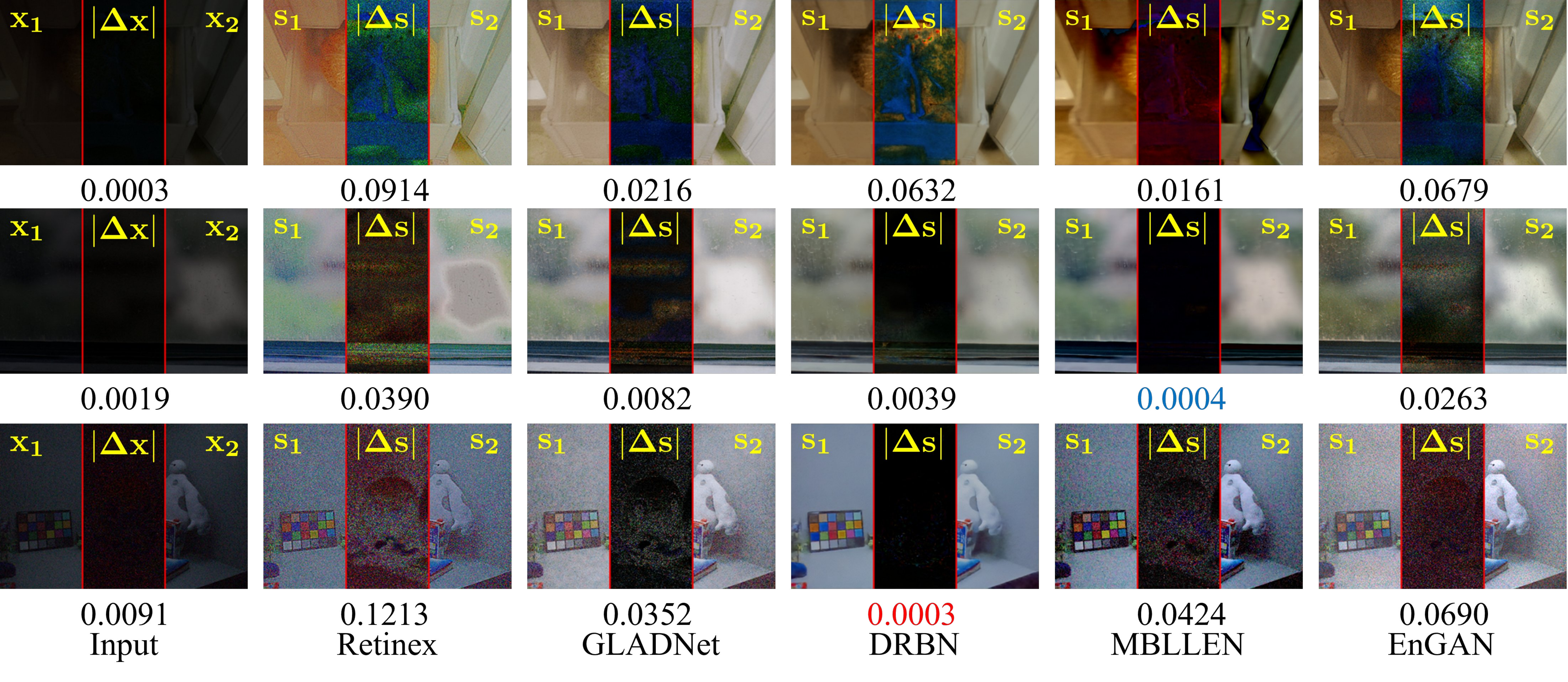} \\
				\includegraphics[width=0.96\linewidth]{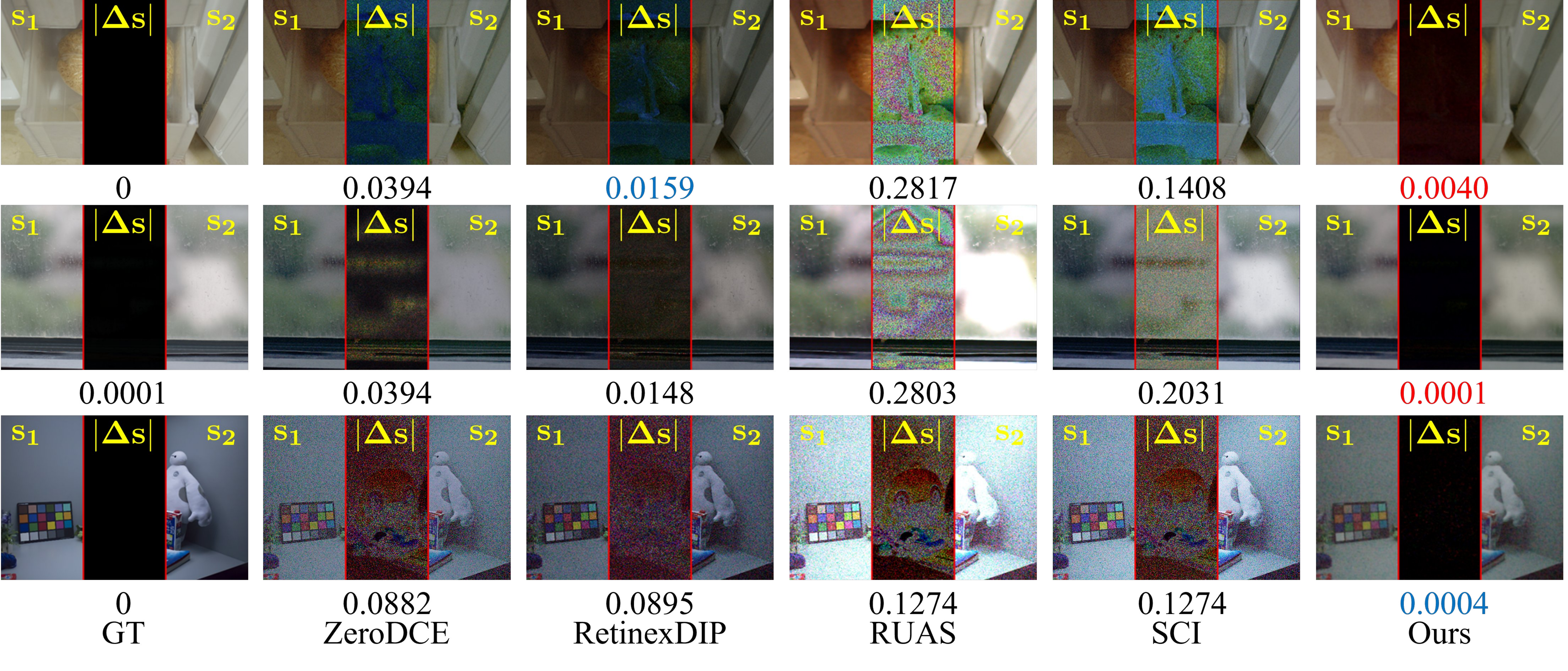}
			\end{center}
			\vspace{-0.4cm}
			\caption{
				Robust enhancement towards different similar images in LOL dataset. 
				$\mathbf{s_i}$ presents the enhanced result of $\mathbf{x_i}$ and $\mathbf{\Delta s} = 10(\mathbf{s_{1} - s_{2}})^2$ means scaled square difference between results for better comparison (darker is better). Mean square error are listed at the bottom to quantify the difference of similar inputs/results (smaller is better). The best result is in red whereas the second best one is in blue.
			}
			\label{fig:robust enhancement}
			%	\label{fig:charm_of_color_loss}\vspace{-0.1cm}
		\end{figure*}

		Further, to demonstrate the stability of our method in dynamic environments, we select pairs of images in test set of LOL with the same content but captured under different lighting conditions, many of which are slight different with each other.
		Despite the minor variations in the input images, achieving stable appearance including pixel intensity remains a challenge.
		Additionally, we use mean square error as the metric to compare the enhanced images and results are revealed in Figure~\ref{fig:robust enhancement}.
		Our method consistently achieves the lowest differences among the unsupervised methods and outperforms most of the supervised methods. 
		Specifically, our proposed method generates results with smaller differences compared to other approaches when handling low-light images with subtle differences. 
		This finding further validates the robustness of our approach and its ability to defend against slight changes while producing stable results.
		
\begin{table}[htbp]
	\renewcommand{\arraystretch}{1.2}
	\caption{Quantitative comparison with traditional methods (Trad.), supervised methods (Super.) and unsupervised methods (Unsuper.) on MIT and LOL dataset regarding two full-reference metrics (i.e., PSNR and SSIM) and two no-reference metrics (i.e., LOE and NIQE) . The best result is in red whereas the second best one is in blue.}
	\vspace{5pt}
	\centering
%	\smaller
	\setlength{\tabcolsep}{0.6mm}{
		\begin{tabular}{|c|c|cccc|cccc|}
			\hline
			\multicolumn{2}{|c|}{Dataset} & \multicolumn{4}{c|}{MIT} & \multicolumn{4}{c|}{LOL} \bigstrut[t]\\
			\multicolumn{2}{|c|}{Metric} & \multicolumn{1}{c}{PSNR$\uparrow$} & \multicolumn{1}{c}{SSIM$\uparrow$} & \multicolumn{1}{c}{LOE$\downarrow$} & \multicolumn{1}{c|}{NIQE$\downarrow$} & \multicolumn{1}{c}{PSNR$\uparrow$} & \multicolumn{1}{c}{SSIM$\uparrow$} & \multicolumn{1}{c}{LOE$\downarrow$} & \multicolumn{1}{c|}{NIQE$\downarrow$} \bigstrut[b]\\
			\hline
			\hline
			
			\multicolumn{1}{|c|}{\multirow{2}[2]{*}{\rotatebox{90}{Trad.}}} & LIME & 17.393  & 0.687  & 587.408  & 3.717  & 17.654  & 0.511  & 366.894  & 9.302  \bigstrut[t]\\
			& SRRM & 17.306  & 0.662  & 501.561  & 5.150  & 14.700  & 0.660  & 416.025  & 6.955  \bigstrut[b]\\
%			& Brain-like &  16.652 & 0.664 & 514.577 & 4.225 & 11.926 & 0.497 & 314.583 & 7.989 \bigstrut[b]\\
			\hline
			\hline
			\multicolumn{1}{|c|}{\multirow{4}[2]{*}{\rotatebox{90}{Super.}}} & Retinex & 12.734  & 0.592  & 1820.125  & 4.387  & 16.308  & 0.436  & 1030.256  & 10.244  \bigstrut[t]\\
			& GLADNet & 16.120  & 0.644  & 261.987  & 3.921  & \textcolor[rgb]{ 0,  0,  1}{\textbf{19.697 }} & 0.656  & 374.294  & 7.438  \\
			& DRBN & 16.292  & 0.606  & 737.147  & 4.710  & 18.262  & \textcolor[rgb]{ 0,  0,  1}{\textbf{0.669 }} & 615.350  & 5.480  \\
			& MBLLEN & 17.519  & 0.616  & \textcolor[rgb]{ 0,  0,  1}{\textbf{184.634 }} & 5.266  & 17.666  & 0.577  & 367.522  & \textcolor[rgb]{ 0,  0,  1}{\textbf{5.466 }} \bigstrut[b]\\
			\hline
			\hline
			\multicolumn{1}{|c|}{\multirow{6}[4]{*}{\rotatebox{90}{Unsuper.}}} & EnGAN & 15.363  & 0.633  & 849.468  & 3.845 & 17.684  & 0.608  & 529.220  & 5.559  \bigstrut[t]\\
			& ZeroDCE & 16.000  & 0.665  & 496.208  & \textcolor[rgb]{ 1,  0,  0}{\textbf{3.678 }} & 16.402  & 0.573  & \textcolor[rgb]{ 0,  0,  1}{\textbf{227.478 }} & 8.895  \\
			& RetinexDIP & \textcolor[rgb]{ 0,  0,  1}{\textbf{17.865 }} & \textcolor[rgb]{ 1,  0,  0}{\textbf{0.716 }} & 505.883  & \textcolor[rgb]{ 0,  0,  1}{\textbf{3.812}}  & 12.530  & 0.505  & 386.879  & 8.819  \\
			& RUAS & 12.277  & 0.599  & 878.720  & 6.597  & 14.966  & 0.499  & 377.342  & 7.808  \\
			& SCI & 17.350  & 0.665  & 225.252  & 4.035  & 15.841  & 0.519  & \textcolor[rgb]{ 1,  0,  0}{\textbf{124.840 }} & 8.930  \bigstrut[b]\\
			\cline{2-10}      & Ours & \textcolor[rgb]{ 1,  0,  0}{\textbf{18.683 }} & \textcolor[rgb]{ 0,  0,  1}{\textbf{0.689 }} & \textcolor[rgb]{ 1,  0,  0}{\textbf{104.057 }} & 3.981  & \textcolor[rgb]{ 1,  0,  0}{\textbf{20.372 }} & \textcolor[rgb]{ 1,  0,  0}{\textbf{0.718 }} & 356.261  & \textcolor[rgb]{ 1,  0,  0}{\textbf{5.413 }} \bigstrut\\
			\hline
		\end{tabular}%
	}
	\label{tab:syn_metric}%
\end{table}%

%		\noindent
\textbf{Quantitative evaluation.} Table~\ref{tab:syn_metric} reports quantitative comparison results on the MIT and LOL datasets.  As can be seen, our method numerically outperforms existing unsupervised learning methods and even supervised learning methods by a large margin and ranks first in almost all metrics. 
%	In particular, compared to a recent study SCI, our method improves the PSNR by 1.3331dB and 4.5314dB on the MIT and LOL datasets, respectively, demonstrating the superiority of the proposed method.
%	 Note that the relevant experiments on MIT are based only on the constructed learnable illumination interpolator module, demonstrating the superiority of the proposed LII even in the absence of SNR. 
%	To ensure a fair comparison, we choose the medium version of SCI$^{\dagger}$ with the best performance and retrain it on the LOL dataset. 
In comparison to the second-best unsupervised state-of-the-art method on the MIT dataset, our approach achieves a PSNR improvement of 0.818dB. Furthermore, when compared to the best performing supervised method, our approach results in a PSNR improvement of 1.164dB. Likewise, on the LOL dataset, our method leads to improvements of 2.688dB and 0.675dB, respectively.
Benefiting from the systematic network architecture and reference-free loss we construct, our method outperforms the results of existing unsupervised methods and supervised learning methods on both MIT and LOL datasets significantly.
To delve deeper into the models' ability to generalize across unforeseen scenarios, we evaluated the unsupervised methods directly on the unpaired datasets. The results presented in Table~\ref{tab:non_ref_syn_metric} indicate comparable performance among the methods based on DE and NIQE metrics, although RUAS stands out due to its tendency for overexposure and a less favorable appearance. Notably, our method exhibits substantial superiority over others in terms of LOE and EME metrics.

\begin{table}[htbp]
	\begin{center} 
		\renewcommand{\arraystretch}{1.2}
		\caption{Quantitative comparison on unpaired datasets with non-reference metrics. The best result is in red whereas the second best one is in blue.}
		\vspace{5pt}
		\label{tab:non_ref_syn_metric}%
%		\smaller
		%				\setlength{\tabcolsep}{1.15mm}
		{
			% Table generated by Excel2LaTeX from sheet 'total'
			\begin{tabular}{|c|c|cccc|c|}
				\hline
				Datasets & Metrics & EnGAN & ZeroDCE & RUAS & SCI & Ours \bigstrut\\
				\hline
				\hline
		\multicolumn{1}{|c|}{\multirow{4}[2]{*}{DICM}} & DE$\uparrow$ & \textcolor[rgb]{ 1,  0,  0}{\textbf{7.173 }} & \textcolor[rgb]{ 0,  0,  1}{\textbf{7.025 }} & 4.367  & 6.752  & 6.962  \bigstrut[t]\\
& EME$\uparrow$ & 14.786  & 29.246  & 14.623  & 44.374  & \textcolor[rgb]{ 1,  0,  0}{\textbf{44.647 }} \\
& LOE$\downarrow$ & 709.359  & 342.108  & 1403.100  & 362.332  & \textcolor[rgb]{ 1,  0,  0}{\textbf{98.421 }} \\
& NIQE$\downarrow$ & \textcolor[rgb]{ 1,  0,  0}{\textbf{3.496 }} & \textcolor[rgb]{ 0,  0,  1}{\textbf{3.564 }} & 7.111  & 4.011  & 3.854  \bigstrut[b]\\
\hline
\hline
\multicolumn{1}{|c|}{\multirow{4}[2]{*}{LIME}} & DE$\uparrow$ & \textcolor[rgb]{ 1,  0,  0}{\textbf{7.313 }} & 7.017  & 6.866  & 7.163  & \textcolor[rgb]{ 0,  0,  1}{\textbf{7.259 }} \bigstrut[t]\\
& EME$\uparrow$ & 15.430  & 35.926  & 19.310  & 38.807  & \textcolor[rgb]{ 1,  0,  0}{\textbf{40.105 }} \\
& LOE$\downarrow$ & 540.728  & 135.035  & 719.906  & 176.336  & \textcolor[rgb]{ 1,  0,  0}{\textbf{79.139 }} \\
& NIQE$\downarrow$ & \textcolor[rgb]{ 1,  0,  0}{\textbf{3.658 }} & \textcolor[rgb]{ 0,  0,  1}{\textbf{3.769 }} & 5.358  & 4.165  & 4.380  \bigstrut[b]\\
\hline
\hline
\multicolumn{1}{|c|}{\multirow{4}[2]{*}{MEF}} & DE$\uparrow$ & \textcolor[rgb]{ 1,  0,  0}{\textbf{7.301 }} & 7.056  & 6.216  & 7.098  & \textcolor[rgb]{ 0,  0,  1}{\textbf{7.214 }} \bigstrut[t]\\
& EME$\uparrow$ & 16.801  & 38.975  & 17.727  & 40.896  & \textcolor[rgb]{ 1,  0,  0}{\textbf{44.167 }} \\
& LOE$\downarrow$ & 589.343  & 164.262  & 784.168  & 200.329  & \textcolor[rgb]{ 1,  0,  0}{\textbf{57.198 }} \\
& NIQE$\downarrow$ & \textcolor[rgb]{ 1,  0,  0}{\textbf{3.221 }} & \textcolor[rgb]{ 0,  0,  1}{\textbf{3.283 }} & 5.426  & 3.681  & 3.644  \bigstrut[b]\\
\hline
\hline
\multicolumn{1}{|c|}{\multirow{4}[2]{*}{NPE}} & DE$\uparrow$ & 7.361  & \textcolor[rgb]{ 0,  0,  1}{\textbf{7.389 }} & 4.523  & 7.318  & \textcolor[rgb]{ 1,  0,  0}{\textbf{7.526 }} \bigstrut[t]\\
& EME$\uparrow$ & 12.985  & 27.063  & 17.418  & 29.476  & \textcolor[rgb]{ 1,  0,  0}{\textbf{33.893 }} \\
& LOE$\downarrow$ & 749.139  & 161.408  & 1357.500  & 462.794  & \textcolor[rgb]{ 1,  0,  0}{\textbf{85.321 }} \\
& NIQE$\downarrow$ & 4.120  & \textcolor[rgb]{ 1,  0,  0}{\textbf{3.926 }} & 7.083  & 4.180  & 4.046  \bigstrut[b]\\
\hline
\hline
\multicolumn{1}{|c|}{\multirow{4}[2]{*}{VV}} & DE$\uparrow$ & \textcolor[rgb]{ 1,  0,  0}{\textbf{7.549 }} & \textcolor[rgb]{ 0,  0,  1}{\textbf{7.444 }} & 5.223  & 7.196  & 7.222  \bigstrut[t]\\
& EME$\uparrow$ & 7.759  & 14.947  & 6.337  & 20.871  & \textcolor[rgb]{ 1,  0,  0}{\textbf{22.336 }} \\
& LOE$\downarrow$ & 461.173  & 145.436  & 583.704  & 128.578  & \textcolor[rgb]{ 1,  0,  0}{\textbf{37.580 }} \\
& NIQE$\downarrow$ & 4.139  & 3.213  & 5.303  & \textcolor[rgb]{ 1,  0,  0}{\textbf{2.753 }} & \textcolor[rgb]{ 0,  0,  1}{\textbf{2.773 }} \bigstrut[b]\\
\hline
			\end{tabular}%
		}
	\end{center} 
\end{table}%

		\textbf{Computational Efficiency.} Our proposed method is characterized by high efficiency
		%\footnote{Further details about the model are provided in the \ref{app:size_eff}.}
		. Specifically, the model size is only 7.7046M, and the GPU runtime is 0.0096s. At full speed, our method can perform high-performance enhancement at a rate of up to 100 frames per second.
		
		In experiments conducted on standard low-light datasets, our method exhibited evident superiority in both qualitative and quantitative enhancement of image quality. Leveraging these outcomes, we expanded the application of our technology to more demanding and practical scenarios.

	\subsection{Ablation Studies}
	We conducted ablation studies to investigate individual network components,  noise estimator and reference-free loss functions. 
	
		\begin{table}[htbp]
		%		\small
		\centering
		%	\begin{threeparttable}
			\renewcommand{\arraystretch}{1.2}
			\caption{Ablation study toward network modules (i.e., SCD and LII)  on the LOL dataset. The best result is in red whereas the second best one is in blue.}
			\vspace{5pt}
%			\smaller
			%			\vspace{-0.2cm}
			% $\dagger$ denotes the original enhanced network (i.e., ZeroDCE and SCI) with the spliced SCD at the forefront.}
		\label{tab:Abl_1}% 
				\tabcolsep=0.045\linewidth
				\begin{tabular}{|c|cccc|}
					\hline
					\textit{Method} & PSNR$\uparrow$ & SSIM$\uparrow$ & LOE$\downarrow$ &NIQE$\downarrow$\\
					\hline 
					\hline
					\multicolumn{5}{|c|}{\textit{Config.: M1 (w/o SCD), M2 (w/o LII, w/SCI), }} \\
					\multicolumn{5}{|c|}{\textit{M3 (w/o $\psi(\nabla^n \mathbf{x})$), M4 (LII-to-SCD)}} \\
					\hline
					M1 & 18.1820 &	0.5175 &	\textcolor[rgb]{ 1,  0,  0}{\textbf{195.1463}} &	8.7650 	 \\
					M2 & 15.2660 &	0.5162 	& 1277.8000 &	5.8762 \\
					M3 & \textcolor[rgb]{ 0,  0,  1}{\textbf{20.0589}} &  \textcolor[rgb]{ 0,  0,  1}{\textbf{0.7009}}  & 373.0177  & \textcolor[rgb]{ 0,  0,  1}{\textbf{5.1661}} \\
					M4 & 19.2083 &	0.6783  &	418.1006 &	\textcolor[rgb]{ 1,  0,  0}{\textbf{4.8808}}  \\ 
					\hline				
					\hline				
					Ours &\textcolor[rgb]{ 1,  0,  0}{\textbf{20.3721}}&	\textcolor[rgb]{ 1,  0,  0}{\textbf{0.7183}}	&\textcolor[rgb]{ 0,  0,  1}{\textbf{356.2606}}&	5.4132\\ 
					\hline
				\end{tabular}%
				%				\vspace{-0.2cm}
				%			} 
		\end{table}% 
	
	%		\noindent
	\textbf{Effects of SCD and LII.}  
	Table~\ref{tab:Abl_1} presents the results of network module (i.e., SCD and LII) ablation on the LOL dataset. It should be noted that in the table, [M1] refers to the model without the SCD module, and [M2] represents the removal of the LII module, which is replaced by the unsupervised illumination learning network proposed in the recent work SCI~\cite{ma2022toward}. We observe that without the SCD module, the performance decreased by $2.1901dB$ and $0.2008$ on the PSNR and SSIM metrics, respectively. Moreover, the substitution of LII for SCI significantly attenuates the performance on all metrics, especially with a $5.1061dB$ decrease in PSNR. We argue that the root cause of this phenomenon is that the substituted Model [M2] does not comply with the loss optimization procedure under the illumination prior constraint condition with a simple loss function. In the other hand, it indicates that our LII module satisfy the illumination prior naturally.

		%		\noindent
		\textbf{High-Order Gradient-Based Noise Estimator.} 
		Having examined the accuracy of the noise estimator, we subsequently investigate its influence on the enhanced results. To be specific, we remove the noise estimator from denoiser and formulated Model [M3] for experimental purposes, as shown in Table \ref{tab:Abl_1}. The outcomes reveal that the noise estimator can proficiently eliminate noise and boost the performance of our approach.
		
				\begin{table}[htbp]
			\centering
			%	\begin{threeparttable}
				\renewcommand{\arraystretch}{1.2}
				\caption{Ablation study regarding the proposed reference-free loss (i.e., $\mathcal{L}_{srr} ~and ~\mathcal{L}_{nr}$) on the LOL dataset. The best result is in red whereas the second best one is in blue.}
				\vspace{5pt}
%				\smaller
				%				\vspace{-0.2cm}
				% $\dagger$ denotes the original enhanced network (i.e., ZeroDCE and SCI) with the spliced SCD at the forefront.}
			\label{tab:Abl_2}% 

				\tabcolsep = 0.019\linewidth
				\begin{tabular}{|c|cccccc|cc|}
					\hline
					\textit{Method} & SCD &$\mathcal{L}_{DCE}$&$\mathcal{L}_{SCI}$&$\mathcal{L}_{srr}$&$\mathcal{L}_{TV}$&$\mathcal{L}_{nr}$ & PSNR$\uparrow$ & SSIM$\uparrow$\\
					\hline
					\hline
					M5 &  & \ding{51} & & & & & 10.8484 &	0.2553  \\ 				
					M6 & & & \ding{51} & & & & 10.5643 &	0.2835  \\
					M7 & & & & \ding{51} & & & 18.1820  &	0.5175	 \\ %w/ TV + Fid
					M8 & \ding{51} & & & \ding{51} & \ding{51} & & \textcolor[rgb]{ 0,  0,  1}{\textbf{20.1610}} & \textcolor[rgb]{ 0,  0,  1}{\textbf{0.6023}} \\
					\hline				
					\hline				
					Ours & \ding{51} & & & \ding{51} & & \ding{51} &\textcolor[rgb]{ 1,  0,  0}{\textbf{20.3721}}&	\textcolor[rgb]{ 1,  0,  0}{\textbf{0.7183}} \\ 
					\hline
				\end{tabular}%
				%			}
			%				\vspace{-0.2cm}
			%		\begin{tablenotes}
				%			\footnotesize 
				%			\item[] $\dagger$ denotes the original enhanced network (i.e., ZeroDCE and SCI) with the spliced SCD at the forefront.
				%		\end{tablenotes} 
			%	\end{threeparttable}	
	\end{table}% 

		%		\noindent
		\textbf{SCD-to-LII v.s. LII-to-SCD.} We explore the effects of executive order of SCD, and construct model [M4] where its front LII is spliced with a post-processor denoiser plug-in.  The corresponding quantitative is reported in Table~\ref{tab:Abl_1}, showing that using ``SCD-to-LII''  improves the performance of LLIE to a certain extent on three performance metrics, compared with ``LII-to-SCD''.

\begin{figure}[htbp]
	\begin{center}
		\begin{tabular}{c@{\extracolsep{0.35em}}c@{\extracolsep{0.35em}}c@{\extracolsep{0.35em}}c@{\extracolsep{0.35em}}} 			
			
			\includegraphics[width=0.232\linewidth]{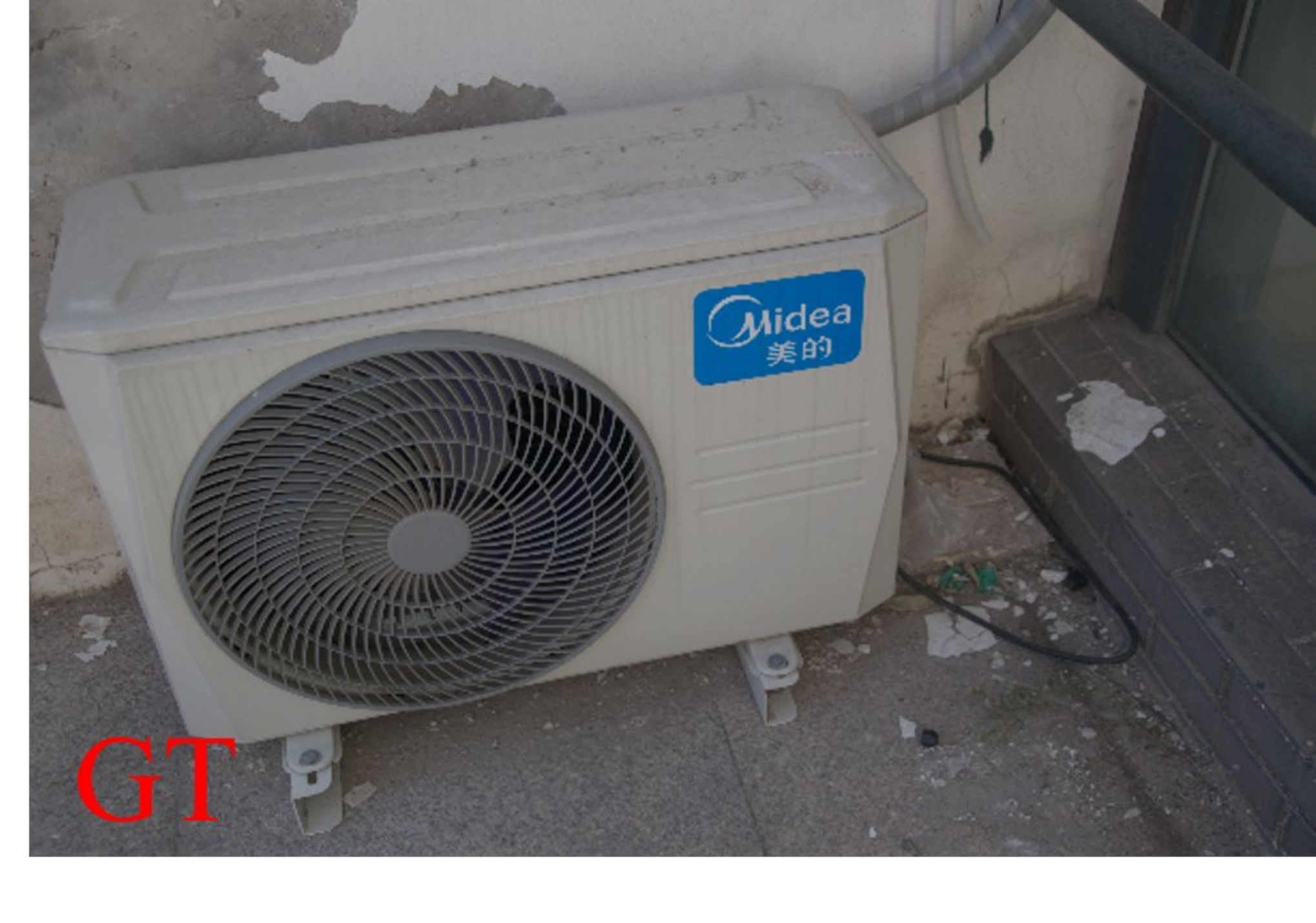}   & \includegraphics[width=0.232\linewidth]{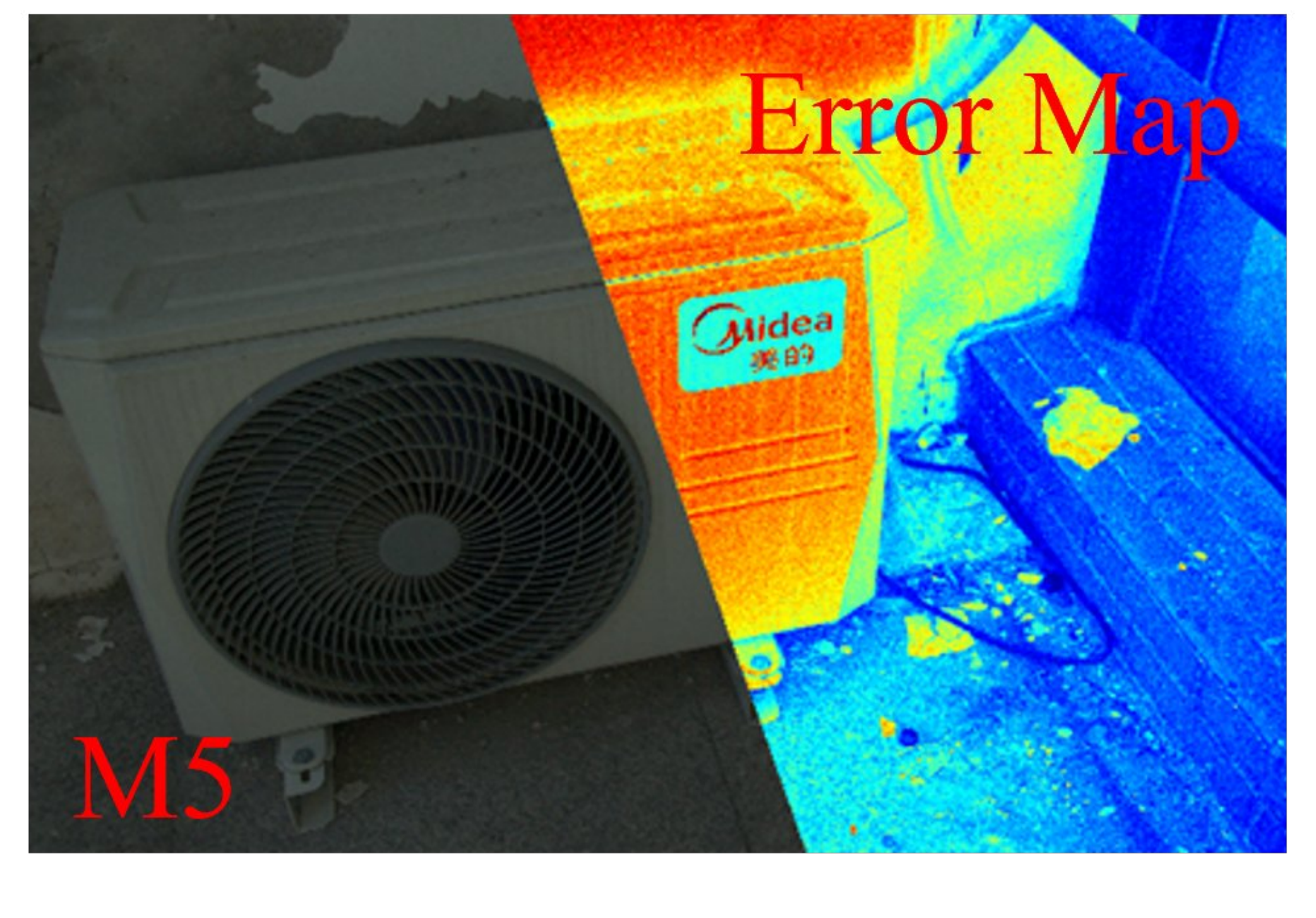}   & \includegraphics[width=0.232\linewidth]{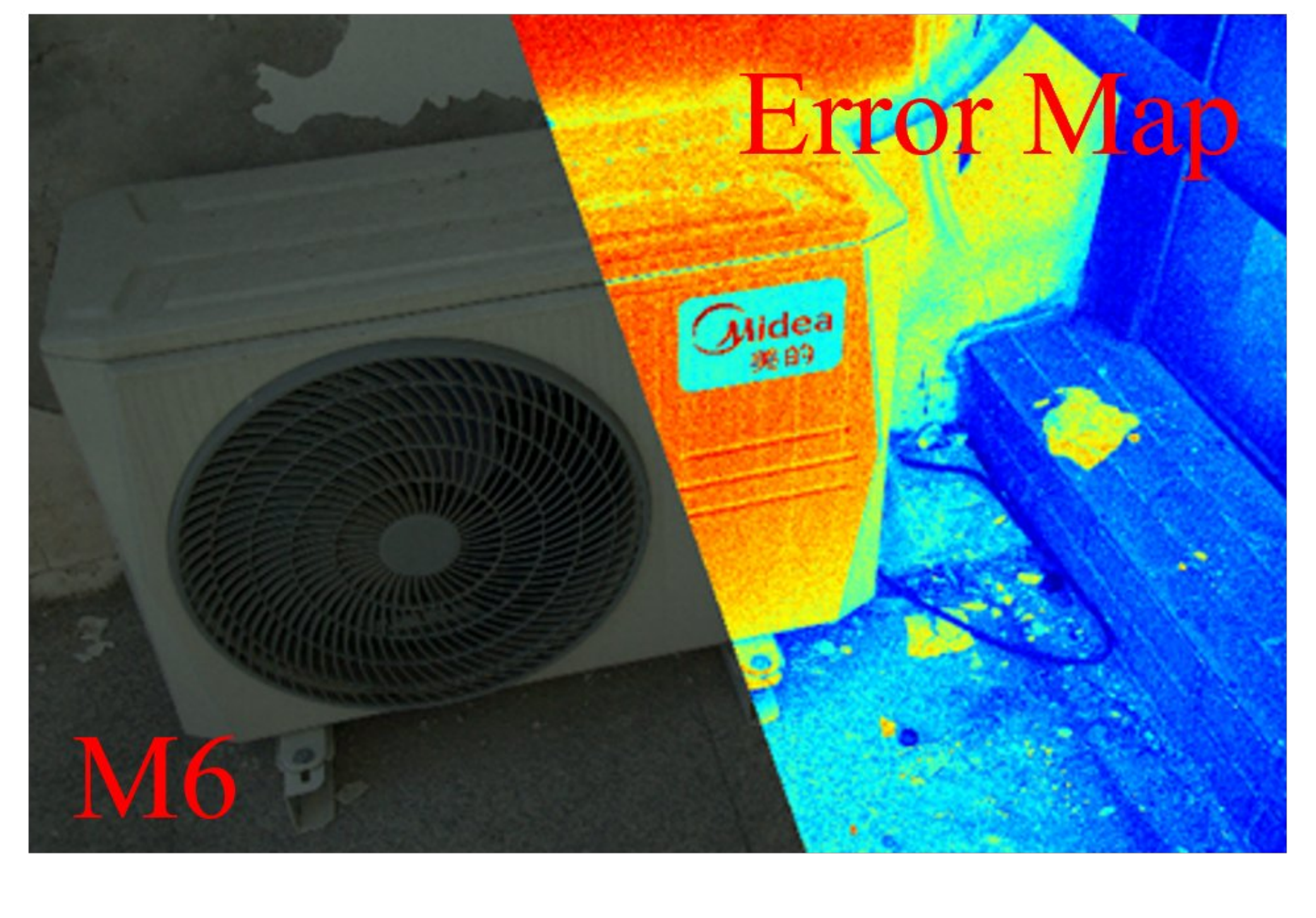} 
			& \includegraphics[width=0.232\linewidth]{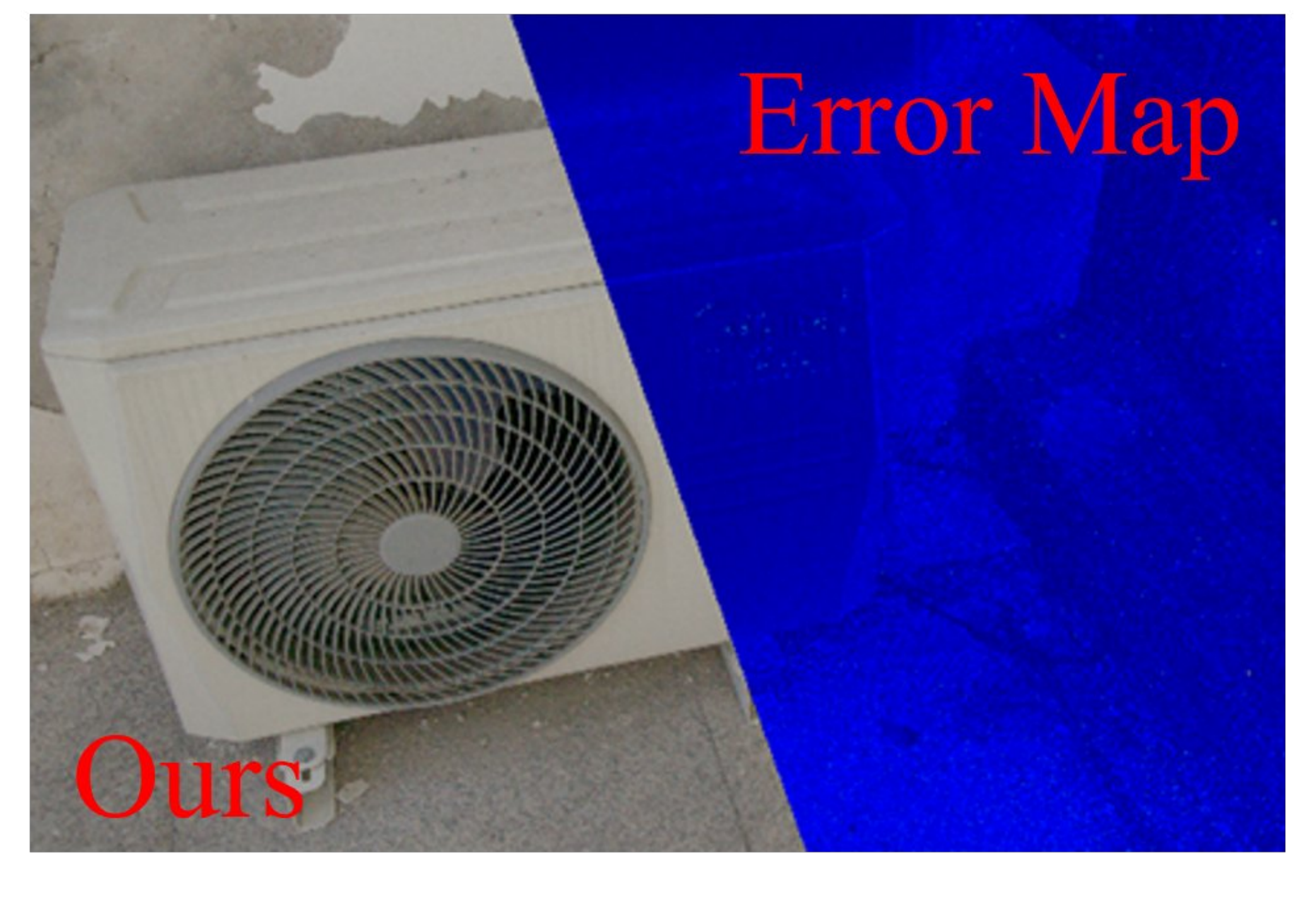}  \\ 			
			\specialrule{0em}{-0.5pt}{-0pt} 
			\footnotesize --- & \footnotesize (13.615, 0.620) &  \footnotesize (13.657, 0.611) &  \footnotesize \textbf{(22.389, 0.754)} \\ \specialrule{0em}{-0.5pt}{-0pt}  
			
			\includegraphics[width=0.232\linewidth]{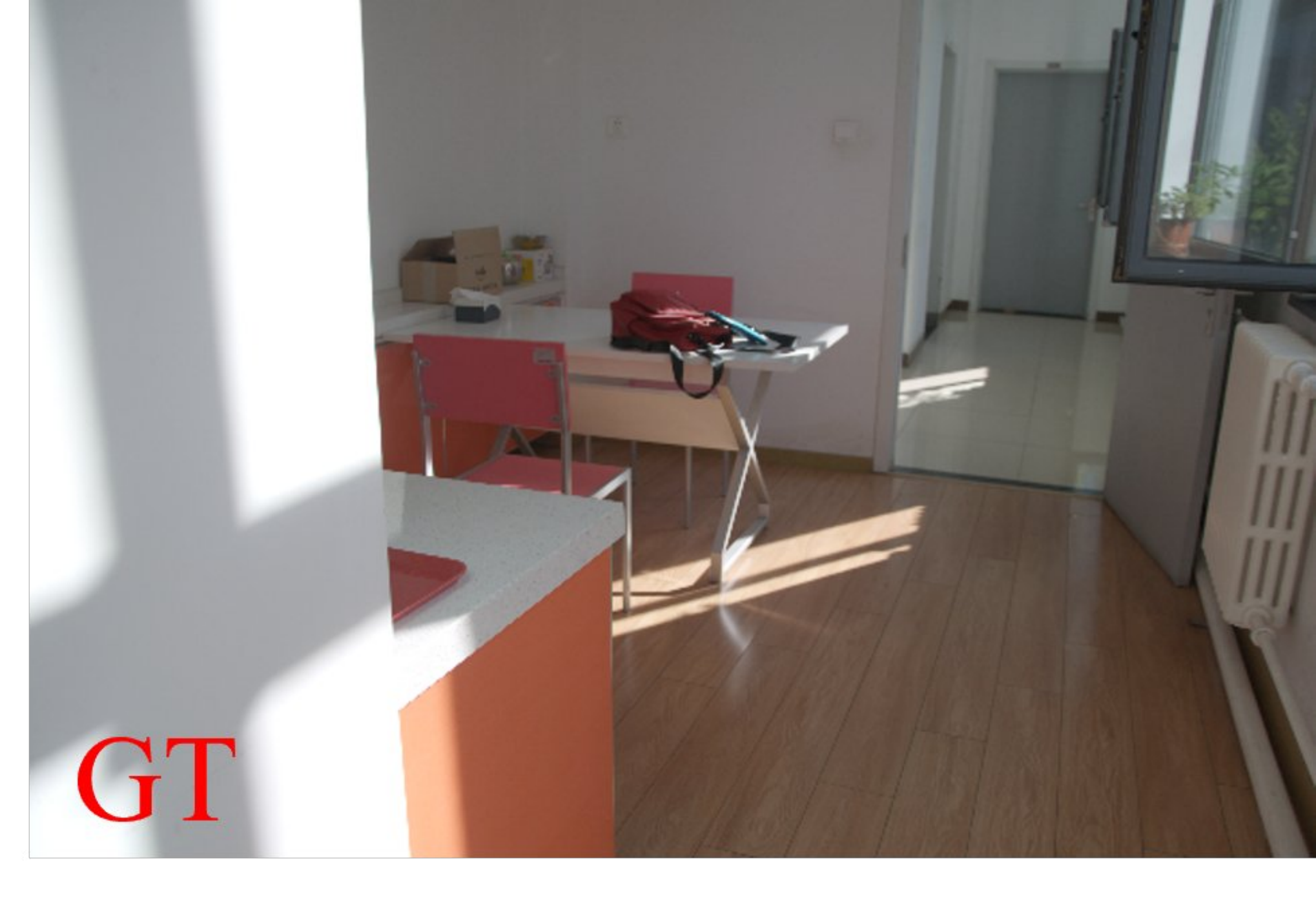}   & \includegraphics[width=0.232\linewidth]{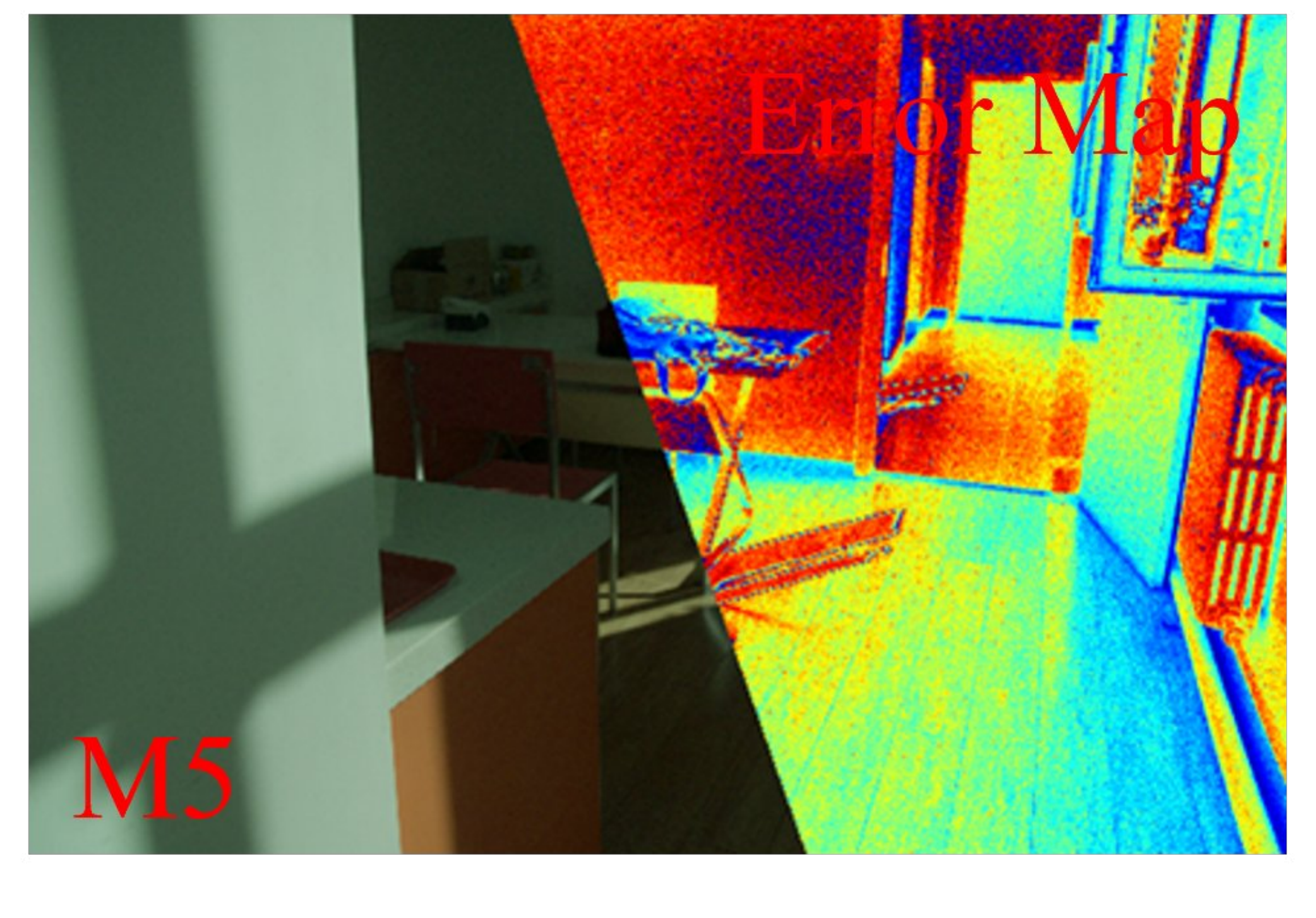}   & \includegraphics[width=0.232\linewidth]{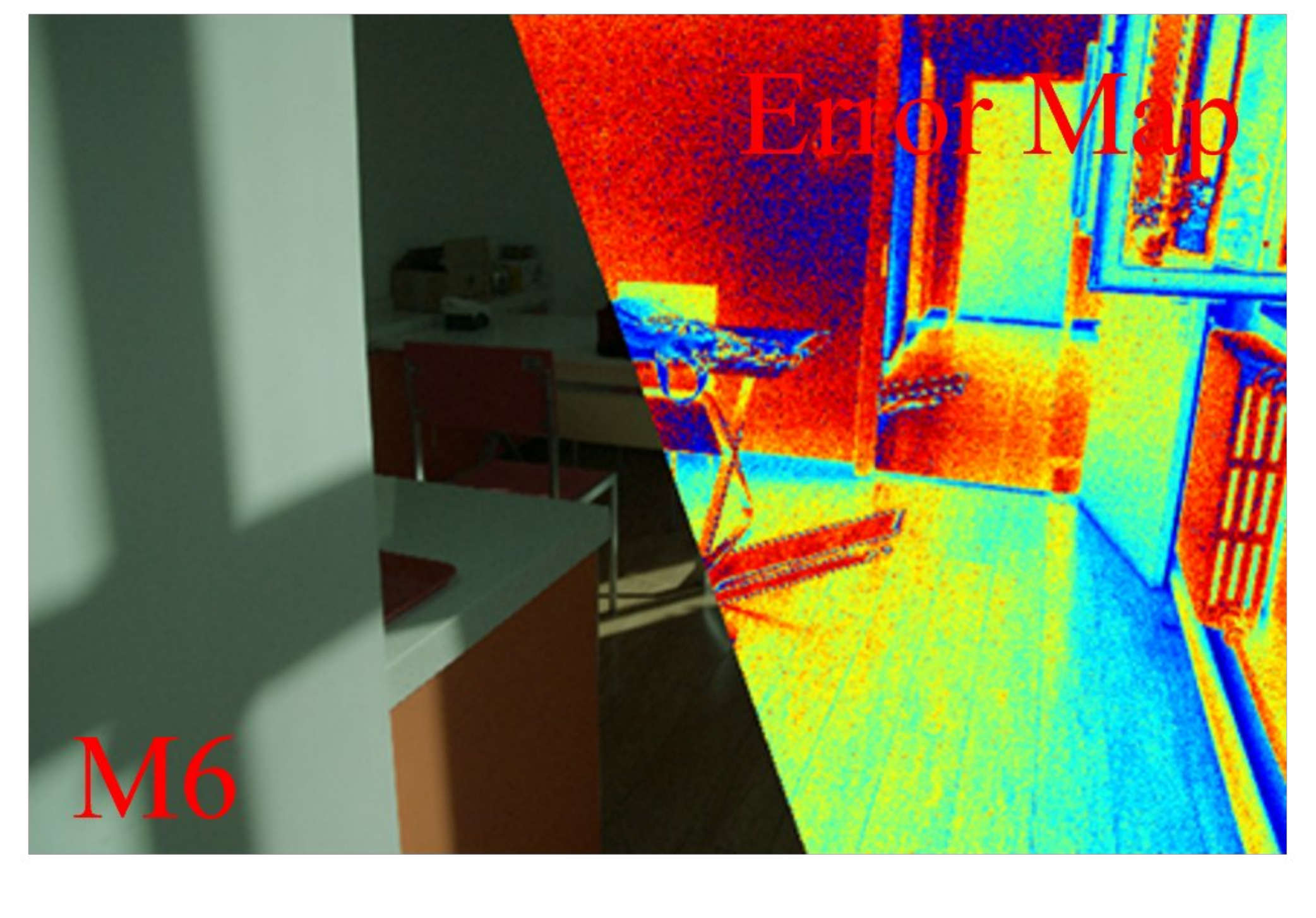} 
			& \includegraphics[width=0.232\linewidth]{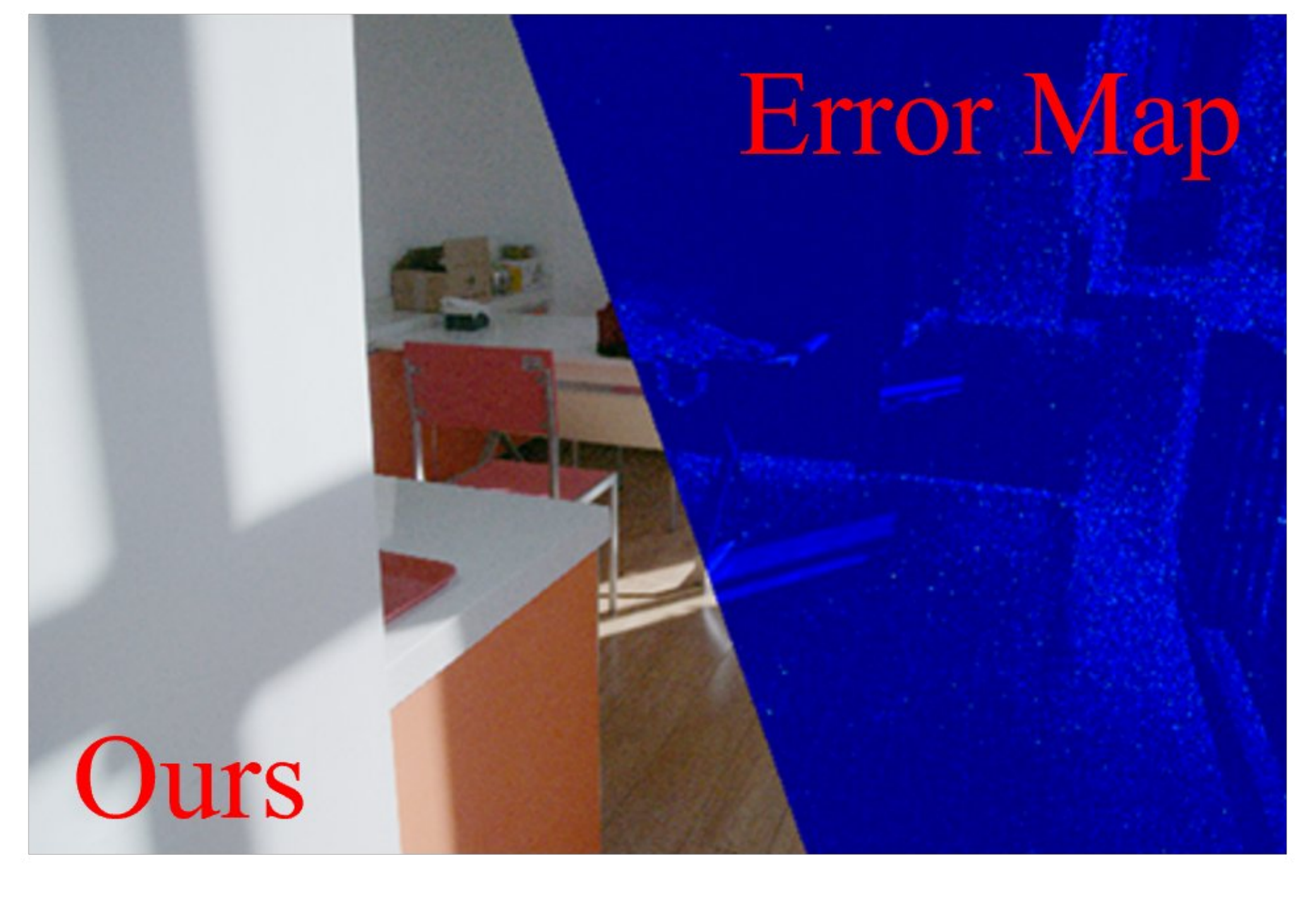}  \\ \specialrule{0em}{-0.5pt}{-1pt} 
			\footnotesize --- & \footnotesize (10.421, 0.340) &  \footnotesize (10.323, 0.368) &   \footnotesize \textbf{(24.779, 0.7614)} \\ 
			%			Reference & M4 & M5 & Ours\\		
			\specialrule{0em}{-2pt}{-2pt} 
		\end{tabular}
	\end{center}
	\vspace{-0.2cm}
	\caption{ 
		Ablation study on the impact of reference-free loss. (PSNR$\uparrow$, SSIM$\uparrow$) are listed at the bottom to quantify the generated image quality, and a scaled heatmap of the square error between ground truth and result is concatenated to the right part of enhanced image for better comparison.
		%		For improved comparison, we present each image as a pair comprising the enhanced result on the left and a scaled heatmap of the square error on the right (darker is better).
	}
	\label{fig:loss}\vspace{-0.4cm}
\end{figure} 
%\noindent
\textbf{Exploration on Reference-Free Loss.}  
%As discussed, we elaborate a more complete system of noise removel and illuminance recovery.
Our study investigates the effectiveness of self-regularized recovery loss in comparison with popular unsupervised learning loss functions, including $\mathcal{L}_{DCE}$ from ZeroDCE~\cite{guo2020zero} and $\mathcal{L}_{SCI}$ from SCI~\cite{ma2022toward}, as shown in Table~\ref{tab:Abl_2}. We built Model [M5] and [M6] separately using the corresponding complex loss configurations. 
%	As discussed, we have adhered to a single guiding principle when designing the illuminance recovery loss: simplicity is paramount, so the hyperparameters of the loss functions are utilized without any coefficient adjustment. 
Hyperparameters of the loss functions are utilized without any coefficient adjustment for evaluate its robustness.
However, due to the lack of a denoising function of these model, we remove the $\mathcal{L}_{nr}$ as well as SCD term and construct Model [M7] for justice. None of the models include an SCD module. Our quantitative results show a substantial improvement in Model [M7] over Models [M5] and [M6], such as a  $67.6\%$ PSNR score improvement, demonstrating robustness and effectiveness in color restoration, as indicated in Figure \ref{fig:loss} and Table \ref{tab:Abl_2}. Additionally, we explore self-adaptable noise removal loss by removing the adaptive factor $\sigma(\mathbf{x})$ from $\mathcal{L}_{nr}$ to train Model [M8], which leads to the degradation of the loss to the most common denoise loss $\mathcal{L}_{TV}$~\cite{rudin1992nonlinear}. Our final model shows improvements in the PSNR and SSIM metrics of $0.2111dB$ and $0.1160$, respectively.

\section{Conclusion}
%Based on the characteristics of low-light images, we propose a formula for rapidly estimating Gaussian noise in low-light images, empowering denoising solutions and emphasizing the enhancement of denoiser performance in low-light scenarios.
%Drawing inspiration from the retinex theory's partitioning scheme, we present a structure-constrained noise-aware illumination interpolator, coupled with a reference-free loss function that is intuitive yet effective. 
%%The proposed approach imposes natural constraints on the illumination of the enhanced image, thereby  preserving the intrinsic relationships between pixels, including edges and semantic entities. 
%Our network's interpretability and the predictability of the enhanced results are attributed to the utilization of these simple yet effective approaches.
%In various datasets, our proposed method outperforms state-of-the-art 
%%	unsupervised methods and even achieves comparability with supervised 
%methods in most metrics. 
%%Moreover, the enhanced results generated by our approach exhibit excellent performance when applied to other low-light datasets, thereby facilitating high-level visual tasks, such as segmentation.

Drawing upon the distinctive characteristics of low-light images, we have devised a rapid Gaussian noise estimation formula tailored specifically for dynamic dark scenarios. This formula empowers denoising solutions and prioritizes enhancing denoiser performance in low-light conditions. Inspired by the partitioning scheme of the retinex theory, we propose a structure-constrained noise-aware illumination interpolator, complemented by an intuitive yet effective reference-free loss function. The simplicity and efficacy of these strategies enhance the interpretability of our network and the predictability of the enhanced results. Our proposed method consistently surpasses state-of-the-art methodologies across various metrics on multiple datasets. 
Notably, the noise estimation approach's lack of requirement for additional sampling operations and its ease of integration into neural networks suggest potential benefits for diffusion models and image information entropy estimation in future applications. Building upon these contributions, our future endeavors will focus on refining data inline distribution to achieve even better enhancement performance.

\section{Acknowledgement}
This work is partially supported by the National Natural Science Foundation of China (Nos. U22B2052 and 62027826) and the Fundamental Research
Funds for the Central Universities.

%During the preparation of this work authors used ChatGPT3.5 in order to polish the paper. After using this tool, authors reviewed and edited the content as needed and take full responsibility for the content of the publication.

\bibliographystyle{elsarticle-num} 
\bibliography{reference}

\end{document}